\documentclass{article} %
\usepackage{iclr2020_conference,times}

\usepackage{amsmath,amsfonts,bm}

\def\eqref#1{equation~\ref{#1}}

\def\1{\bm{1}}

\DeclareMathAlphabet{\mathsfit}{\encodingdefault}{\sfdefault}{m}{sl}
\SetMathAlphabet{\mathsfit}{bold}{\encodingdefault}{\sfdefault}{bx}{n}

\usepackage{hyperref}
\usepackage[utf8]{inputenc}
\usepackage[T1]{fontenc}
\usepackage{booktabs}
\usepackage{graphicx}
\usepackage{hypcap}
\usepackage{nicefrac}
\usepackage{microtype}
\usepackage{multirow}
\usepackage{placeins}
\usepackage{rotating}
\usepackage{subfig}
\usepackage{xcolor}
\usepackage{xspace}

\title{Meta-Dataset: A Dataset of Datasets for Learning to Learn from Few Examples}

\author{%
Eleni Triantafillou$^{*\dagger}$,
Tyler Zhu$^{\dagger}$,
Vincent Dumoulin$^{\dagger}$,
Pascal Lamblin$^{\dagger}$,
Utku Evci$^{\dagger}$, \\ \bf
Kelvin Xu$^{\ddagger\dagger}$,
Ross Goroshin$^{\dagger}$,
Carles Gelada$^{\dagger}$,
Kevin Swersky$^{\dagger}$, \\ \bf
Pierre-Antoine Manzagol$^{\dagger}$ \&
Hugo Larochelle$^{\dagger}$ \\
$^{*}$University of Toronto and Vector Institute,
$^{\dagger}$Google AI,
$^{\ddagger}$University of California, Berkeley \\
Correspondence to: \texttt{eleni@cs.toronto.edu} \\
}

\newcommand{\benchmark}{\textsc{Meta-Dataset}\xspace}

\iclrfinalcopy %
\begin{document}

\maketitle

\begin{abstract}
Few-shot classification refers to learning a classifier for new classes given
only a few examples. While a plethora of models have emerged to tackle it, we find the procedure and datasets that are used to assess their progress lacking. To address this limitation, we propose \benchmark: a new benchmark for training and evaluating models that is large-scale, consists of diverse datasets, and presents
more realistic tasks. We experiment with popular baselines and meta-learners on \benchmark, along with a competitive method that we propose. We analyze performance as a function of various characteristics of test tasks and examine the models' ability to leverage diverse training sources for improving their generalization. We also propose a new set of baselines for quantifying the benefit of meta-learning in \benchmark. Our extensive experimentation has uncovered important research challenges and we hope to inspire work in these directions. 
\end{abstract}

\section{Introduction}
Few-shot learning refers to learning new concepts from few examples, an ability
that humans naturally possess, but machines still lack. Improving on this
aspect would lead to more efficient algorithms that can flexibly expand their
knowledge without requiring large labeled datasets. We focus on
few-shot classification: classifying unseen examples into one of $N$ new `test'
classes, given only a few reference examples of each. Recent progress
in this direction has been made by considering a meta-problem: though we are
not interested in learning about any training class in particular, we can exploit the training classes for the purpose of {\em learning to learn new
classes from few examples}, thus acquiring a learning procedure that can be
directly applied to new few-shot learning problems too.

This intuition has inspired numerous models of increasing complexity (see Related Work for some examples). However, we believe that the
commonly-used setup for measuring success in this direction is lacking.
Specifically, two datasets have emerged as {\em de facto}
benchmarks for few-shot learning: Omniglot~\citep{lake2015human}, and {\em
mini}-ImageNet~\citep{vinyals2016matching}, and we believe that both of them
are approaching their limit in terms of allowing one to discriminate between
the merits of different approaches. Omniglot is a dataset of 1623 handwritten
characters from 50 different alphabets and contains 20 examples per class
(character). Most recent methods obtain very high accuracy on Omniglot,
rendering the comparisons between them mostly uninformative. {\em
mini}-ImageNet is formed out of 100 ImageNet~\citep{russakovsky2015imagenet}
classes (64/16/20 for train/validation/test) and contains 600 examples per
class. Albeit harder than Omniglot, it has the same property that most recent
methods trained on it present similar accuracy when controlling for model
capacity. We advocate that a more challenging and realistic benchmark is
required for further progress in this area.

More specifically, current benchmarks: 1) Consider homogeneous learning tasks. In contrast, real-life learning experiences are heterogeneous: they vary in terms of the number of classes and examples per class, and are unbalanced. 2) Measure only within-dataset generalization. However, we are eventually after models that can generalize to entirely new distributions (e.g., datasets). 3) Ignore the relationships between classes when forming episodes. Specifically, the coarse-grained classification of dogs and chairs may present different difficulties than the fine-grained classification of dog breeds, and current benchmarks do not establish a distinction between the two.

\benchmark aims to improve upon previous benchmarks in the above directions: it
is significantly larger-scale and is comprised of multiple datasets of diverse
data distributions; its task creation is informed by class structure for
ImageNet and Omniglot; it introduces realistic class imbalance; and it varies
the number of classes in each task and the size of the training set, thus
testing the robustness of models across the spectrum from very-low-shot
learning onwards.

The main contributions of this work are:
1) A more realistic, large-scale and diverse environment for training and testing few-shot learners.
2) Experimental evaluation of popular models, and a new set of baselines combining inference algorithms of meta-learners with non-episodic training.
3) Analyses of whether different models benefit from more data, heterogeneous training sources, pre-trained weights, and meta-training.
4) A novel meta-learner that performs strongly on \benchmark.
\section{Few-shot Classification: Task Formulation and Approaches}

\paragraph{Task Formulation} The end-goal of few-shot classification is to
produce a model which, given a new learning {\em episode} with $N$ classes and
a few labeled examples ($k_c$ per class, $c \in {1, \ldots, N}$), is able to
generalize to unseen examples for that episode. In other words, the model
learns from a training \emph{(support)} set $\mathcal{S} = \{(\mathbf{x}_1,
y_1), (\mathbf{x}_2, y_2), \ldots, (\mathbf{x}_K, y_K)\}$ (with $K = \sum_c
k_c$) and is evaluated on a held-out test \emph{(query)} set $\mathcal{Q} =
\{(\mathbf{x}^*_1, y^*_1), (\mathbf{x}^*_2, y^*_2), \ldots, (\mathbf{x}^*_T,
y^*_T)\}$. Each example $(\mathbf{x}, y)$ is formed of an input vector
$\mathbf{x} \in \mathbb{R}^D$ and a class label $y \in \{1, \ldots, N\}$.
Episodes with balanced training sets (i.e., $k_c = k, ~\forall c$) are usually
described as `$N$-way, $k$-shot' episodes. Evaluation episodes are constructed by sampling their $N$ classes from a
larger set $\mathcal{C}_{test}$ of classes and sampling the desired number of
examples per class. 

A disjoint set $\mathcal{C}_{train}$ of classes is available to
train the model; note that this notion of training is distinct from the
training that occurs within a few-shot learning episode. Few-shot learning does not prescribe a specific procedure for exploiting $\mathcal{C}_{train}$, but a
common approach matches the conditions in which the model is trained
and evaluated~\citep{vinyals2016matching}. In other words, training often (but
not always) proceeds in an episodic fashion. Some authors use {\em training}
and {\em testing} to refer to what happens {\em within} any given episode, and
{\em meta-training} and {\em meta-testing} to refer to using
$\mathcal{C}_{train}$ to turn the model into a  learner capable of fast
adaptation and $\mathcal{C}_{test}$ for evaluating its success to learn using
few shots, respectively. This nomenclature highlights the {\em meta-learning} perspective
alluded to earlier, but to avoid confusion we will adopt another common
nomenclature and refer to the training and test sets of an episode as the {\em
support} and {\em query} sets and to the process of learning from
$\mathcal{C}_{train}$ simply as training. We use the term `meta-learner' to
describe a model that is trained episodically, i.e., learns to learn across
multiple tasks that are sampled from the training set $\mathcal{C}_{train}$.

\paragraph{Non-episodic Approaches to Few-shot Classification} A natural non-episodic approach simply 
trains a classifier over all of the training classes $\mathcal{C}_{train}$ at once, which
can be parameterized by a neural network with a linear layer on top with one output unit per class.
After training, this neural network is used as an embedding function $g$ that maps images into
a meaningful representation space. The hope of using this model for few-shot learning is that
this representation space is useful even for examples of classes that were not included in training.
It would then remain to define an algorithm for performing few-shot classification on top of these
representations of the images of a task. We consider two choices for this algorithm, yielding 
the `$k$-NN' and `Finetune' variants of this baseline.

Given a test episode, the `$k$-NN' baseline classifies each query example as the class that its `closest' support example belongs to. Closeness is measured by either Euclidean or cosine distance in the 
learned embedding space; a choice that we treat as a hyperparameter. On the other hand, the 
`Finetune' baseline uses the support set of the given test episode to train a new `output
layer' on top of the embeddings $g$, and optionally finetune those embedding too (another hyperparameter), for the purpose of classifying between the $N$ new classes of the associated task. 

A variant of the `Finetune' baseline has recently become popular:
Baseline++ \citep{chen2019closer}, originally inspired by \citet{GidarisS2018,qi2018low}.
It uses a `cosine classifier' as the final layer
($\ell^2$-normalizing embeddings and weights before taking the dot product),
both during the non-episodic training phase, and for evaluation on test episodes.
We incorporate this idea in our codebase by adding a hyperparameter that optionally enables using a cosine classifier for the `$k$-NN' (training only) and `Finetune' (both phases) baselines.

\paragraph{Meta-Learners for Few-shot Classification} In the episodic setting,
models are trained end-to-end for the purpose of learning to build classifiers
from a few examples. We choose to experiment with Matching Networks
\citep{vinyals2016matching}, Relation Networks \citep{sung2018learning},
Prototypical Networks \citep{snell2017prototypical} and Model Agnostic
Meta-Learning (MAML, \citealp{finn2017model}) since they cover a diverse set of
approaches to few-shot learning. We also introduce a novel meta-learner which
is inspired by the last two models.

In each training episode, episodic models compute for each query example
$\mathbf{x}^* \in \mathcal{Q}$, the distribution for its label $p(y^*
|\mathbf{x}^*, \mathcal{S})$ conditioned on the support set $\mathcal{S}$ and
allow to train this differentiably-parameterized conditional distribution
end-to-end via gradient descent. The different models are distinguished by the
manner in which this conditioning on the support set is realized. In all cases,
the performance on the query set drives the update of the meta-learner's
weights, which include (and sometimes consist only of) the embedding weights.
We briefly describe each method below.

\paragraph{Prototypical Networks} Prototypical Networks construct a prototype
for each class and then classify each query example as the class whose
prototype is `nearest' to it under Euclidean distance. More concretely, the
probability that a query example $\mathbf{x}^*$ belongs to class $k$ is defined
as:
\begin{equation*}
  p(y^* = k | \mathbf{x}^*, \mathcal{S})
  = \frac{\exp(-||g(\mathbf{x}^*) - \mathbf{c}_k||^2_2)}
	 {\sum_{k' \in \{1, \dots, N\}}
    \exp(-||g(\mathbf{x}^*) - \mathbf{c}_{k'}||^2_2)}
\end{equation*}
where $\mathbf{c}_k$ is the `prototype' for class $k$: the average of the
embeddings of class $k$'s support examples.

\paragraph{Matching Networks} Matching Networks (in their simplest form) label
each query example as a (cosine) distance-weighted linear combination of the
support labels:
\begin{equation*}
  p(y^*=k | \mathbf{x}^*, \mathcal{S})
  = \displaystyle \sum_{i=1}^{|\mathcal{S}|} \alpha (\mathbf{x}^*, \mathbf{x}_i) \mathbf{1}_{y_i=k},
\end{equation*}
where $\mathbf{1}_{A}$ is the indicator function and $\alpha(\mathbf{x}^*, \mathbf{x}_i)$
is the cosine similarity between $g(\mathbf{x}^*)$ and
$g(\mathbf{x}_i)$, softmax-normalized over all support examples $\mathbf{x}_i$, where $1 \le i \le |\mathcal{S}|$.
\paragraph{Relation Networks}
Relation Networks are comprised of an embedding function $g$ as usual, and a
`relation module' parameterized by some additional neural network layers. They
first embed each support and query using $g$ and create a prototype
$p_c$ for each class $c$ by averaging its support embeddings. Each prototype
$p_c$ is concatenated with each embedded query and fed through the
relation module which outputs a number in $[0, 1]$ representing the predicted
probability that that query belongs to class $c$.  The query loss is then
defined as the mean square error of that prediction compared to the (binary)
ground truth. Both $g$ and the relation module are trained to minimize this
loss.

\paragraph{MAML} MAML uses a linear layer parametrized by $\mathbf{W}$ and
$\mathbf{b}$ on top of the embedding function $g(\cdot; \theta)$ and
classifies a query example as
\begin{equation*}
  p(y^*| \mathbf{x}^*, \mathcal{S}) = {\rm softmax}(\mathbf{b}' + \mathbf{W}' g(\mathbf{x}^*;\theta')),
\end{equation*}
where the output layer parameters $\mathbf{W}'$ and $\mathbf{b}'$ and the
embedding function parameters $\theta'$ are obtained by performing a small
number of within-episode training steps on the support set $S$, starting from
initial parameter values $(\mathbf{b},\mathbf{W},\theta)$. The model is trained
by backpropagating the query set loss through the within-episode gradient
descent procedure and into $(\mathbf{b},\mathbf{W},\theta)$. This normally
requires computing second-order gradients, which can be expensive to obtain
(both in terms of time and memory). For this reason, an approximation is often
used whereby gradients of the within-episode descent steps are ignored. This
variant is referred to as first-order MAML (fo-MAML) and was used in our
experiments. We did attempt to use the full-order version, but found it to be
impractically expensive (e.g., it caused frequent out-of-memory problems).

Moreover, since in our setting the number of ways varies between episodes,
$\mathbf{b},\mathbf{W}$ are set to zero and are not trained (i.e.,
$\mathbf{b}',\mathbf{W}'$ are the result of within-episode gradient descent
initialized at 0), leaving only $\theta$ to be trained. In other words, MAML
focuses on learning the within-episode initialization $\theta$ of the embedding
network so that it can be rapidly adapted for a new task.

\paragraph{Introducing Proto-MAML} We introduce a novel meta-learner that combines the complementary strengths of Prototypical Networks and MAML: the former's simple inductive bias that is evidently effective for very-few-shot learning, and the latter's flexible adaptation mechanism.

As explained by \citet{snell2017prototypical}, Prototypical Networks can be
re-interpreted as a linear classifier applied to a learned representation
$g(\mathbf{x})$. The use of a squared Euclidean distance means that output
logits are expressed as
\begin{equation*}
  -||g(\mathbf{x}^*) - \mathbf{c}_k||^2
  = -g(\mathbf{x}^*)^Tg(\mathbf{x}^*) + 2\mathbf{c}_k^Tg(\mathbf{x}^*)
     - \mathbf{c}_k^T\mathbf{c}_k
  = 2\mathbf{c}_k^Tg(\mathbf{x}^*) - ||\mathbf{c}_k||^2 + constant
\end{equation*}
where $constant$ is a class-independent scalar which can be ignored, as it
leaves output probabilities unchanged. The $k$-th unit of the equivalent
linear layer therefore has weights $\mathbf{W}_{k,\cdot} = 2\mathbf{c}_k$ and
biases $b_k = -||\mathbf{c}_k||^2$, which are both differentiable with respect
to $\theta$ as they are a function of $g(\cdot; \theta)$.

We refer to (fo-)Proto-MAML as the (fo-)MAML model where the task-specific linear
layer of each episode is initialized from the Prototypical Network-equivalent
weights and bias defined above and subsequently optimized as usual on the given
support set. When computing the update for $\theta$, we allow gradients to
flow through the Prototypical Network-equivalent linear layer initialization.
We show that this simple modification significantly helps the optimization of
this model and outperforms vanilla fo-MAML by a large margin on \benchmark.

\section{\benchmark: A New Few-Shot Classification Benchmark}

\benchmark aims to offer an environment for measuring progress in realistic
few-shot classification tasks. Our approach is twofold: 1) changing the data
and 2) changing the formulation of the task (i.e., how episodes are generated). %
The following sections describe these modifications in detail.
\ificlrfinal
The code is open source and publicly available\footnote{\href
{https://github.com/google-research/meta-dataset}
{github.com/google-research/meta-dataset}}.
\else
Our code is provided with this submission.\footnote{\scriptsize\url{https://storage.googleapis.com/meta-dataset-source-code/meta-dataset-iclr2020.tar.gz}}
\fi

\subsection{\benchmark's Data}

\benchmark's data is much larger in size than any previous benchmark,
and is comprised of \emph{multiple existing datasets}. This invites
research into how diverse sources of data can be exploited by a meta-learner,
and allows us to evaluate a more challenging generalization problem, to new
datasets altogether. Specifically, \benchmark leverages data from the following
10 datasets: ILSVRC-2012 (ImageNet, \citealp{russakovsky2015imagenet}),
Omniglot~\citep{lake2015human}, Aircraft~\citep{maji13finegrained}, CUB-200-2011
(Birds, \citealp{WahCUB_200_2011}), Describable Textures~\citep{cimpoi14describing},
Quick Draw~\citep{jongejan2016quick}, Fungi~\citep{fungi}, VGG
Flower~\citep{Nilsback08}, Traffic Signs~\citep{Houben-IJCNN-2013} and
MSCOCO~\citep{lin2014microsoft}. These datasets were chosen because they are
free and easy to obtain, span a variety of visual concepts (natural and
human-made) and vary in how fine-grained the class definition is. More
information about each of these datasets is provided in the Appendix.

To ensure that episodes correspond to realistic classification
problems, each episode generated in \benchmark uses classes from a single
dataset. Moreover, two of these datasets, Traffic Signs and
MSCOCO, are fully reserved for evaluation, meaning that no classes from them
participate in the training set. The remaining ones contribute some classes to
each of the training, validation and test splits of classes, roughly with 70\%
/ 15\% / 15\% proportions. Two of these datasets, ImageNet and Omniglot,
possess a class hierarchy that we exploit in \benchmark. %
\ificlrfinal
For each dataset, the composition of splits is available online\footnote{\href
{https://github.com/google-research/meta-dataset/tree/master/meta_dataset/dataset_conversion/splits}
{github.com/google-research/meta-dataset/tree/master/meta\_dataset/dataset\_conversion/splits}}.
\fi

\paragraph{ImageNet} ImageNet is comprised of 82,115 `synsets', i.e., concepts of the WordNet ontology, and it provides `is-a' relationships for its synsets, thus
defining a DAG over them. \benchmark uses the 1K synsets that were chosen for 
the ILSVRC 2012 classification challenge 
and defines a new class split for it and a novel procedure for sampling classes from it for episode creation, both informed by its class hierarchy.

Specifically, we construct a sub-graph of the overall DAG whose leaves are the 1K classes of ILSVRC-2012. We then `cut' this sub-graph
into three pieces, for the training, validation, and test splits, such that there is no overlap between the leaves of any of these pieces. For this, we selected the synsets `carnivore' and `device' as the roots of the validation and test sub-graphs, respectively. The leaves that are reachable from `carnivore' and `device' form the sets of the validation and test classes, respectively. All of
the remaining leaves constitute the training classes. This method of splitting
ensures that the training classes are semantically different from the test classes. 
We end up with 712 training, 158 validation and 130 test classes, roughly adhering 
to the standard 70 / 15 / 15 (\%) splits.

\paragraph{Omniglot} This dataset is one of the established benchmarks for
few-shot classification as mentioned earlier. However, contrary to the common setup that flattens and ignores its two-level hierarchy of alphabets and characters, we allow it to influence the episode class selection in \benchmark, yielding finer-grained tasks. 
We also use the original splits proposed in \citet{lake2015human}:
(all characters of) the `background' and `evaluation' alphabets are used for
training and testing, respectively. However, we reserve the 5 smallest
alphabets from the `background' set for validation.

\subsection{Episode Sampling} In this section we outline \benchmark's algorithm for sampling episodes, featuring hierarchically-aware procedures for sampling classes of ImageNet and Omniglot, and an algorithm that yields realistically imbalanced episodes of variable shots and ways. 
The steps for sampling an episode for a given split are: Step 0) uniformly sample a dataset $\mathcal{D}$, Step 1) sample a set of classes $\mathcal{C}$ from the classes of $\mathcal{D}$ assigned to the requested split, and Step 2) sample support and query examples from $\mathcal{C}$.

\paragraph{\underline{Step 1}: Sampling the episode's class set} This procedure differs depending
on which dataset is chosen. For datasets without a known class organization, we
sample the `way' uniformly from the range $[5, \textsc{MAX-CLASSES}]$, where
$\textsc{MAX-CLASSES}$ is either $50$ or as many as there are available. Then
we sample `way' many classes uniformly at random from the requested class split
of the given dataset. ImageNet and Omniglot use class-structure-aware procedures outlined below.

\paragraph{ImageNet class sampling} We adopt a hierarchy-aware sampling
procedure: First, we sample an internal (non-leaf) node uniformly
from the DAG of the given split. The chosen set of classes is
then the set of leaves spanned by that node (or a random subset of it, if more than 50).
We prevent nodes that are too close to the root to be selected as the internal node, 
as explained in more detail in the Appendix. This procedure enables the creation of tasks of varying degrees of
fine-grainedness: the larger the height of the internal node, the more coarse-grained the resulting episode.

\paragraph{Omniglot class sampling} We sample classes from Omniglot by first
sampling an alphabet uniformly at random from the chosen split of alphabets
(train, validation or test). Then, the `way' of the episode is sampled
uniformly at random using the same restrictions as for the rest of the
datasets, but taking care not to sample a larger number than the number of
characters that belong to the chosen alphabet. Finally, the prescribed number
of characters of that alphabet are randomly sampled. This ensures that each
episode presents a within-alphabet fine-grained classification.

\paragraph{\underline{Step 2}: Sampling the episode's examples} Having already selected a set of classes, the choice of the examples from them that will
populate an episode can be broken down into three steps. We provide a high-level description here and elaborate in the Appendix with the accompanying formulas.

\textbf{Step 2a: Compute the query set size } The query set is class-balanced, reflecting
the fact that we care equally to perform well on all classes of an episode. The number
of query images \textit{per class} is set to a number such that all chosen classes
have enough images to contribute that number and still remain with roughly half on their
images to possibly add to the support set (in a later step). This number is capped to 10 images per class.

\textbf{Step 2b: Compute the support set size } We allow each chosen class to contribute to the
support set at most 100 of its remaining examples (i.e., excluding the ones added to the query set).
We multiply this remaining number by a scalar sampled uniformly from the interval $(0, 1]$ to
enable the potential generation of `few-shot' episodes even when multiple images are available,
as we are also interested in studying that end of the spectrum. We do enforce, however, that
each chosen class has a budget for at least one image in the support set, and we cap the total support set size to 500 examples.
\

\textbf{Step 2c: Compute the shot of each class } We now discuss how to distribute the total support set size chosen above across the participating classes. The un-normalized proportion of the support set that will be occupied by a given chosen class is a noisy version of the total number of images of that class in the dataset. This design choice is made in the hopes of obtaining realistic class ratios, under the hypothesis that the dataset class statistics are a reasonable approximation of the real-world statistics of appearances of the corresponding classes. We ensure that each class has at least one image in the support set and distribute the rest according to the above rule.

After these steps, we complete the episode creation process by choosing the
prescribed number of examples of each chosen class uniformly at random to
populate the support and query sets.

\section{Related Work}

In this work we evaluate four meta-learners on \benchmark 
that we believe capture a good diversity of well-established models. Evaluating other few-shot classifiers on \benchmark
is beyond the scope of this paper, but we discuss some additional related models below. 

Similarly to MAML, some train a meta-learner for quick adaptation to new
tasks~\citep{RaviS2017,munkhdalai17a,rusu2018metalearning,NIPS2018_7963}.
Others relate to Prototypical Networks by learning a representation on which
differentiable training can be performed on some form of
classifier~\citep{bertinetto2018metalearning,GidarisS2018,OreshkinB2018}.
Others relate to Matching Networks in that they perform
comparisons between pairs of support and query examples, using either a graph
neural network~\citep{garcia2018fewshot} or an attention
mechanism~\citep{mishra2018a}. Finally, some make use of memory-augmented
recurrent networks~\citep{pmlr-v48-santoro16}, some learn to perform data
augmentation~\citep{hariharan2017low,wang2018low} in a low-shot learning setting,
and some learn to predict the parameters of a large-shot classifier from the
parameters learned in a few-shot setting~\citep{wang2016learning,wang2017learning}.
Of relevance to Proto-MAML is MAML++~\citep{AntoniouA2019}, which consists of a
collection of adjustments to MAML, such as multiple meta-trained inner loop
learning rates and derivative-order annealing. Proto-MAML instead modifies the output weight initialization scheme and could be combined with those adjustments.

Finally, \benchmark relates to other recent image classification benchmarks.
The CVPR 2017 Visual Domain Decathlon Challenge trains a model on 10 different
datasets, many of which are included in our benchmark, and measures its
ability to generalize to held-out examples for those same datasets but does not measure generalization to new classes (or datasets).
\citet{hariharan2017low} propose a benchmark where a model is given abundant data from certain {\em base} ImageNet classes and is tested on few-shot learning {\em novel} ImageNet classes in a way that doesn't compromise its knowledge of the base classes. 
\citet{wang2018low} build upon that benchmark and propose a new evaluation protocol for it. 
\citet{chen2019closer} investigate fine-grained few-shot classification using the CUB
dataset~(\citealp{WahCUB_200_2011}, also featured in our benchmark) and
cross-domain transfer between {\em mini}-ImageNet and CUB. Larger-scale few-shot
classification benchmarks were also proposed using
CIFAR-100~\citep{krizhevsky2009learning,bertinetto2018metalearning,OreshkinB2018},
{\em tiered}-ImageNet~\citep{ren2018meta}, and ImageNet-21k~\citep{dhillon2019baseline}. Compared to these, \benchmark contains the largest set of diverse datasets in the context of few-shot learning and is additionally accompanied by an algorithm for creating learning scenarios from that data that we advocate are more realistic than the previous ones.

\section{Experiments}

\begin{table*}[t]
\scriptsize
\centering
\caption{Few-shot classification results on \benchmark using models \textbf{trained on ILSVRC-2012 only (top)} and \textbf{trained on all datasets (bottom)}.}
\label{tab:iclr2020_tables_without_ci}
\begin{tabular}{|c|c|c|c|c|c|c|c|c|}
\hline
Test Source                 & $k$-NN               & Finetune               & MatchingNet               & ProtoNet               & fo-MAML               & RelationNet                & fo-Proto-MAML \\\hline\hline
ILSVRC                      & 41.03                & 45.78                  & 45.00                     & \bf{50.50}             & 45.51                 & 34.69                      & \bf{49.53}        \\\hline
Omniglot                    & 37.07                & 60.85                  & 52.27                     & 59.98                  & 55.55                 & 45.35                      & \bf{63.37}        \\\hline
Aircraft                    & 46.81                & \bf{68.69}             & 48.97                     & 53.10                  & 56.24                 & 40.73                      & 55.95             \\\hline
Birds                       & 50.13                & 57.31                  & 62.21                     & \bf{68.79}             & 63.61                 & 49.51                      & \bf{68.66}        \\\hline
Textures                    & 66.36                & \bf{69.05}             & 64.15                     & 66.56                  & \bf{68.04}            & 52.97                      & 66.49             \\\hline
Quick Draw                  & 32.06                & 42.60                  & 42.87                     & 48.96                  & 43.96                 & 43.30                      & \bf{51.52}        \\\hline
Fungi                       & 36.16                & 38.20                  & 33.97                     & \bf{39.71}             & 32.10                 & 30.55                      & \bf{39.96}        \\\hline
VGG Flower                  & 83.10                & 85.51                  & 80.13                     & 85.27                  & 81.74                 & 68.76                      & \bf{87.15}        \\\hline
Traffic Signs               & 44.59                & \bf{66.79}             & 47.80                     & 47.12                  & 50.93                 & 33.67                      & 48.83             \\\hline
MSCOCO                      & 30.38                & 34.86                  & 34.99                     & 41.00                  & 35.30                 & 29.15                      & \bf{43.74}        \\\hline\hline
\textbf{Avg. rank}          & 5.7 & 2.9 & 4.65 & 2.65 & 3.7 & 6.55 & \bf{1.85}\\\hline
\end{tabular}

\vspace{\baselineskip}
\begin{tabular}{|c|c|c|c|c|c|c|c|c|}
\hline
Test Source                 & $k$-NN               & Finetune               & MatchingNet               & ProtoNet               & fo-MAML               & RelationNet                & fo-Proto-MAML \\\hline\hline
ILSVRC                      & 38.55                & 43.08                  & 36.08                     & 44.50                  & 37.83                 & 30.89                      & \bf{46.52}        \\\hline
Omniglot                    & 74.60                & 71.11                  & 78.25                     & 79.56                  & 83.92                 & \bf{86.57}                 & 82.69             \\\hline
Aircraft                    & 64.98                & 72.03                  & 69.17                     & 71.14                  & \bf{76.41}            & 69.71                      & 75.23             \\\hline
Birds                       & 66.35                & 59.82                  & 56.40                     & 67.01                  & 62.43                 & 54.14                      & \bf{69.88}        \\\hline
Textures                    & 63.58                & \bf{69.14}             & 61.80                     & 65.18                  & 64.16                 & 56.56                      & \bf{68.25}        \\\hline
Quick Draw                  & 44.88                & 47.05                  & 60.81                     & 64.88                  & 59.73                 & 61.75                      & \bf{66.84}        \\\hline
Fungi                       & 37.12                & 38.16                  & 33.70                     & 40.26                  & 33.54                 & 32.56                      & \bf{41.99}        \\\hline
VGG Flower                  & 83.47                & 85.28                  & 81.90                     & 86.85                  & 79.94                 & 76.08                      & \bf{88.72}        \\\hline
Traffic Signs               & 40.11                & \bf{66.74}             & 55.57                     & 46.48                  & 42.91                 & 37.48                      & 52.42             \\\hline
MSCOCO                      & 29.55                & 35.17                  & 28.79                     & 39.87                  & 29.37                 & 27.41                      & \bf{41.74}        \\\hline\hline
\textbf{Avg. rank}          & 5.05 & 3.6 & 4.95 & 2.85 & 4.25 & 5.8 & \bf{1.5}\\\hline
\end{tabular}
\end{table*}

\paragraph{Training procedure} \benchmark does not prescribe a procedure for learning from the training data.
In these experiments, keeping with the spirit of matching training and testing conditions, we trained
the meta-learners via training episodes sampled using the same algorithm as we used for \benchmark's evaluation episodes, described
above. The choice of the dataset from which to sample the next episode was random uniform. The non-episodic baselines are trained to solve the large classification problem that results from `concatenating' the training classes of all datasets. 

\paragraph{Validation} Another design choice was to perform validation on (the validation split of) ImageNet
only, ignoring the validation sets of the other datasets. The rationale behind
this choice is that the performance on ImageNet has been known to be a good
proxy for the performance on different datasets. 
We used this validation performance to select our hyperparameters, including backbone architectures, image resolutions and model-specific ones. We describe these further in the Appendix.

\paragraph{Pre-training} We gave each meta-learner the opportunity to initialize its embedding function from the
embedding weights to which the $k$-NN Baseline model trained on ImageNet
converged to. We treated the choice of starting from scratch or starting from this initialization as a hyperparameter. For a fair comparison with the baselines, we allowed the non-episodic models to start from this initialization too. This is especially important for the baselines in the case of training on all datasets since it offers the opportunity to start from ImageNet-pretrained weights.

\paragraph{Main results} Table~\ref{tab:iclr2020_tables_without_ci} displays the accuracy of each model on the test set of each dataset, after they were trained on
ImageNet-only or all datasets. Traffic Signs and MSCOCO are not used for training
in either case, as they are reserved for evaluation. 
We propose to use the average (over the datasets) rank of each method as our metric for
comparison, where smaller is better. A method receives rank 1 if it has the
highest accuracy, rank 2 if it has the second highest, and so on. If two models share the best accuracy, they both get rank 1.5, and so on. We find that fo-Proto-MAML is the top-performer according to this metric, Prototypical Networks also perform strongly, and the Finetune Baseline notably presents a worthy opponent\footnote{We improved MAML's performance significantly in the time between the acceptance decision and publication, thanks to a suggestion to use a more aggressive inner-loop learning rate. In the Appendix, we present the older MAML results side by side with the new ones, and discuss the importance of this hyperparameter for MAML.}. We include more detailed versions of these tables displaying confidence intervals and per-dataset ranks in the Appendix. 

\begin{figure*}[h]
\centering
    \includegraphics[scale=0.2]{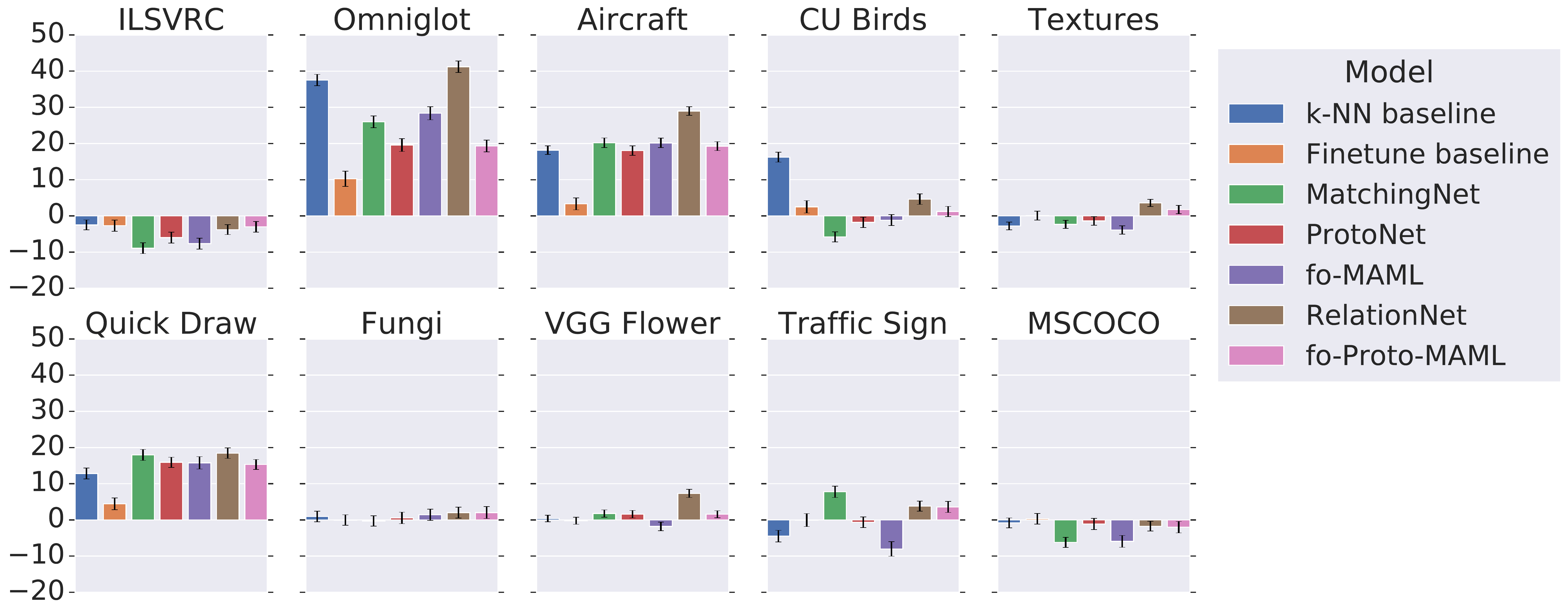}
  \caption{The performance difference on test datasets, when training on all datasets instead of ILSVRC only. A positive value indicates an improvement from all-dataset training.}
  \label{fig:trainsource_diff}
\end{figure*}

\paragraph{Effect of training on all datasets instead of ImageNet only}
It's interesting to examine whether training on (the training splits of) all
datasets leads to improved generalizaton compared to training on (the training
split of) ImageNet only. Specifically, while we might expect that training on
more data helps improve generalization, it is an empirical question whether
that still holds for heterogeneous data. We can examine this by comparing the
performance of each model between the top and bottom sets of results of
Table~\ref{tab:iclr2020_tables_without_ci}, corresponding to the two training
sources (ImageNet only and all datasets, respectively). For convenience,
Figure~\ref{fig:trainsource_diff} visualizes this difference in a barplot.
Notably, for Omniglot, Quick Draw and Aircraft we observe a substantial
increase across the board from training on all sources. This is reasonable for
datasets whose images are significantly different from ImageNet's: we indeed
expect to gain a large benefit from training on some images from (the training
classes of) these datasets.
Interestingly though, on the remainder of the test sources, we don't observe a
gain from all-dataset training. This result invites research into methods for
exploiting heterogeneous data for generalization to unseen classes of diverse
sources. Our experiments show that learning `naively' across the training
datasets (e.g., by picking the next dataset to use uniformly at random) does
not automatically lead to that desired benefit in most cases.

\begin{figure*}[h]
  \centering
	\subfloat[\label{fig:imagenet_way_analysis} Ways analysis]{\includegraphics[scale=0.19]{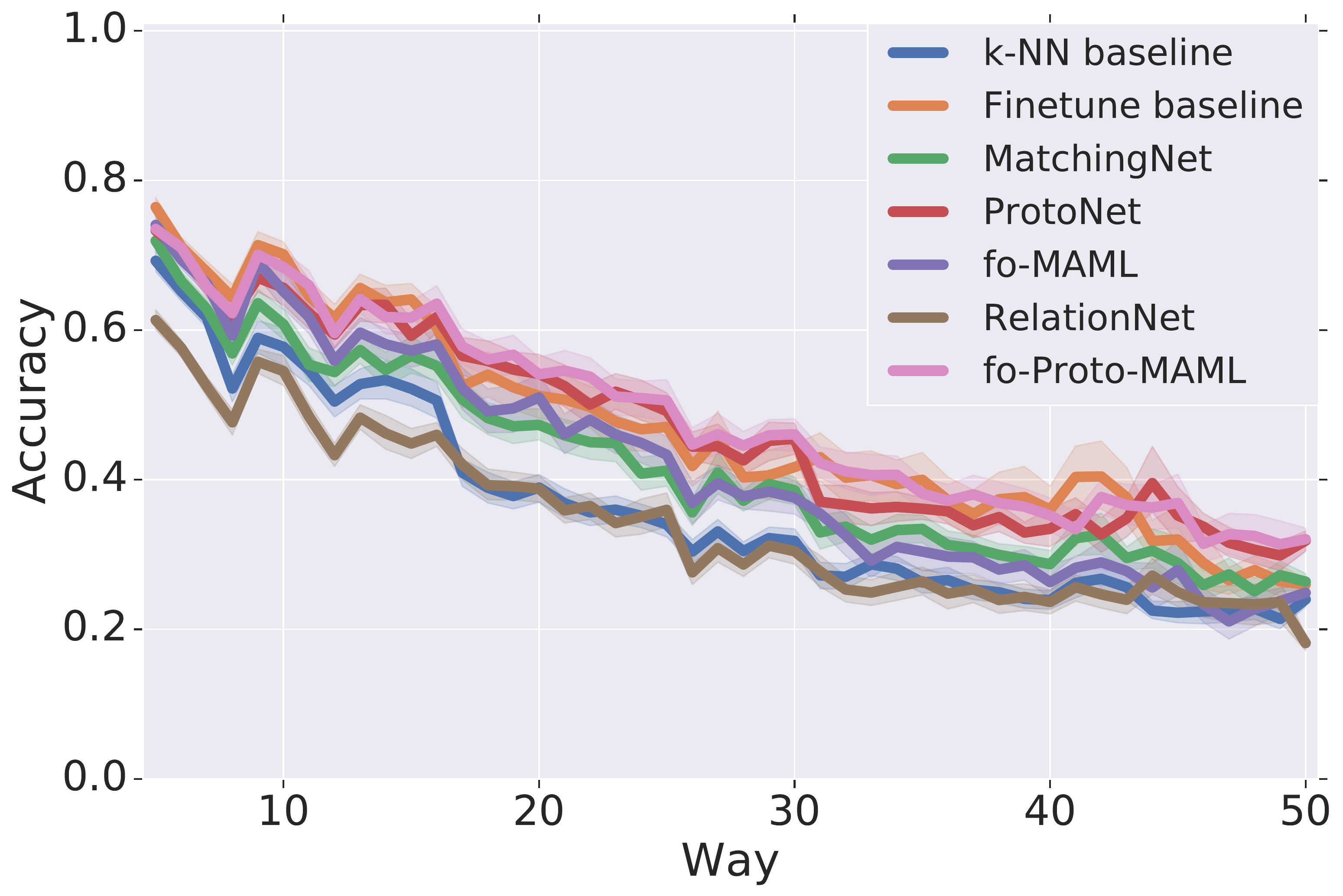}}\hspace{0.001cm}
	\subfloat[\label{fig:imagenet_shots_analysis} Shots analysis]{\includegraphics[scale=0.19]{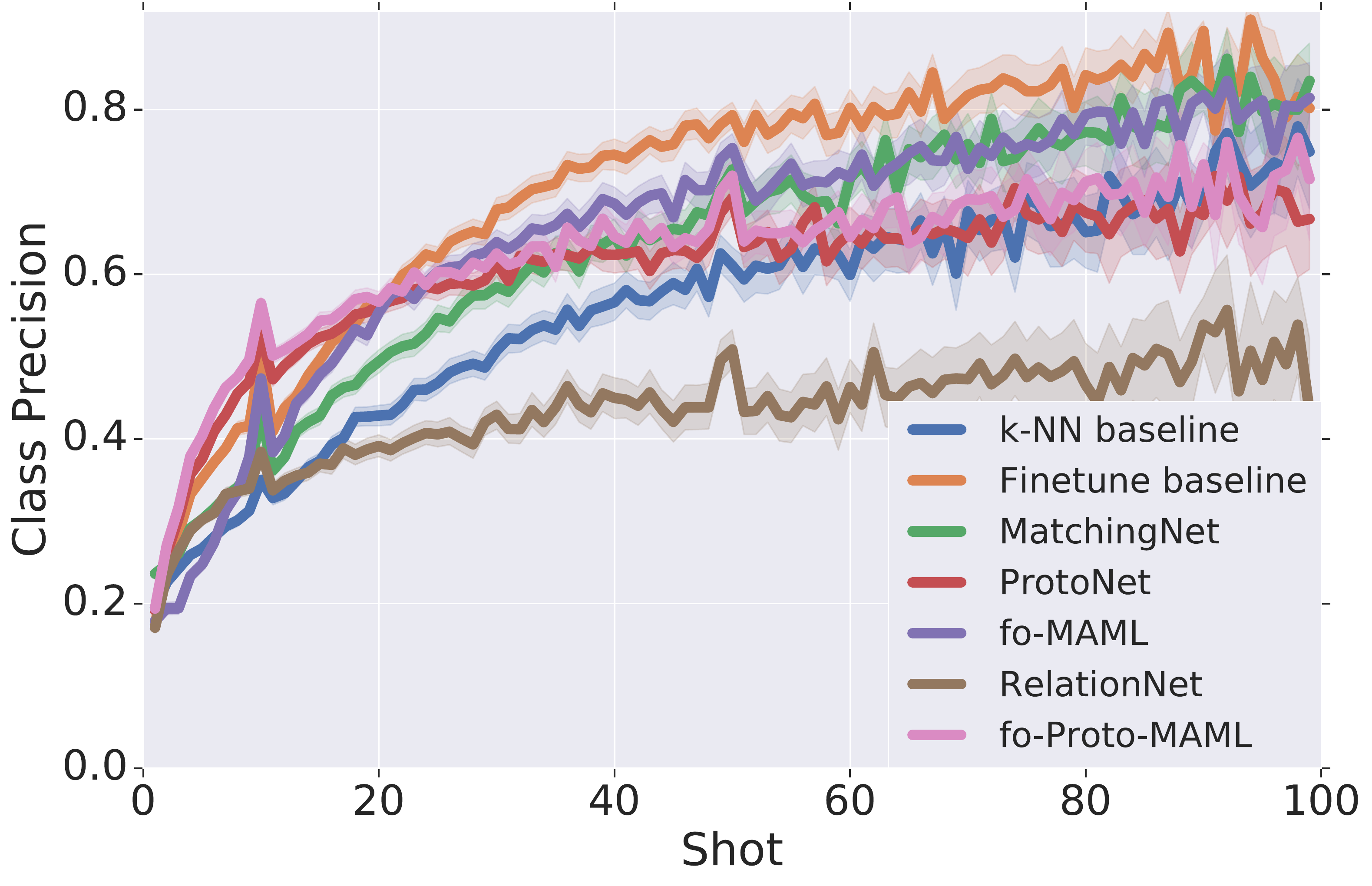}}\hspace{0.001cm}
	\caption{The effect of different ways and shots on test performance (w/ 95\% confidence intervals) when training on ImageNet.}
\end{figure*}
\paragraph{Ways and shots analysis} We further study the accuracy as a function of `ways' (Figure~\ref{fig:imagenet_way_analysis})
and the class precision as a function of `shots' (Figure~\ref{fig:imagenet_shots_analysis}). As expected, we found that the difficulty increases as the way increases, and performance degrades. More examples per class, on the other hand, indeed make it easier to correctly classify that class. Interestingly, though, not all models benefit at the same rate from more data: Prototypical Networks and
fo-Proto-MAML outshine other models in very-low-shot settings but saturate faster, whereas the Finetune baseline, Matching Networks, and fo-MAML improve at a higher rate when the shot increases. We draw the same conclusions when performing this analysis on all datasets, and include those plots in the Appendix. As discussed in the Appendix, we recommend including this analysis when reporting results on Meta-Dataset, aside from the main table. The rationale is that we're not only interested in performing well on average, but also in performing well under different specifications of test tasks.

\begin{figure*}[h]
  \centering
	\subfloat[\label{fig:effect_of_pretrain_imagenet} Effect of pre-training (ImageNet)]{\includegraphics[scale=0.145]{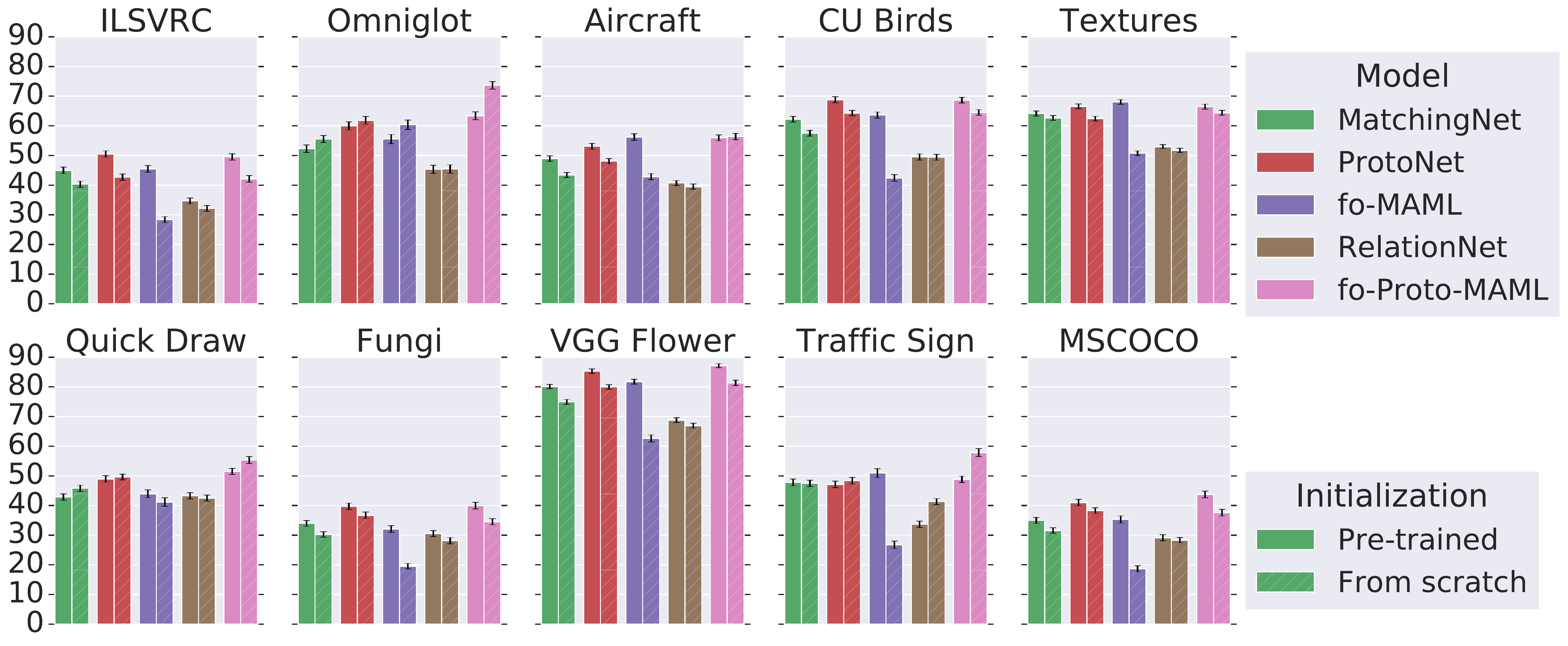}}\hspace{0.001cm}
	\subfloat[\label{fig:effect_of_pretrain_all_datasets} Effect of pre-training (All datasets)]{\includegraphics[scale=0.145]{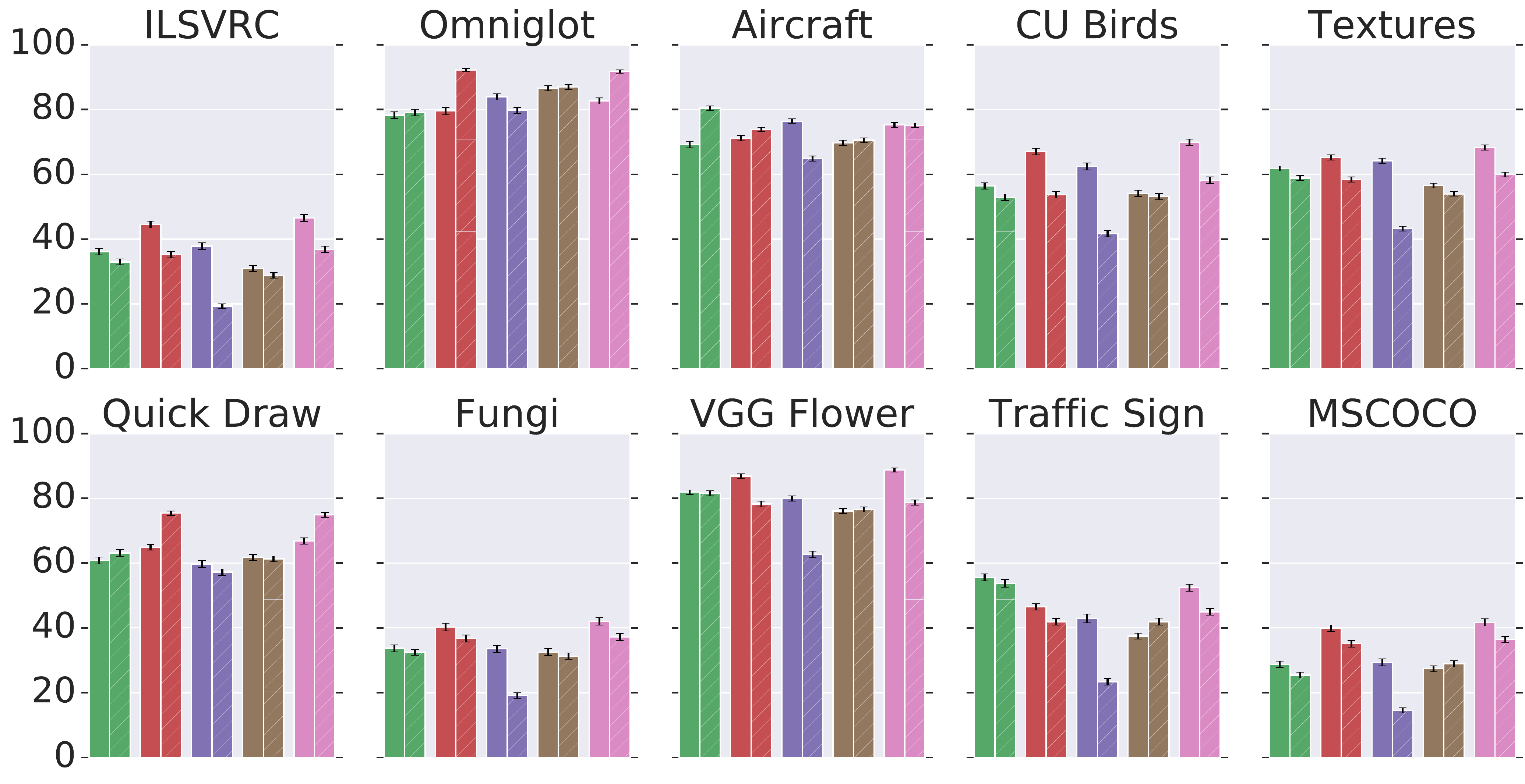}}\hspace{0.001cm}
	\subfloat[\label{fig:effect_of_meta_learn_imagenet} Effect of meta-learning (ImageNet)]{\includegraphics[scale=0.145]{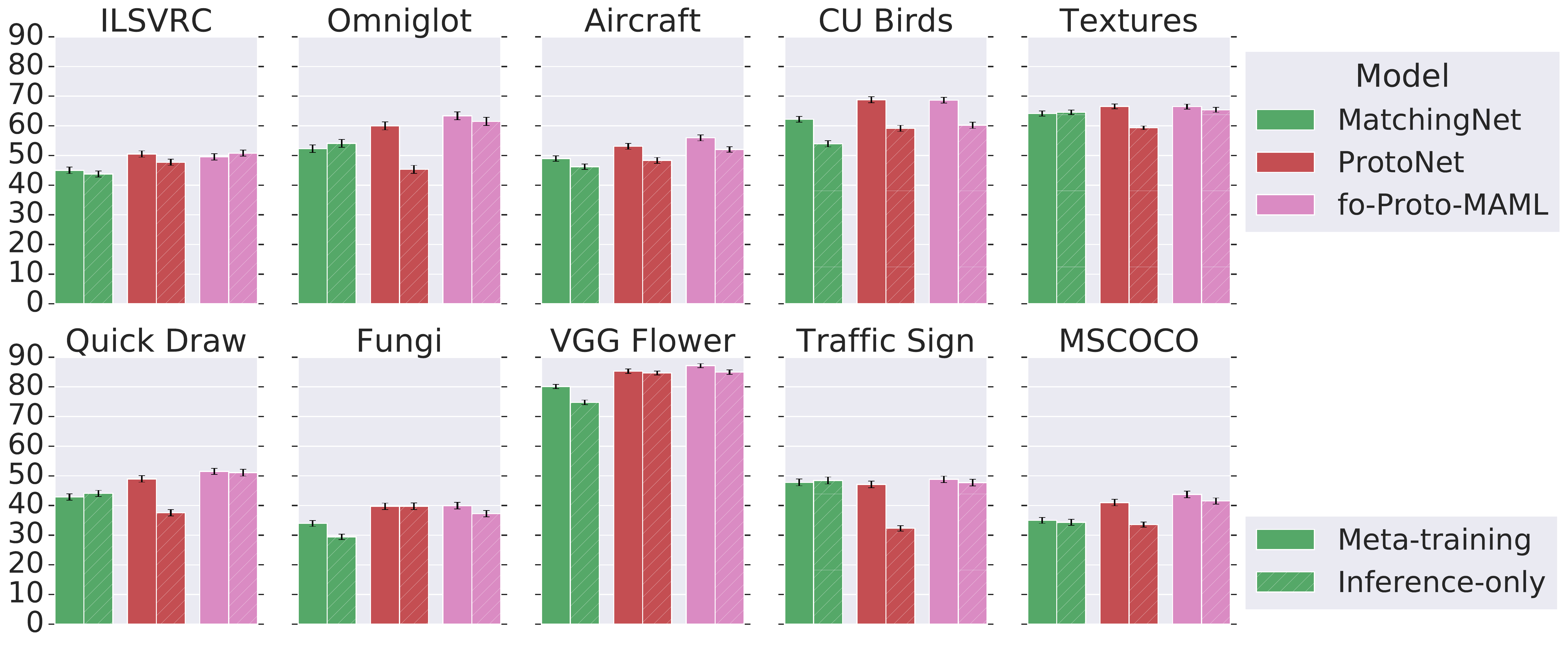}}\hspace{0.001cm}
	\subfloat[\label{fig:effect_of_meta_learn_all_datasets} Effect of meta-learning (All datasets)]{\includegraphics[scale=0.145]{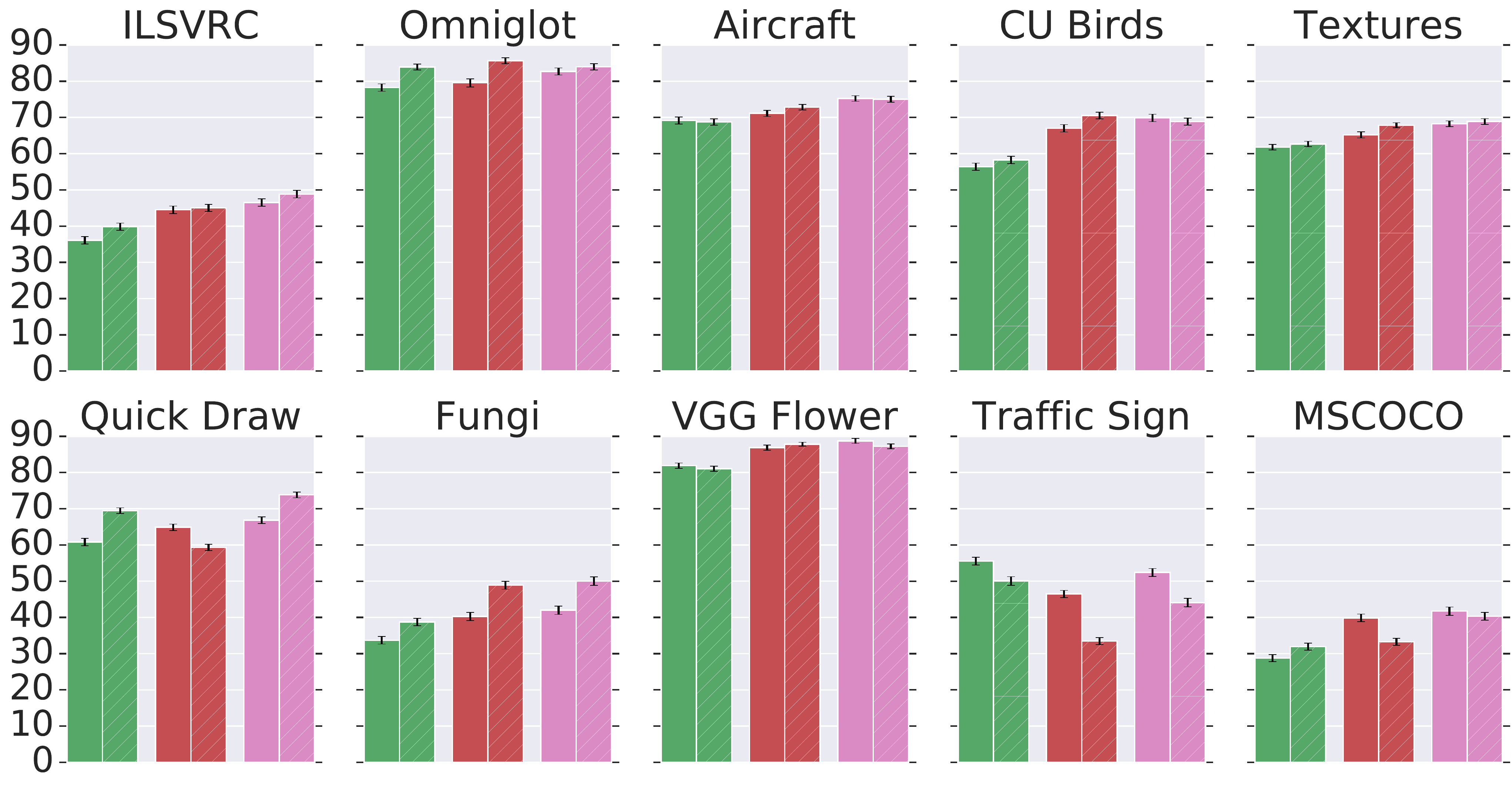}}\hspace{0.001cm}
	\caption{The effects of pre-training and meta-training (w/ 95\% confidence intervals). (ImageNet) or (All datasets) is the training source.}
\end{figure*}

\paragraph{Effect of pre-training}
In Figures~\ref{fig:effect_of_pretrain_imagenet}
and~\ref{fig:effect_of_pretrain_all_datasets}, we quantify how beneficial it is
to initialize the embedding network of meta-learners using the weights of the $k$-NN
baseline pre-trained on ImageNet, as opposed to starting their episodic training from scratch.
We find this procedure to often be beneficial, both for ImageNet-only training
and for training on all datasets. It seems that this ImageNet-influenced
initialization drives the meta-learner towards a solution which yields
increased performance on natural image test datasets, especially ILSVRC, Birds,
Fungi, Flowers and MSCOCO. Perhaps unsusprisingly, though, it underperforms on
significantly different datasets such as Omniglot and Quick Draw. These
findings show that, aside from the choice of the training data source(s) (e.g.,
ImageNet only or all datasets, as discussed above), the choice of the
initialization scheme can also influence to an important degree the final
solution and consequently the aptness of applying the resulting meta-learner to
different data sources at test time. Finally, an interesting observation is
that MAML seems to benefit the most from the pre-trained initialization, which
may speak to the difficulty of optimization associated with that model.

\paragraph{Effect of meta-training}
We propose to disentangle the inference algorithm of each meta-learner from the
fact that it is meta-learned, to assess the benefit of meta-learning on
\benchmark. To this end, we propose a new set of baselines: `Prototypical
Networks Inference', `Matching Networks Inference', and `fo-Proto-MAML
Inference', that are trained non-episodically but evaluated episodically
(for validation and testing) using the inference algorithm of the respective meta-learner. This
is possible for these meta-learners as they don't have any additional
parameters aside from the embedding function that explicitly need to be learned
episodically (as opposed to the relation module of Relation Networks, for example). We
compare each Inference-only method to its corresponding meta-learner in
Figures~\ref{fig:effect_of_meta_learn_imagenet}
and~\ref{fig:effect_of_meta_learn_all_datasets}.
We find that these baselines are strong: when training on ImageNet only, we can
usually observe a small benefit from meta-learning the embedding weights but
this benefit often disappears when training on all datasets, in which case
meta-learning sometimes actually hurts. We find this result very interesting
and we believe it emphasizes the need for research on how to meta-learn across
multiple diverse sources, an important challenge that \benchmark puts forth.

\paragraph{Fine-grainedness analysis} We use ILVRC-2012 to investigate the hypothesis that finer-grained tasks are harder than coarse-grained ones. Our findings suggest that while the test sub-graph is not rich enough to exhibit any trend, the performance on the train sub-graph does seem to agree with this hypothesis. We include the experimental setup and results for this analysis in the Appendix.

\section{Conclusion}

We have introduced a new large-scale, diverse, and realistic environment for few-shot classification. We believe that our exploration of various models on \benchmark has uncovered
interesting directions for future work pertaining to meta-learning across heterogeneous data: it remains unclear what is the best strategy for creating training episodes, the most appropriate validation creation and the most appropriate initialization. Current models don't always improve when trained on multiple sources and meta-learning is not always beneficial across datasets. Current models are also not robust to the amount of data in test episodes, each excelling in a different part of the spectrum. We believe that addressing these shortcomings consitutes an important research goal moving forward.

\ificlrfinal
\subsubsection*{Author Contributions}

Eleni, Hugo, and Kevin came up with the benchmark idea and requirements.  Eleni
developed the core of the project, and worked on the experiment design and
management with Tyler and Kevin, as well as experiment analysis. Carles, Ross,
Kelvin, Pascal, Vincent, and Tyler helped extend the benchmark by adding
datasets. Eleni, Vincent, and Utku contributed the Prototypical Networks,
Matching Networks, and Relation Networks implementations, respectively.
Tyler implemented baselines, MAML (with
Kevin) and Proto-MAML models, and updated the backbones to support them.
Writing was mostly led by Eleni, with contributions by Hugo, Vincent, and Kevin
and help from Tyler and Pascal for visualizations. Pascal and Pierre-Antoine
worked on code organization, efficiency, and open-sourcing, Pascal and Vincent
optimized the efficiency of the data input pipeline. Pierre-Antoine supervised
the code development process and reviewed most of the changes, Hugo and Kevin
supervised the overall direction of the research.

\subsubsection*{Acknowledgments}

We would like to thank Chelsea Finn for fruitful discussions and
advice on tuning fo-MAML and ensuring the correctness of implementation, as
well as Zack Nado and Dan Moldovan for the initial dataset code that was
adapted, and Cristina Vasconcelos for spotting an issue in the ranking of models.
Finally, we'd like to thank John Bronskill for suggesting that we
experiment with a larger inner-loop learning rate for MAML which indeed
significantly improved our fo-MAML results on \benchmark.
\fi

\newpage

\bibliography{references}

\begin{thebibliography}{35}
\providecommand{\natexlab}[1]{#1}
\providecommand{\url}[1]{\texttt{#1}}
\expandafter\ifx\csname urlstyle\endcsname\relax
  \providecommand{\doi}[1]{doi: #1}\else
  \providecommand{\doi}{doi: \begingroup \urlstyle{rm}\Url}\fi

\bibitem[Antoniou et~al.(2019)Antoniou, Edwards, and Storkey]{AntoniouA2019}
Antreas Antoniou, Harrison Edwards, and Amos Storkey.
\newblock How to train your {MAML}.
\newblock In \emph{Proceedings of the International Conference on Learning
  Representations}, 2019.

\bibitem[Bertinetto et~al.(2019)Bertinetto, Henriques, Torr, and
  Vedaldi]{bertinetto2018metalearning}
Luca Bertinetto, Joao~F. Henriques, Philip Torr, and Andrea Vedaldi.
\newblock Meta-learning with differentiable closed-form solvers.
\newblock In \emph{Proceedings of the International Conference on Learning
  Representations}, 2019.

\bibitem[Chen et~al.(2019)Chen, Liu, Kira, Wang, and Huang]{chen2019closer}
Wei-Yu Chen, Yen-Cheng Liu, Zsolt Kira, Yu-Chiang~Frank Wang, and Jia-Bin
  Huang.
\newblock A closer look at few-shot classification.
\newblock In \emph{Proceedings of the International Conference on Learning
  Representations}, 2019.

\bibitem[Cimpoi et~al.(2014)Cimpoi, Maji, Kokkinos, Mohamed, and
  Vedaldi]{cimpoi14describing}
M.~Cimpoi, S.~Maji, I.~Kokkinos, S.~Mohamed, and A.~Vedaldi.
\newblock Describing textures in the wild.
\newblock In \emph{{IEEE} Conference on Computer Vision and Pattern
  Recognition}, 2014.

\bibitem[Dhillon et~al.(2019)Dhillon, Chaudhari, Ravichandran, and
  Soatto]{dhillon2019baseline}
Guneet~S. Dhillon, Pratik Chaudhari, Avinash Ravichandran, and Stefano Soatto.
\newblock A baseline for few-shot image classification.
\newblock \emph{arXiv}, abs/1909.02729, 2019.

\bibitem[Finn et~al.(2017)Finn, Abbeel, and Levine]{finn2017model}
Chelsea Finn, Pieter Abbeel, and Sergey Levine.
\newblock Model-agnostic meta-learning for fast adaptation of deep networks.
\newblock In \emph{Proceedings of the International Conference of Machine
  Learning}, 2017.

\bibitem[Gidaris \& Komodakis(2018)Gidaris and Komodakis]{GidarisS2018}
Spyros Gidaris and Nikos Komodakis.
\newblock Dynamic few-shot visual learning without forgetting.
\newblock In \emph{{IEEE} Conference on Computer Vision and Pattern
  Recognition}, 2018.

\bibitem[Hariharan \& Girshick(2017)Hariharan and Girshick]{hariharan2017low}
Bharath Hariharan and Ross Girshick.
\newblock Low-shot visual recognition by shrinking and hallucinating features.
\newblock In \emph{Proceedings of the IEEE International Conference on Computer
  Vision}, pp.\  3018--3027, 2017.

\bibitem[Houben et~al.(2013)Houben, Stallkamp, Salmen, Schlipsing, and
  Igel]{Houben-IJCNN-2013}
Sebastian Houben, Johannes Stallkamp, Jan Salmen, Marc Schlipsing, and
  Christian Igel.
\newblock Detection of traffic signs in real-world images: The {German}
  {Traffic} {Sign} {Detection} {Benchmark}.
\newblock In \emph{International Joint Conference on Neural Networks}, 2013.

\bibitem[Jongejan et~al.(2016)Jongejan, Rowley, Kawashima, Kim, and
  Fox-Gieg]{jongejan2016quick}
Jonas Jongejan, Henry Rowley, Takashi Kawashima, Jongmin Kim, and Nick
  Fox-Gieg.
\newblock The {Quick}, {Draw}! -- {A.I.} experiment.
\newblock \url{quickdraw.withgoogle.com}, 2016.

\bibitem[Krizhevsky et~al.(2009)]{krizhevsky2009learning}
Alex Krizhevsky et~al.
\newblock Learning multiple layers of features from tiny images.
\newblock Technical report, University of Toronto, 2009.

\bibitem[Lake et~al.(2015)Lake, Salakhutdinov, and Tenenbaum]{lake2015human}
Brenden~M Lake, Ruslan Salakhutdinov, and Joshua~B Tenenbaum.
\newblock Human-level concept learning through probabilistic program induction.
\newblock \emph{Science}, 350\penalty0 (6266):\penalty0 1332--1338, 2015.

\bibitem[Lin et~al.(2014)Lin, Maire, Belongie, Hays, Perona, Ramanan,
  Doll{\'a}r, and Zitnick]{lin2014microsoft}
Tsung-Yi Lin, Michael Maire, Serge Belongie, James Hays, Pietro Perona, Deva
  Ramanan, Piotr Doll{\'a}r, and C~Lawrence Zitnick.
\newblock Microsoft {COCO}: Common objects in context.
\newblock In \emph{European Conference on Computer Vision}, pp.\  740--755,
  2014.

\bibitem[Maji et~al.(2013)Maji, Rahtu, Kannala, Blaschko, and
  Vedaldi]{maji13finegrained}
Subhransu Maji, Esa Rahtu, Juho Kannala, Matthew Blaschko, and Andrea Vedaldi.
\newblock Fine-grained visual classification of aircraft.
\newblock \emph{arXiv}, abs/1306.5151, 2013.

\bibitem[Mishra et~al.(2018)Mishra, Rohaninejad, Chen, and Abbeel]{mishra2018a}
Nikhil Mishra, Mostafa Rohaninejad, Xi~Chen, and Pieter Abbeel.
\newblock A simple neural attentive meta-learner.
\newblock In \emph{Proceedings of the International Conference on Learning
  Representations}, 2018.

\bibitem[Munkhdalai \& Yu(2017)Munkhdalai and Yu]{munkhdalai17a}
Tsendsuren Munkhdalai and Hong Yu.
\newblock Meta networks.
\newblock In \emph{Proceedings of the International Conference on Machine
  Learning}, pp.\  2554--2563, 2017.

\bibitem[Nilsback \& Zisserman(2008)Nilsback and Zisserman]{Nilsback08}
M-E. Nilsback and A.~Zisserman.
\newblock Automated flower classification over a large number of classes.
\newblock In \emph{Proceedings of the Indian Conference on Computer Vision,
  Graphics and Image Processing}, 2008.

\bibitem[Oreshkin et~al.(2018)Oreshkin, Rodriguez, and Lacoste]{OreshkinB2018}
Boris~N.\ Oreshkin, Pau Rodriguez, and Alexandre Lacoste.
\newblock {TADAM}: Task dependent adaptive metric for improved few-shot
  learning.
\newblock In \emph{Advances in Neural Information Processing Systems}, pp.\
  719--729, 2018.

\bibitem[Qi et~al.(2018)Qi, Brown, and Lowe]{qi2018low}
Hang Qi, Matthew Brown, and David~G Lowe.
\newblock Low-shot learning with imprinted weights.
\newblock In \emph{Proceedings of the IEEE Conference on Computer Vision and
  Pattern Recognition}, pp.\  5822--5830, 2018.

\bibitem[Ravi \& Larochelle(2017)Ravi and Larochelle]{RaviS2017}
Sachin Ravi and Hugo Larochelle.
\newblock Optimization as a model for few-shot learning.
\newblock In \emph{Proceedings of the International Conference on Learning
  Representations}, 2017.

\bibitem[Ren et~al.(2018)Ren, Triantafillou, Ravi, Snell, Swersky, Tenenbaum,
  Larochelle, and Zemel]{ren2018meta}
Mengye Ren, Eleni Triantafillou, Sachin Ravi, Jake Snell, Kevin Swersky,
  Joshua~B Tenenbaum, Hugo Larochelle, and Richard~S Zemel.
\newblock Meta-learning for semi-supervised few-shot classification.
\newblock In \emph{Proceedings of the International Conference on Learning
  Representations}, 2018.

\bibitem[Russakovsky et~al.(2015)Russakovsky, Deng, Su, Krause, Satheesh, Ma,
  Huang, Karpathy, Khosla, Bernstein, Berg, and
  Fei-Fei]{russakovsky2015imagenet}
Olga Russakovsky, Jia Deng, Hao Su, Jonathan Krause, Sanjeev Satheesh, Sean Ma,
  Zhiheng Huang, Andrej Karpathy, Aditya Khosla, Michael Bernstein, Alexander~C
  Berg, and Li~Fei-Fei.
\newblock Imagenet large scale visual recognition challenge.
\newblock \emph{International Journal of Computer Vision}, 115\penalty0
  (3):\penalty0 211--252, 2015.

\bibitem[Rusu et~al.(2019)Rusu, Rao, Sygnowski, Vinyals, Pascanu, Osindero, and
  Hadsell]{rusu2018metalearning}
Andrei~A. Rusu, Dushyant Rao, Jakub Sygnowski, Oriol Vinyals, Razvan Pascanu,
  Simon Osindero, and Raia Hadsell.
\newblock Meta-learning with latent embedding optimization.
\newblock In \emph{Proceedings of the International Conference on Learning
  Representations}, 2019.

\bibitem[Salimans \& Kingma(2016)Salimans and Kingma]{salimans2016weight}
Tim Salimans and Durk~P Kingma.
\newblock Weight normalization: A simple reparameterization to accelerate
  training of deep neural networks.
\newblock In \emph{Advances in Neural Information Processing Systems}, pp.\
  901--909, 2016.

\bibitem[Santoro et~al.(2016)Santoro, Bartunov, Botvinick, Wierstra, and
  Lillicrap]{pmlr-v48-santoro16}
Adam Santoro, Sergey Bartunov, Matthew Botvinick, Daan Wierstra, and Timothy
  Lillicrap.
\newblock Meta-learning with memory-augmented neural networks.
\newblock In \emph{Proceedings of the International Conference on Machine
  Learning}, pp.\  1842--1850, 2016.

\bibitem[Satorras \& Estrach(2018)Satorras and Estrach]{garcia2018fewshot}
Victor~Garcia Satorras and Joan~Bruna Estrach.
\newblock Few-shot learning with graph neural networks.
\newblock In \emph{Proceedings of the International Conference on Learning
  Representations}, 2018.

\bibitem[Schroeder \& Cui(2018)Schroeder and Cui]{fungi}
Brigit Schroeder and Yin Cui.
\newblock {FGVCx} fungi classification challenge 2018.
\newblock \url{github.com/visipedia/fgvcx_fungi_comp}, 2018.

\bibitem[Snell et~al.(2017)Snell, Swersky, and Zemel]{snell2017prototypical}
Jake Snell, Kevin Swersky, and Richard Zemel.
\newblock Prototypical networks for few-shot learning.
\newblock In \emph{Advances in Neural Information Processing Systems}, pp.\
  4077--4087, 2017.

\bibitem[Sung et~al.(2018)Sung, Yang, Zhang, Xiang, Torr, and
  Hospedales]{sung2018learning}
Flood Sung, Yongxin Yang, Li~Zhang, Tao Xiang, Philip~HS Torr, and Timothy~M
  Hospedales.
\newblock Learning to compare: Relation network for few-shot learning.
\newblock In \emph{Proceedings of the IEEE Conference on Computer Vision and
  Pattern Recognition}, pp.\  1199--1208, 2018.

\bibitem[Vinyals et~al.(2016)Vinyals, Blundell, Lillicrap, and
  Wierstra]{vinyals2016matching}
Oriol Vinyals, Charles Blundell, Tim Lillicrap, and Daan Wierstra.
\newblock Matching networks for one shot learning.
\newblock In \emph{Advances in Neural Information Processing Systems}, pp.\
  3630--3638, 2016.

\bibitem[Wah et~al.(2011)Wah, Branson, Welinder, Perona, and
  Belongie]{WahCUB_200_2011}
C.~Wah, S.~Branson, P.~Welinder, P.~Perona, and S.~Belongie.
\newblock {The Caltech-UCSD Birds-200-2011 Dataset}.
\newblock Technical Report CNS-TR-2011-001, California Institute of Technology,
  2011.

\bibitem[Wang \& Hebert(2016)Wang and Hebert]{wang2016learning}
Yu-Xiong Wang and Martial Hebert.
\newblock Learning to learn: Model regression networks for easy small sample
  learning.
\newblock In \emph{European Conference on Computer Vision}, pp.\  616--634.
  Springer, 2016.

\bibitem[Wang et~al.(2017)Wang, Ramanan, and Hebert]{wang2017learning}
Yu-Xiong Wang, Deva Ramanan, and Martial Hebert.
\newblock Learning to model the tail.
\newblock In \emph{Advances in Neural Information Processing Systems}, pp.\
  7029--7039, 2017.

\bibitem[Wang et~al.(2018)Wang, Girshick, Hebert, and Hariharan]{wang2018low}
Yu-Xiong Wang, Ross Girshick, Martial Hebert, and Bharath Hariharan.
\newblock Low-shot learning from imaginary data.
\newblock In \emph{Proceedings of the IEEE Conference on Computer Vision and
  Pattern Recognition}, pp.\  7278--7286, 2018.

\bibitem[Yoon et~al.(2018)Yoon, Kim, Dia, Kim, Bengio, and Ahn]{NIPS2018_7963}
Jaesik Yoon, Taesup Kim, Ousmane Dia, Sungwoong Kim, Yoshua Bengio, and Sungjin
  Ahn.
\newblock Bayesian model-agnostic meta-learning.
\newblock In \emph{Advances in Neural Information Processing Systems}, 2018.

\end{thebibliography}
\bibliographystyle{iclr2020_conference}

\newpage

\appendix
\section*{Appendix}

\subsection{Recommendation for reporting results on \benchmark}
We recommend that future work on \benchmark reports two sets of results:
\begin{enumerate}
	\item The main tables storing the average (over 600 test episodes) accuracy of each method on each dataset, after it has been trained on ImageNet only and on All datasets, where the evaluation metric is the average rank. This corresponds to Table~\ref{tab:iclr2020_tables_without_ci} in our case (or the more complete version in Table~\ref{tab:iclr2020_tables} in the Appendix).
	\item The plots that measure robustness in variations of shots and ways. In our case these are Figures~\ref{fig:imagenet_shots_analysis} and~\ref{fig:imagenet_way_analysis} in the main text for ImageNet-only training, and Figures~\ref{fig:all_shot_vs_precision} and~\ref{fig:all_way_vs_accuracy} in the Appendix for the case of training on all datasets.
\end{enumerate}
We propose to use both of these aspects to evaluate performance on \benchmark: it is not only desirable to perform well on average, but also to perform well under different specifications of test tasks, as it is not realistic in general to assume that we will know in advance what setup (number of ways and shots) will be encountered at test time. 
Our final source code will include scripts for generating these plots and for automatically computing ranks given a table to help standardize the procedure for reporting results.

\subsection{Details of \benchmark's Sampling Algorithm}
We now provide a complete description of certain steps that were explained on a higher level in the main paper.
\subsubsection*{Step 1: Sampling the episode's class set}
\paragraph{ImageNet class sampling} The procedure we use for sampling classes for an ImageNet
episode is the following. First, we sample a node uniformly at random
from the set of `eligible' nodes of the DAG structure corresponding to the
specified split (train, validation or test). An internal node is `eligible' for
this selection if it spans at least 5 leaves, but no more than 392 leaves.  The
number 392 was chosen because it is the smallest number so that, collectively,
all eligible internal nodes span all leaves in the DAG. Once an eligible node is 
selected, some of the leaves that it spans will constitute the classes of the episode. Specifically, if the number of those leaves is no greater than 50, we use all of them. Otherwise, we randomly choose 50 of them.

This procedure enables the creation of tasks of varying degrees of
fine-grainedness. For instance, if the sampled internal node has a small
height, the leaf classes that it spans will represent semantically-related
concepts, thus posing a fine-grained classification task. As the height of the
sampled node increases, we `zoom out' to consider a broader scope from which we
sample classes and the resulting episodes are more coarse-grained.

\subsubsection*{Step 2: Sampling the episode's examples}
\paragraph{a) Computing the query set size} The query set is class-balanced, reflecting the fact that we care equally to
perform well on all classes of an episode. The number of query images
\textit{per class} is computed as:
\begin{equation*}
  q = \min \left\{10, \left(\min_{c \in \mathcal{C}} \left\lfloor0.5 * |Im(c)|\right\rfloor\right)\right\}
\end{equation*}

where $\mathcal{C}$ is the set of selected classes and $Im(c)$ denotes the set
of images belonging to class $c$. The min over classes ensures that each class
has at least $q$ images to add to the query set, thus allowing it to be
class-balanced. The $0.5$ multiplier ensures that enough images of each class
will be available to add to the support set, and the minimum with $10$ prevents
the query set from being too large.

\paragraph{b) Computing the support set size} We compute the total support set size as:
\begin{equation*}
  |\mathcal{S}| = \min \left\{500, \displaystyle \sum_{c \in \mathcal{C}}\left\lceil \beta  \min \{ 100, |Im(c)| - q\} \right\rceil \right\}
\end{equation*}
where $\beta$ is a scalar sampled uniformly from interval $(0, 1]$.
Intuitively, each class on average contributes either all its remaining
examples (after placing $q$ of them in the query set) if there are less than
$100$ or $100$ otherwise, to avoid having too large support sets. The
multiplication with $\beta$ enables the potential generation of smaller support
sets even when multiple images are available, since we are also interested in
examining the very-low-shot end of the spectrum. The `ceiling' operation
ensures that each selected class will have at least one image in the support
set. Finally, we cap the total support set size to $500$.

\paragraph{c) Computing the shot of each class} We are now ready to compute the `shot' of each class. Specifically, the
proportion of the support set that will be devoted to class $c$ is computed as:
\begin{equation*}
  R_c = \displaystyle\frac{\exp(\alpha_c) |Im(c)|}{\displaystyle\sum_{c' \in \mathcal{C}} \exp(\alpha_c') |Im(c')|}
\end{equation*}
where $\alpha_c$ is sampled uniformly from the interval $[\log(0.5), \log(2))$.
Intuitively, the un-normalized proportion of the support set that will be
occupied by class $c$ is a noisy version of the total number of images of that
class in the dataset $Im(c)$. This design choice is made in the hopes of
obtaining realistic class ratios, under the hypothesis that the dataset class
statistics are a reasonable approximation of the real-world statistics of
appearances of the corresponding classes. The shot of a class $c$ is then set
to:
\begin{equation*}
  k_c = \min \left\{\left\lfloor R_c * \left(|\mathcal{S}| - |\mathcal{C}|\right) \right\rfloor + 1, |Im(c)| - q \right\}
\end{equation*}
which ensures that at least one example is selected for each class, with
additional examples selected proportionally to $R_c$, if enough are available.

\subsection{Datasets}
\label{sec:appendix_datasets}

\benchmark is formed of data originating from 10 different image datasets. A
complete list of the datasets we use is the following.

\begin{figure*}[htp]
  \centering
  \subfloat[\label{fig:dataset_samples_imagenet} ImageNet]{\includegraphics[width=0.18\textwidth]{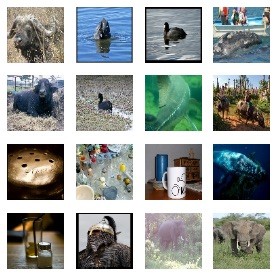}}\hspace{0.1cm}
  \subfloat[\label{fig:dataset_samples_omniglot} Omniglot]{\includegraphics[width=0.18\textwidth]{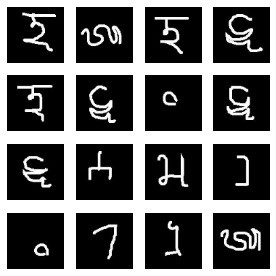}}\hspace{0.1cm}
  \subfloat[\label{fig:dataset_samples_aircraft} Aircraft]{\includegraphics[width=0.18\textwidth]{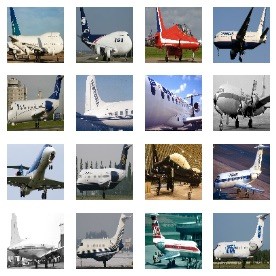}}\hspace{0.1cm}
  \subfloat[\label{fig:dataset_samples_birds} Birds]{\includegraphics[width=0.18\textwidth]{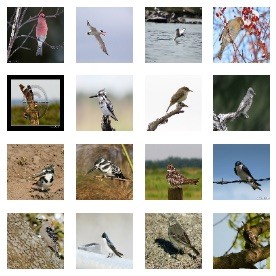}}\hspace{0.1cm}
  \subfloat[\label{fig:dataset_samples_dtd} DTD]{\includegraphics[width=0.18\textwidth]{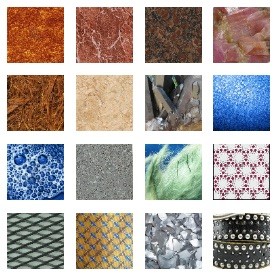}}\\
  \subfloat[\label{fig:dataset_samples_quickdraw} Quick Draw]{\includegraphics[width=0.18\textwidth]{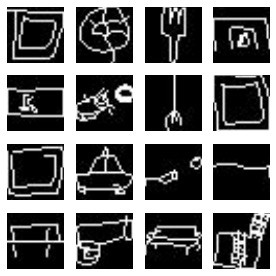}}\hspace{0.1cm}
  \subfloat[\label{fig:dataset_samples_fungi} Fungi]{\includegraphics[width=0.18\textwidth]{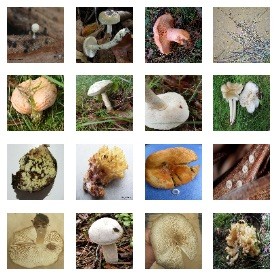}}\hspace{0.1cm}
  \subfloat[\label{fig:dataset_samples_flower} VGG Flower]{\includegraphics[width=0.18\textwidth]{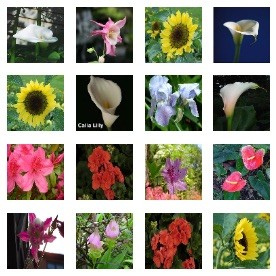}}\hspace{0.1cm}
  \subfloat[\label{fig:dataset_samples_traffic} Traffic Signs]{\includegraphics[width=0.18\textwidth]{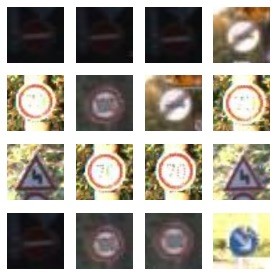}}\hspace{0.1cm}
  \subfloat[\label{fig:dataset_samples_mscoco} MSCOCO]{\includegraphics[width=0.18\textwidth]{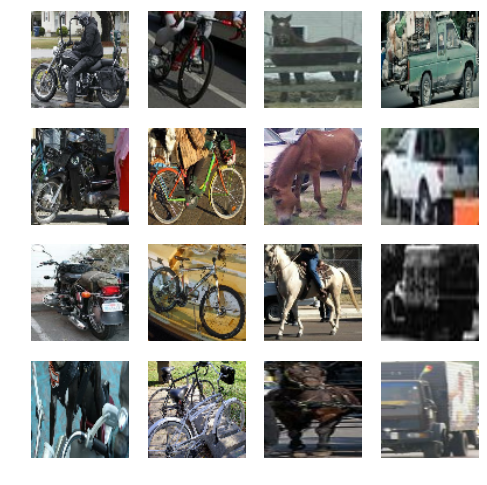}}
  \caption{\label{fig:dataset_samples} Training examples taken from the various datasets forming \benchmark.}
\end{figure*}

\paragraph{ILSVRC-2012 (ImageNet, \citealp{russakovsky2015imagenet})} A dataset of
natural images from 1000 categories (Figure~\ref{fig:dataset_samples_imagenet}).
We removed some images that were duplicates of images in another dataset in
\benchmark (43 images that were also part of Birds) or other standard datasets
of interest (92 from Caltech-101 and 286 from Caltech-256). The complete list
of duplicates is part of the source code release.

\paragraph{Omniglot \citep{lake2015human}} A dataset of images of 1623
handwritten characters from 50 different alphabets, with 20 examples per class
(Figure~\ref{fig:dataset_samples_omniglot}). While recently \cite{vinyals2016matching}
proposed a new split for this dataset, we instead make use of the original
intended split \cite{lake2015human} which is more challenging since the split
is on the level of alphabets (30 training alphabets and 20 evaluation
alphabets), not characters from those alphabets, therefore posing a more
challenging generalization problem. Out of the 30 training alphabets, we hold
out the 5 smallest ones (i.e., with the least number of character classes) to
form our validation set, and use the remaining 25 for training.

\paragraph{Aircraft \citep{maji13finegrained}} A dataset of images of aircrafts
spanning 102 model variants, with 100 images per class
(Figure~\ref{fig:dataset_samples_aircraft}).
The images are cropped according to the providing bounding boxes, in order not
to include other aircrafts, or the copyright text at the bottom of images.

\paragraph{CUB-200-2011 (Birds, \citealp{WahCUB_200_2011})} A dataset for
fine-grained classification of 200 different bird species
(Figure~\ref{fig:dataset_samples_birds}).
We did not use the provided bounding boxes to crop the images, instead the full
images are used, which provides a harder challenge.

\paragraph{Describable Textures (DTD, \citealp{cimpoi14describing})} A texture
database, consisting of 5640 images, organized according to a list of 47 terms
(categories) inspired from human perception
(Figure~\ref{fig:dataset_samples_dtd}).

\paragraph{Quick Draw \citep{jongejan2016quick}} A dataset of 50 million
black-and-white drawings across 345 categories, contributed by players of the
game Quick, Draw!  (Figure~\ref{fig:dataset_samples_quickdraw}).

\paragraph{Fungi \citep{fungi}} A large dataset of approximately 100K images of
nearly 1,500 wild mushrooms species (Figure~\ref{fig:dataset_samples_fungi}).

\paragraph{VGG Flower \citep{Nilsback08}} A dataset of natural images of 102
flower categories. The flowers chosen to be ones commonly occurring in the
United Kingdom. Each class consists of between 40 and 258 images
(Figure~\ref{fig:dataset_samples_flower}).

\paragraph{Traffic Signs \citep{Houben-IJCNN-2013}} A dataset of 50,000 images
of German road signs in 43 classes (Figure~\ref{fig:dataset_samples_traffic}).

\paragraph{MSCOCO \cite{lin2014microsoft}} A dataset of images collected from
Flickr with 1.5 million object instances belonging to 80 classes labelled and
localized using bounding boxes. We choose the train2017 split and create images
crops from original images using each object instance's groundtruth bounding
box (Figure~\ref{fig:dataset_samples_mscoco}).

\subsection{Hyperparameters}
We used three architectures: a commonly-used four-layer convolutional network,
an 18-layer residual network and a wide residual network. While some of the
baseline models performed best with the latter, we noticed that the
meta-learners preferred the resnet-18 backbone and rarely the
four-layer-convnet.
For Relation Networks only, we also allow the option to use another
architecture, aside from the aforementioned three, inspired by the
four-layer-convnet used in the Relation Networks paper
\citep{sung2018learning}.
The main difference is that they used the usual max-pooling operation only in
the first two layers, omitting it in the last two, yielding activations of
larger spatial dimensions. In our case, we found that these increased spatial
dimensions did not fit in memory, so as a compromise we used max-pooling on the
first 3 out of the 4 layer of the convnet.

For fo-MAML and fo-Proto-MAML, we tuned the inner-loop learning rate, the
number of inner loop steps, and the number of additional such steps to be
performed in evaluation (i.e., validation or test) episodes.

For the baselines, we tuned whether the cosine classifier of Baseline++ will be
used, as opposed to a standard forward pass through a linear classification
layer. Also, since \citet{chen2019closer} added weight normalization
\citep{salimans2016weight} to their implementation of the cosine classifier
layer, we also implemented this and created a hyperparameter choice for whether
or not it is enabled. This hyperparameter is independent from the one that
decides if the cosine classifier is used. Both are applicable to the $k$-NN
Basline (for its all-way training classification task) and to the Finetune
Baseline (both for its all-way training classification and for its
within-episode classification at validation and test times). For the Finetune
Baseline, we tuned a binary hyperparameter deciding if gradient descent or ADAM
is used for the within-task optimization. We also tuned the decision of whether
all embedding layers are finetuned or, alternatively, the embedding is held
fixed and only the final classifier on top of it is optimized. Finally, we tuned
the number of finetuning steps that will be carried out.

We also tried two different image resolutions: the commonly-used 84x84 and 126x126. Finally, we tuned the learning rate schedule and weight decay and we used ADAM to train all of our models. All other details, dataset splits and the complete set of best hyperparameters discovered for each model are included in the source code.

\subsection{Complete Main Results and Rank Computation}

\paragraph{Rank computation}
We rank models by decreasing order of accuracy and handle
ties by assigning tied models the average of their ranks. A tie between two
models occurs when a 95\% confidence interval statistical test on the
difference between their mean accuracies is inconclusive in rejecting the null
hypothesis that this difference is 0.
Our recommendation is that this test is ran to determine if ties occur.
As mentioned earlier, our source code will include this computation.

\paragraph{Complete main tables}
For completeness, Table~\ref{tab:iclr2020_tables} presents a more detailed version of
Table~\ref{tab:iclr2020_tables_without_ci} that also displays confidence intervals
and per-dataset ranks computed using the above procedure.
\begin{sidewaystable*}
\centering

\caption{Few-shot classification results on \benchmark.}
\label{tab:iclr2020_tables}

\subfloat[][Models trained on \textbf{ILSVRC-2012 only}.]{
\begin{tabular}{|l *{7}{r@{$\pm$}l|}}
\hline
\multirow{2}{*}{Test Source}& \multicolumn{14}{c|}{Method: Accuracy (\%) $\pm$ confidence (\%)} \\ \cline{2-15}
        & \multicolumn{2}{c|}{$k$-NN}    & \multicolumn{2}{c|}{Finetune}    & \multicolumn{2}{c|}{MatchingNet}    & \multicolumn{2}{c|}{ProtoNet}    & \multicolumn{2}{c|}{fo-MAML}    & \multicolumn{2}{c|}{RelationNet}     & \multicolumn{2}{c|}{Proto-MAML} \\\hline\hline
ILSVRC                      & 41.03       & 1.01 (6)         & 45.78       & 1.10 (4)           & 45.00       & 1.10 (4)              & \bf{50.50}  & 1.08 (1.5)         & 45.51       & 1.11 (4)          & 34.69       & 1.01 (7)               & \bf{49.53}  & 1.05 (1.5)      \\\hline
Omniglot                    & 37.07       & 1.15 (7)         & 60.85       & 1.58 (2.5)         & 52.27       & 1.28 (5)              & 59.98       & 1.35 (2.5)         & 55.55       & 1.54 (4)          & 45.35       & 1.36 (6)               & \bf{63.37}  & 1.33 (1)        \\\hline
Aircraft                    & 46.81       & 0.89 (6)         & \bf{68.69}  & 1.26 (1)           & 48.97       & 0.93 (5)              & 53.10       & 1.00 (4)           & 56.24       & 1.11 (2.5)        & 40.73       & 0.83 (7)               & 55.95       & 0.99 (2.5)      \\\hline
Birds                       & 50.13       & 1.00 (6.5)       & 57.31       & 1.26 (5)           & 62.21       & 0.95 (3.5)            & \bf{68.79}  & 1.01 (1.5)         & 63.61       & 1.06 (3.5)        & 49.51       & 1.05 (6.5)             & \bf{68.66}  & 0.96 (1.5)      \\\hline
Textures                    & 66.36       & 0.75 (4)         & \bf{69.05}  & 0.90 (1.5)         & 64.15       & 0.85 (6)              & 66.56       & 0.83 (4)           & \bf{68.04}  & 0.81 (1.5)        & 52.97       & 0.69 (7)               & 66.49       & 0.83 (4)        \\\hline
Quick Draw                  & 32.06       & 1.08 (7)         & 42.60       & 1.17 (4.5)         & 42.87       & 1.09 (4.5)            & 48.96       & 1.08 (2)           & 43.96       & 1.29 (4.5)        & 43.30       & 1.08 (4.5)             & \bf{51.52}  & 1.00 (1)        \\\hline
Fungi                       & 36.16       & 1.02 (4)         & 38.20       & 1.02 (3)           & 33.97       & 1.00 (5)              & \bf{39.71}  & 1.11 (1.5)         & 32.10       & 1.10 (6)          & 30.55       & 1.04 (7)               & \bf{39.96}  & 1.14 (1.5)      \\\hline
VGG Flower                  & 83.10       & 0.68 (4)         & 85.51       & 0.68 (2.5)         & 80.13       & 0.71 (6)              & 85.27       & 0.77 (2.5)         & 81.74       & 0.83 (5)          & 68.76       & 0.83 (7)               & \bf{87.15}  & 0.69 (1)        \\\hline
Traffic Signs               & 44.59       & 1.19 (6)         & \bf{66.79}  & 1.31 (1)           & 47.80       & 1.14 (3.5)            & 47.12       & 1.10 (5)           & 50.93       & 1.51 (2)          & 33.67       & 1.05 (7)               & 48.83       & 1.09 (3.5)      \\\hline
MSCOCO                      & 30.38       & 0.99 (6.5)       & 34.86       & 0.97 (4)           & 34.99       & 1.00 (4)              & 41.00       & 1.10 (2)           & 35.30       & 1.23 (4)          & 29.15       & 1.01 (6.5)             & \bf{43.74}  & 1.12 (1)        \\\hline\hline
\textbf{Avg. rank}          & \multicolumn{2}{c|}{5.7}       & \multicolumn{2}{c|}{2.9}         & \multicolumn{2}{c|}{4.65}           & \multicolumn{2}{c|}{2.65}        & \multicolumn{2}{c|}{3.7}        & \multicolumn{2}{c|}{6.55}            & \multicolumn{2}{c|}{\bf{1.85}}\\\hline
\end{tabular}}

\subfloat[][Models trained on \textbf{all datasets}.]{
\begin{tabular}{|l *{7}{r@{$\pm$}l|}}
\hline
\multirow{2}{*}{Test Source}& \multicolumn{14}{c|}{Method: Accuracy (\%) $\pm$ confidence (\%) (rank)} \\ \cline{2-15}
        & \multicolumn{2}{c|}{$k$-NN}    & \multicolumn{2}{c|}{Finetune}    & \multicolumn{2}{c|}{MatchingNet}    & \multicolumn{2}{c|}{ProtoNet}    & \multicolumn{2}{c|}{fo-MAML}    & \multicolumn{2}{c|}{RelationNet}     & \multicolumn{2}{c|}{Proto-MAML} \\\hline\hline
ILSVRC                      & 38.55       & 0.94 (4.5)       & 43.08       & 1.08 (2.5)         & 36.08       & 1.00 (6)              & 44.50       & 1.05 (2.5)         & 37.83       & 1.01 (4.5)        & 30.89       & 0.93 (7)               & \bf{46.52}  & 1.05 (1)        \\\hline
Omniglot                    & 74.60       & 1.08 (6)         & 71.11       & 1.37 (7)           & 78.25       & 1.01 (4.5)            & 79.56       & 1.12 (4.5)         & 83.92       & 0.95 (2.5)        & \bf{86.57}  & 0.79 (1)               & 82.69       & 0.97 (2.5)      \\\hline
Aircraft                    & 64.98       & 0.82 (7)         & 72.03       & 1.07 (3.5)         & 69.17       & 0.96 (5.5)            & 71.14       & 0.86 (3.5)         & \bf{76.41}  & 0.69 (1)          & 69.71       & 0.83 (5.5)             & 75.23       & 0.76 (2)        \\\hline
Birds                       & 66.35       & 0.92 (2.5)       & 59.82       & 1.15 (5)           & 56.40       & 1.00 (6)              & 67.01       & 1.02 (2.5)         & 62.43       & 1.08 (4)          & 54.14       & 0.99 (7)               & \bf{69.88}  & 1.02 (1)        \\\hline
Textures                    & 63.58       & 0.79 (5)         & \bf{69.14}  & 0.85 (1.5)         & 61.80       & 0.74 (6)              & 65.18       & 0.84 (3.5)         & 64.14       & 0.83 (3.5)        & 56.56       & 0.73 (7)               & \bf{68.25}  & 0.81 (1.5)      \\\hline
Quick Draw                  & 44.88       & 1.05 (7)         & 47.05       & 1.16 (6)           & 60.81       & 1.03 (3.5)            & 64.88       & 0.89 (2)           & 59.73       & 1.10 (5)          & 61.75       & 0.97 (3.5)             & \bf{66.84}  & 0.94 (1)        \\\hline
Fungi                       & 37.12       & 1.06 (3.5)       & 38.16       & 1.04 (3.5)         & 33.70       & 1.04 (6)              & 40.26       & 1.13 (2)           & 33.54       & 1.11 (6)          & 32.56       & 1.08 (6)               & \bf{41.99}  & 1.17 (1)        \\\hline
VGG Flower                  & 83.47       & 0.61 (4)         & 85.28       & 0.69 (3)           & 81.90       & 0.72 (5)              & 86.85       & 0.71 (2)           & 79.94       & 0.84 (6)          & 76.08       & 0.76 (7)               & \bf{88.72}  & 0.67 (1)        \\\hline
Traffic Signs               & 40.11       & 1.10 (6)         & \bf{66.74}  & 1.23 (1)           & 55.57       & 1.08 (2)              & 46.48       & 1.00 (4)           & 42.91       & 1.31 (5)          & 37.48       & 0.93 (7)               & 52.42       & 1.08 (3)        \\\hline
MSCOCO                      & 29.55       & 0.96 (5)         & 35.17       & 1.08 (3)           & 28.79       & 0.96 (5)              & 39.87       & 1.06 (2)           & 29.37       & 1.08 (5)          & 27.41       & 0.89 (7)               & \bf{41.74}  & 1.13 (1)        \\\hline\hline
\textbf{Avg. rank}          & \multicolumn{2}{c|}{5.05}      & \multicolumn{2}{c|}{3.6}         & \multicolumn{2}{c|}{4.95}           & \multicolumn{2}{c|}{2.85}        & \multicolumn{2}{c|}{4.25}       & \multicolumn{2}{c|}{5.8}             & \multicolumn{2}{c|}{\bf{1.5}} \\\hline
\end{tabular}
}
\end{sidewaystable*}

\subsection{Analysis of Performance Across Shots and Ways}
\label{sec:appendix_analysis}

\begin{figure*}[htp]
  \centering
  \subfloat[\label{fig:all_way_vs_accuracy} Ways Analysis]{\includegraphics[width=0.49\textwidth]{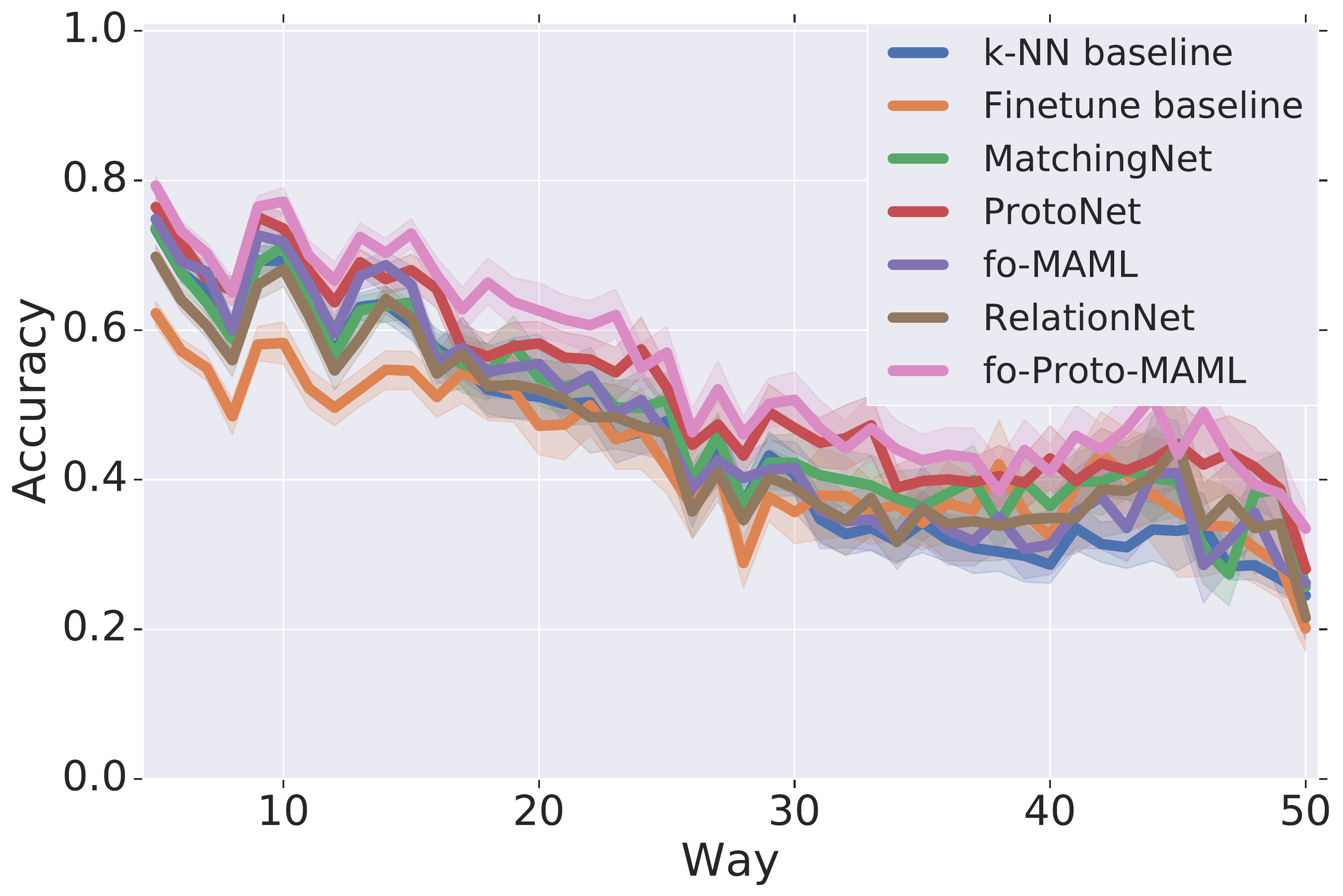}}\hspace{0.1cm}
	\subfloat[\label{fig:all_shot_vs_precision} Shots Analysis]{\includegraphics[width=0.49\textwidth]{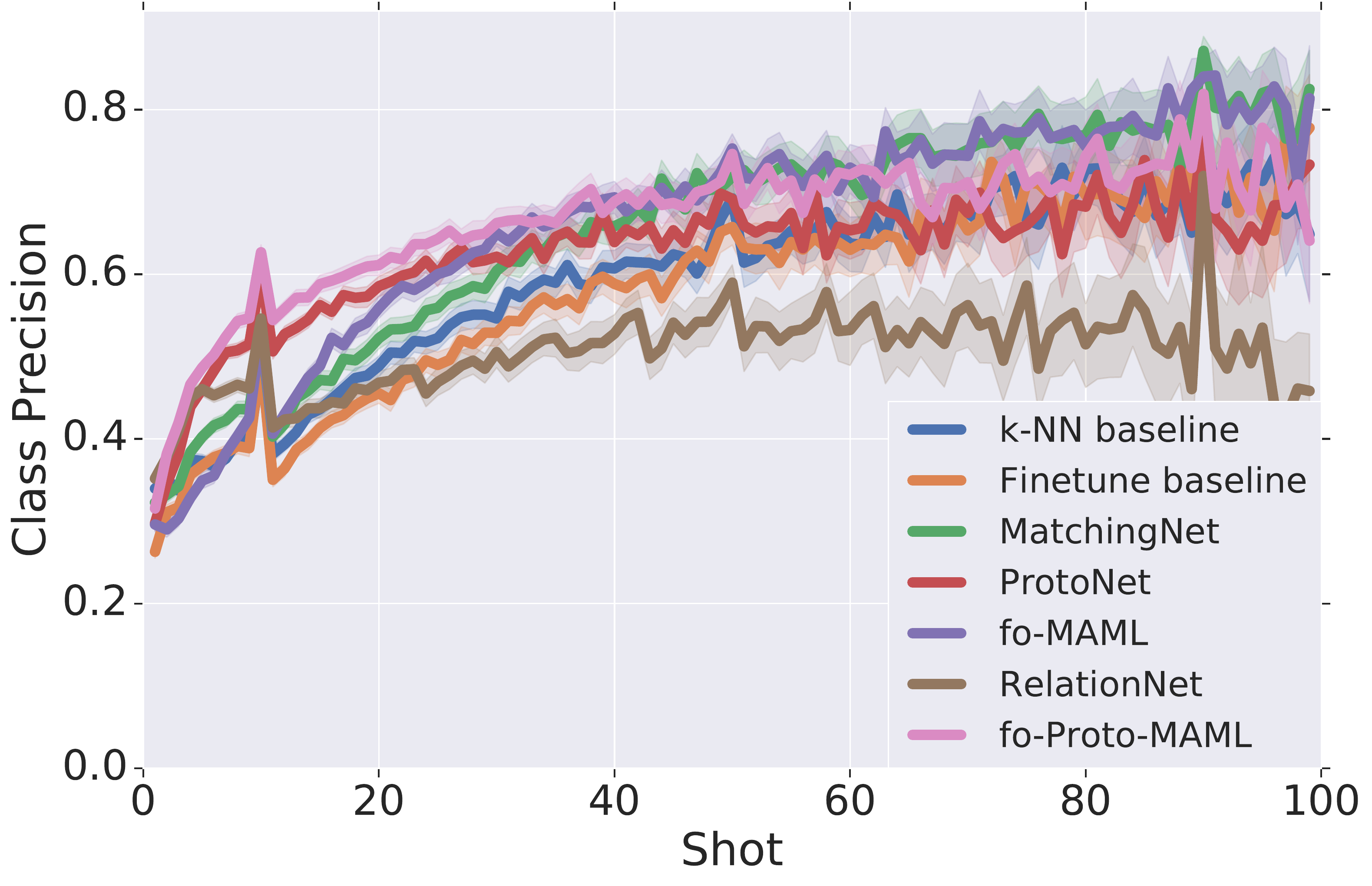}}\hspace{0.1cm}
	\caption{\label{fig:analyses_ways_and_shots_extra} Analysis of performance as a function of the episode's way, shots for models whose training source is (the training data of) all datasets. The bands display 95\% confidence intervals.}
\end{figure*}

For completeness, in Figure~\ref{fig:analyses_ways_and_shots_extra} we show the results of the analysis of the robustness to
different ways and shots for the variants of the models that were trained on
all datasets. We observe the same trends as discussed in our Experiments
section for the variants of the models that were trained on ImageNet.

\subsection{Effect of training on all datasets over training on ILSVRC-2012 only}
\label{sec:appendix_trainsource}

\begin{figure*}[htp]
	\caption{Accuracy on the test datasets, when training on ILSVRC only or All datasets (same results as shown in the main tables). The bars display 95\% confidence intervals.}
  \label{fig:trainsource}
    \includegraphics[width=\textwidth]{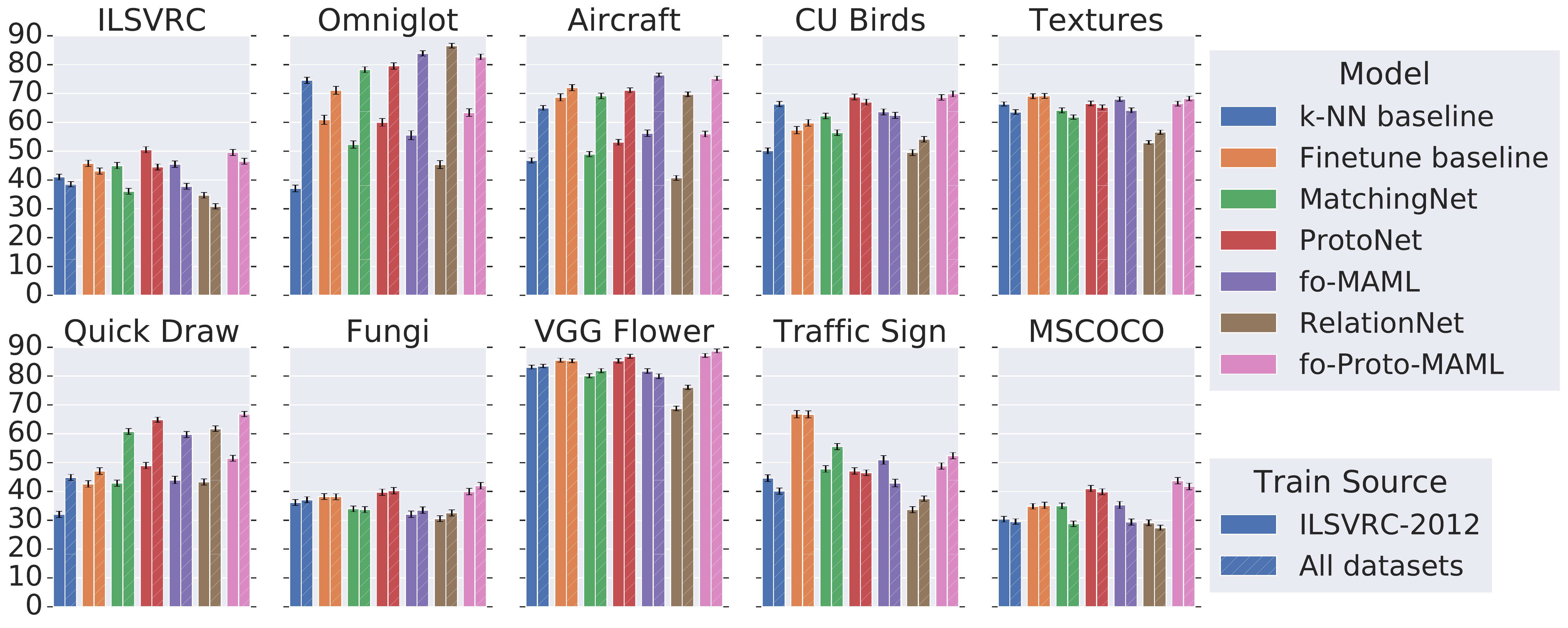}
\end{figure*}

For more clearly observing whether training on all datasets leads to improved
generalization over training on ImageNet only, Figure~\ref{fig:trainsource}
shows side-to-side the performance of each model trained on ILSVRC only vs. all datasets.
The difference between the performance of the all-dataset trained models versus the ImageNet-only trained ones is also visualized in Figure~\ref{fig:trainsource_diff} in the main paper.

As discussed in the main paper, we notice that we do not always observe a clear generalization advantage in training from a 
wider collection of image datasets. While some of the
datasets that were added to the meta-training phase did see an improvement
across all models, in particular for Omniglot and Quick Draw, this was not true
across the board. In fact, in certain cases the performance is slightly worse. We believe
that more successfully leveraging diverse sources of data is an interesting
open research problem.

\subsection{Effect of pre-training versus training from scratch}
For each meta-learner, we selected the best model (based on validation on ImageNet's validation split) out of the ones that used the pre-trained initialization, and the best out of the ones that trained from scratch. We then ran the evaluation of each on (the test split of) all datasets in order to quantify how beneficial this pre-trained initialization is. We performed this experiment twice: for the models that are trained on ImageNet only and for the models that are trained on (the training splits of) all datasets. 

\begin{figure*}[htp]
   \centering
	\subfloat[\label{fig:pretrain_analysis_imagenet_bigger} ImageNet.]{\includegraphics[scale=0.2]{figures_iclr2020/imagenet_pretrained_vs_scratch.pdf}}\hspace{0.01cm}
	\subfloat[\label{fig:pretrain_analysis_all_datasets_bigger} All datasets.]{\includegraphics[scale=0.2]{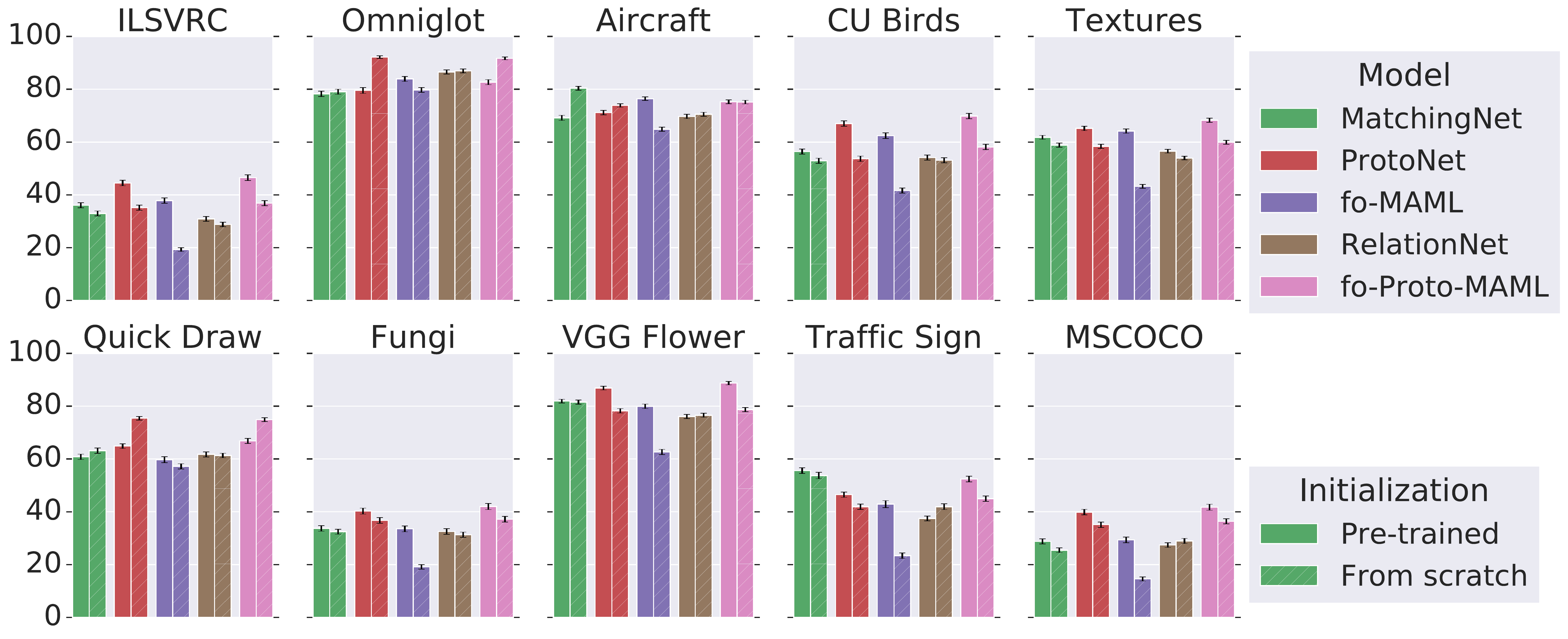}}\hspace{0.01cm}
   \caption{\label{fig:pretrain_analysis_bigger} Comparing pre-training to starting from scratch. Same plots as Figure~\ref{fig:effect_of_pretrain_imagenet} and Figure~\ref{fig:effect_of_pretrain_all_datasets}, only larger.}
 \end{figure*}

\begin{figure*}[htp]
   \centering
	\subfloat[\label{fig:pretrain_analysis_extra_imagenet} The gain from pre-training (ImageNet).]{\includegraphics[width=0.49\textwidth]{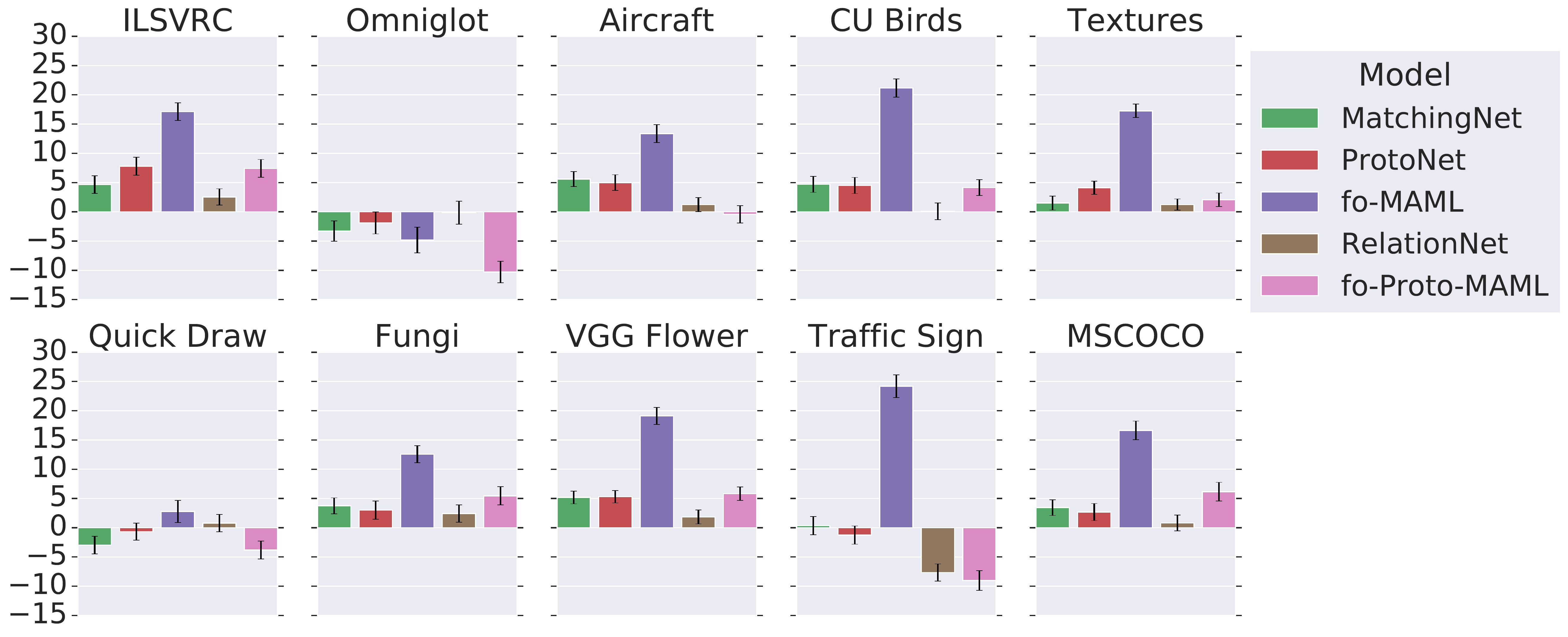}}\hspace{0.01cm}
	\subfloat[\label{fig:pretrain_analysis_extra_all_datasets} The gain from pre-training (All datasets).]{\includegraphics[width=0.49\textwidth]{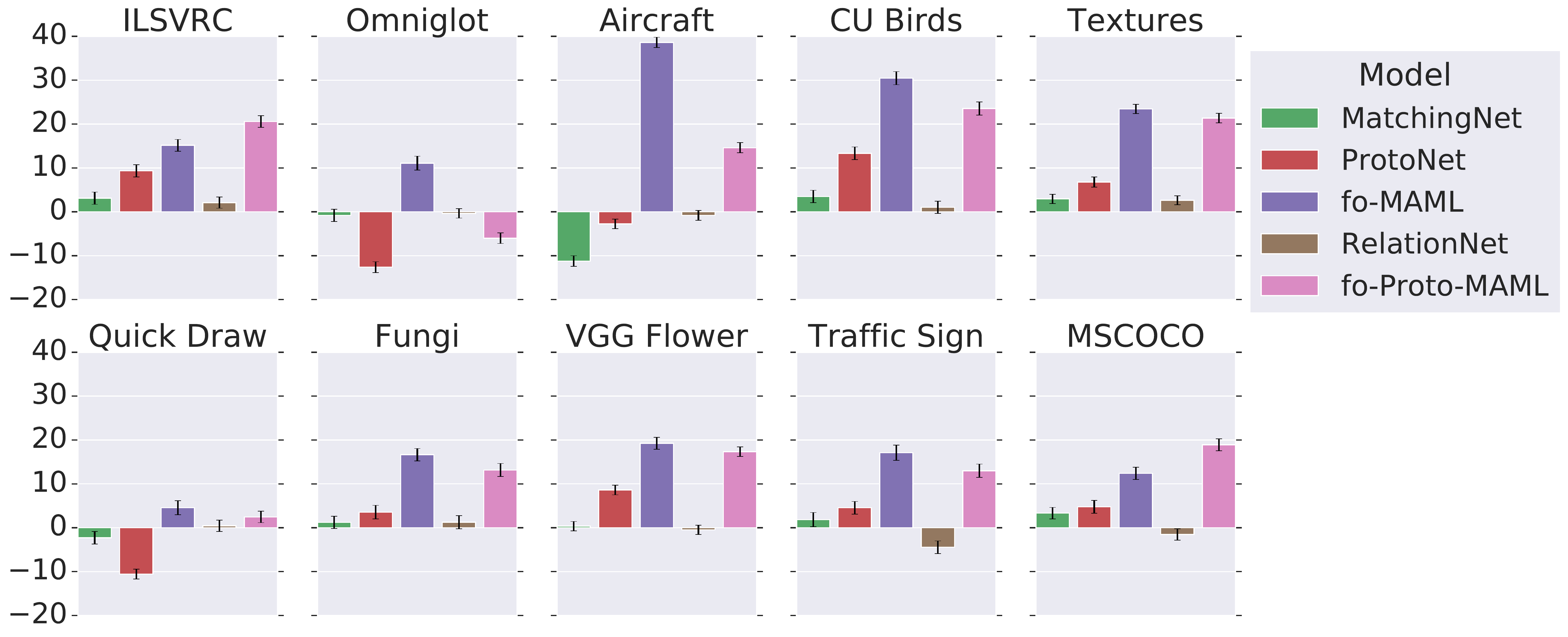}}\hspace{0.01cm}
   \caption{\label{fig:pretrain_analysis_extra} The performance difference of initializing the embedding weights from a pre-trained solution, before episodically training on ImageNet or all datasets, over using a random initialization of those weights. The pre-trained weights that we consider are the ones that the $k$-NN baseline converged to when it was trained on ImageNet. Positive values indicate that this pre-training was beneficial.}
 \end{figure*}

The results of this investigation were reported in the main paper in Figure~\ref{fig:effect_of_pretrain_imagenet} and Figure~\ref{fig:effect_of_pretrain_all_datasets}, for ImageNet-only training and all dataset training, respectively. 
We show the same results in Figure~\ref{fig:pretrain_analysis_bigger}, printed larger to facilitate viewing of error bars.
For easier comparison, we also plot the difference in performance of the models that were pre-trained over the ones that weren't, in Figures~\ref{fig:pretrain_analysis_extra_imagenet} and~\ref{fig:pretrain_analysis_extra_all_datasets}. These figures make it easier to spot that while using the pre-trained solution usually helps for datasets that are visually not too different from ImageNet, it may hurt for datasets that are significantly different from it, such as Omniglot, Quickdraw (and surprisingly Aircraft). Note that these three datasets are the same three that we found benefit from training on All datasets instead of ImageNet-only. It appears that using the pre-trained solution biases the final solution to specialize on ImageNet-like datasets. 

\subsection{Effect of meta-learning versus inference-only}
\begin{figure*}[htp]
   \centering
	\subfloat[\label{fig:meta_train_analysis_imagenet_bigger} ImageNet.]{\includegraphics[scale=0.2]{figures_iclr2020/imagenet_meta_train_vs_inference_only.pdf}}\hspace{0.01cm}
	\subfloat[\label{fig:meta_train_analysis_all_datasets_bigger} All datasets.]{\includegraphics[scale=0.2]{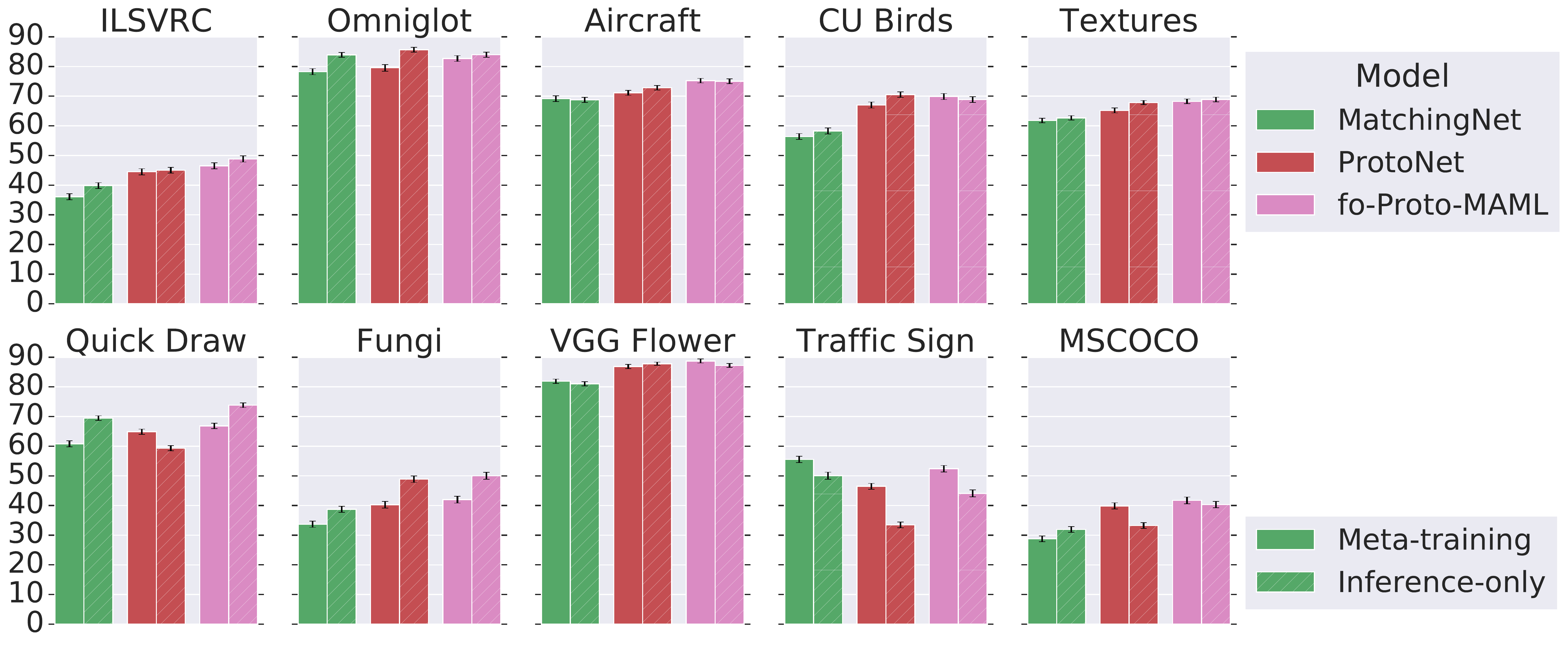}}\hspace{0.01cm}
   \caption{\label{fig:meta_train_analysis_bigger} Comparing the meta-trained variant of meta-learners against their inference-only counterpart. Same plots as Figure~\ref{fig:effect_of_meta_learn_imagenet} and Figure~\ref{fig:effect_of_meta_learn_all_datasets}, only larger.}
 \end{figure*}

\begin{figure*}[htp]
   \centering
\subfloat[\label{fig:meta_train_analysis_extra_imagenet} The gain from meta-training on ImageNet.]{\includegraphics[width=0.49\textwidth]{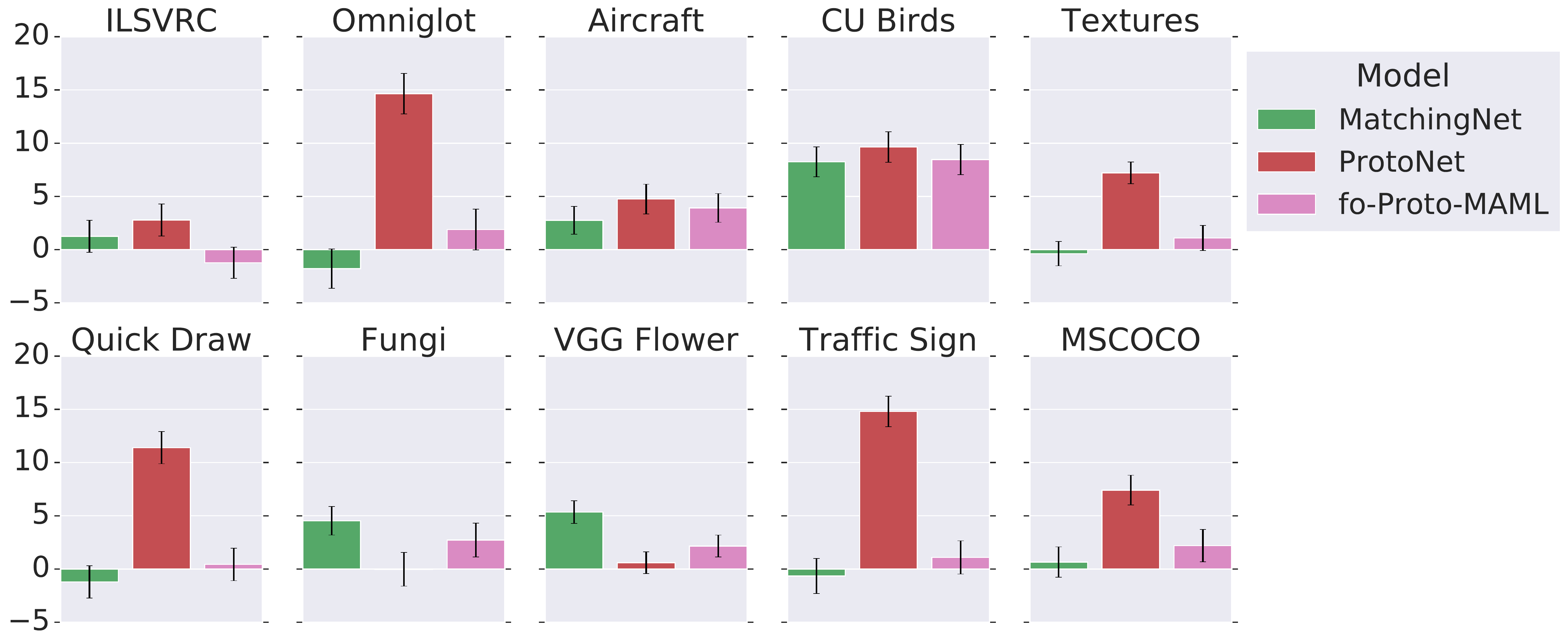}}\hspace{0.01cm}
\subfloat[\label{fig:meta_train_analysis_extra_all_datasets} The gain from meta-training on All datasets.]{\includegraphics[width=0.49\textwidth]{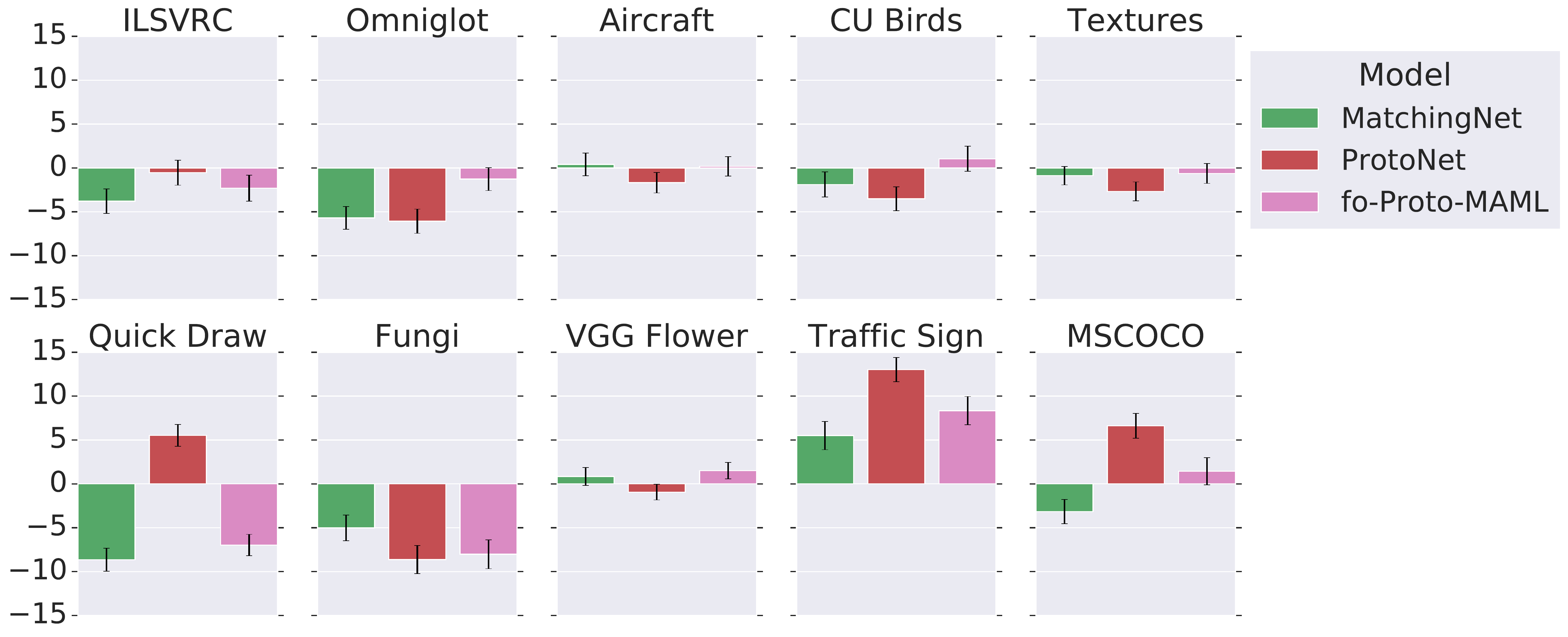}}\hspace{0.01cm}
   \caption{\label{fig:meta_train_analysis_extra} The performance difference of meta-learning over the corresponding inference-only baseline of each meta-learner. Positive values indicate that meta-learning was beneficial.}
 \end{figure*}

Figure~\ref{fig:meta_train_analysis_bigger} shows the same plots as in Figures~\ref{fig:effect_of_meta_learn_imagenet} and~\ref{fig:effect_of_meta_learn_all_datasets} but printed larger to facilitate viewing of error bars.
Furthermore, as we have done for visualizing the observed gain of pre-training, we also present in Figures~\ref{fig:meta_train_analysis_extra_imagenet} and~\ref{fig:meta_train_analysis_extra_all_datasets} the gain observed from meta-learning as opposed to training the corresponding inference-only baseline, as explained in the Experiments section of the main paper. This visulization makes it clear that while meta-training usually helps on ImageNet (or doesn't hurt too much), it sometimes hurts when it is performed on all datasets, emphasizing the need for further research into best practices of meta-learning across heterogeneous sources.

\subsection{Finegrainedness Analysis}
\begin{figure*}[htp]
   \centering
\subfloat[\label{fig:test_finegrain_extra} Fine-grainedness Analysis (on ImageNet's test graph)]{\includegraphics[width=0.49\textwidth]{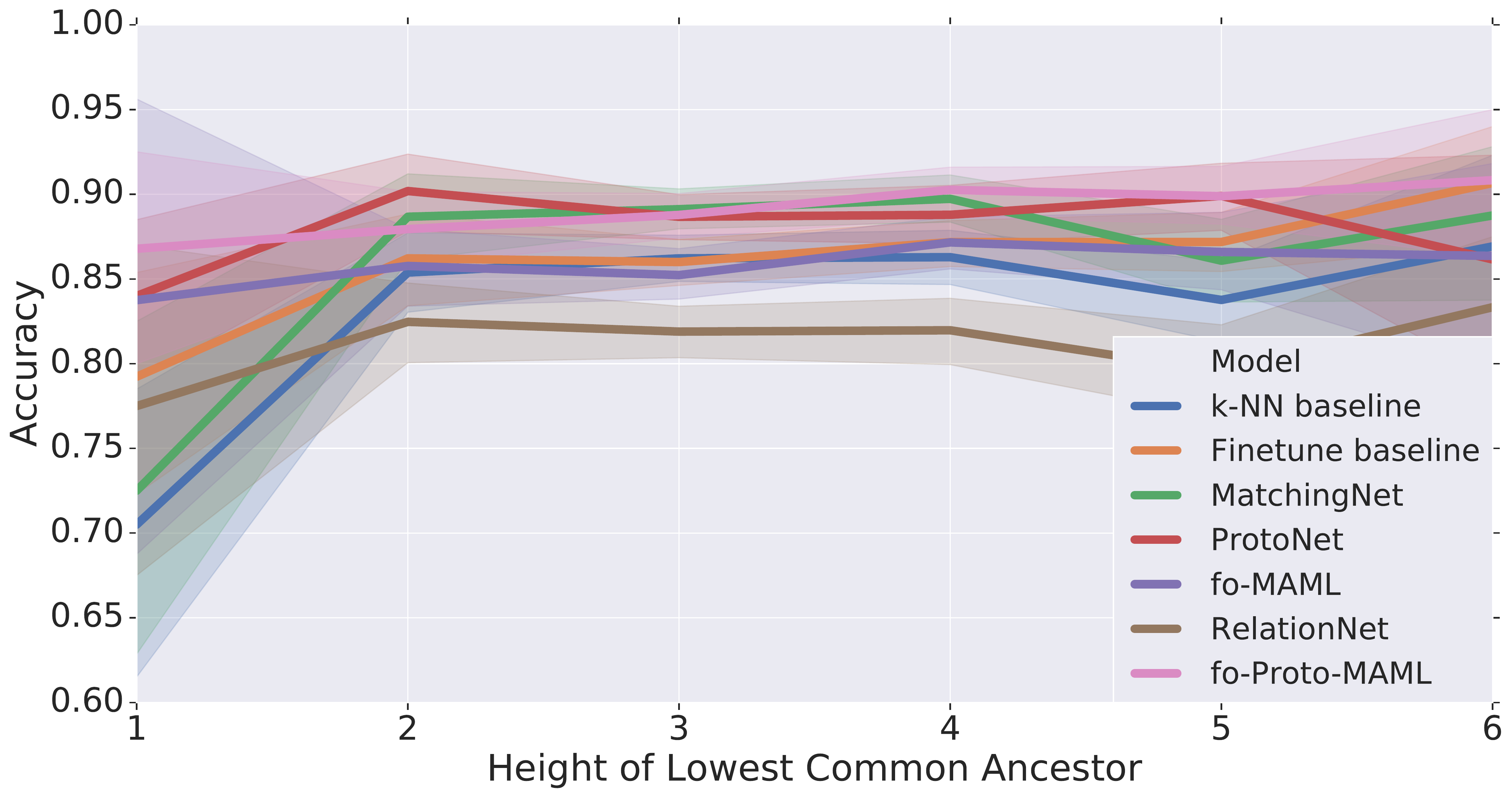}}\hspace{0.01cm}
\subfloat[\label{fig:train_finegrain_extra} Fine-grainedness Analysis (on ImageNet's train graph graph)]{\includegraphics[width=0.49\textwidth]{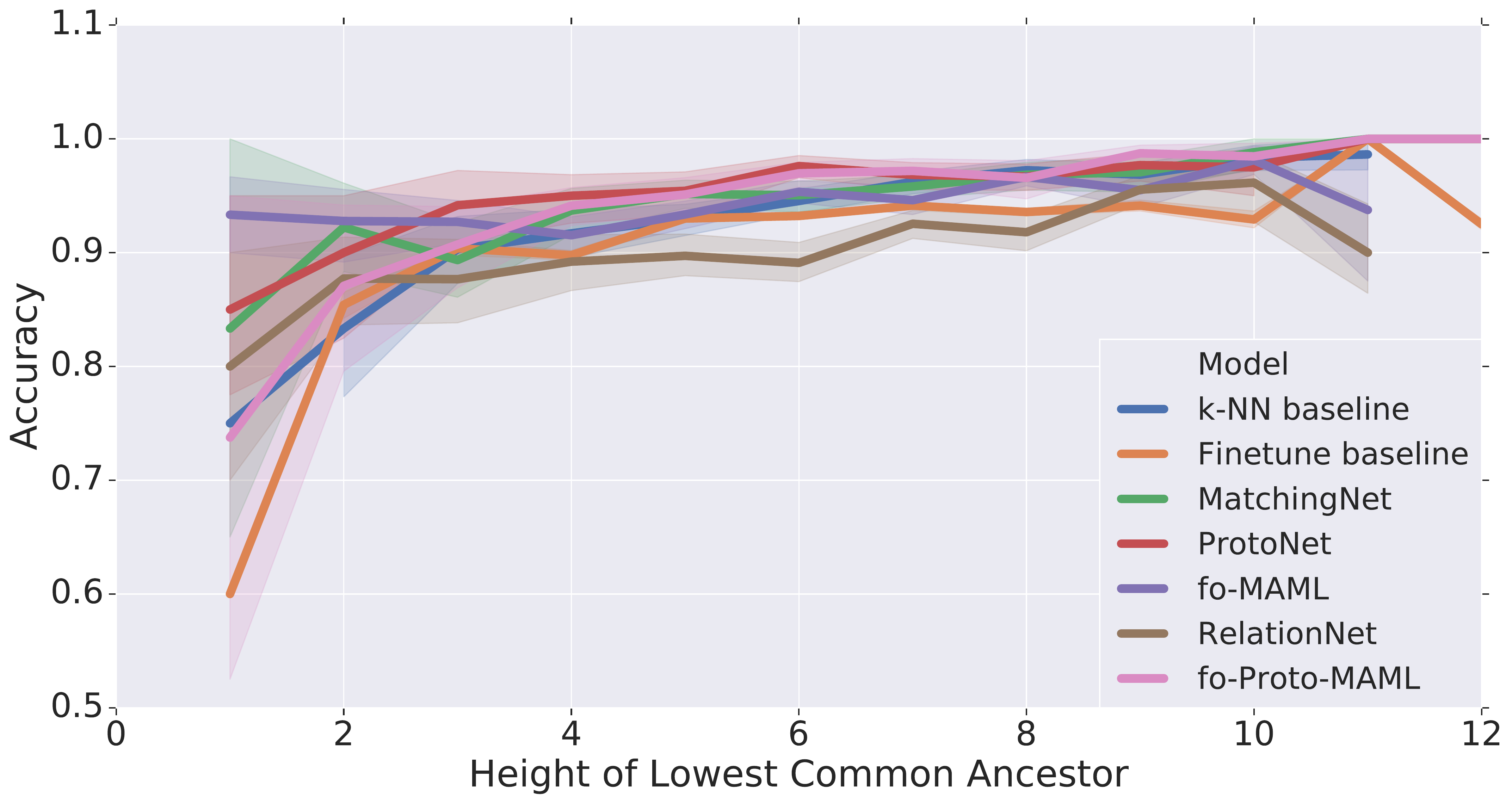}}\hspace{0.01cm}
   \caption{\label{fig:analyses_finegrain_extra} Analysis of performance as a function of the degree of fine-grainedness. Larger heights correspond to coarser-grained tasks. The bands display 95\% confidence intervals.}
 \end{figure*}

We investigate the hypothesis that finer-grained tasks are more challenging than coarse-grained ones by creating binary ImageNet episodes with the two classes chosen uniformly at random from the DAG's set of leaves. We then define the degree of coarse-grainedness of a
task as the height of the lowest common ancestor of the two chosen leaves,
where the height is defined as the length of the longest path from the lowest
common ancestor to one of the selected leaves. Larger heights then correspond
to coarser-grained tasks. We present these results in Figure~\ref{fig:analyses_finegrain_extra}. We do not detect a significant trend when
performing this analysis on the test DAG. The results on the training DAG, though, do seem to indicate that our
hypothesis holds to some extent. We conjecture that this may be due to the
richer structure of the training DAG, but we encourage further investigation.

\subsection{The importance of MAML's inner-loop learning rate hyperparameter.}
The camera-ready version includes updated results for MAML and Proto-MAML
following an external suggestion to experiment with larger values for the
inner-loop learning rate $\alpha$ of MAML. We found that re-doing our
hyperparameter search with a revised range that includes larger $\alpha$ values
significantly improved fo-MAML's performance on \benchmark. For consistency, we
applied the same change to fo-Proto-MAML and re-ran those experiments too.

We found that the value of this $\alpha$ that performs best for fo-MAML both
for training on ImageNet only and training on all datasets is approximately
0.1, which is an order of magnitude larger than our previous best value.
Interestingly, fo-Proto-MAML does not choose such a large $\alpha$ value, with
best $\alpha$ being 0.0054 when training on ImageNet only and 0.02 when
training on all datasets. Plausibly this difference can be attributed to the
better initialization of Proto-MAML which requires a less aggressive
optimization for the adaptation to each new task. This hypothesis is also
supported by the fact that fo-Proto-MAML chooses to take fewer adaptation steps
than fo-MAML does. The complete set of best discovered hyperparameters is
available in our public code.

To emphasize the importance of properly tuning this hyperparameter,
Table~\ref{tab:maml_before_and_after} displays the previous best and the new
best fo-MAML results side-by-side, showcasing the large performance gap when
using the appropriate value for $\alpha$.
\begin{table*}[t]
\centering

\caption{Improvement of fo-MAML when using a larger inner learning rate $\alpha$.}
\label{tab:maml_before_and_after}

\subfloat[][Models trained on \textbf{ILSVRC-2012 only}.]{
\begin{tabular}{|l *{2}{r@{$\pm$}l|}}
\hline
\multirow{2}{*}{Test Source}& \multicolumn{4}{c|}{Method: Accuracy (\%) $\pm$ confidence (\%)} \\ \cline{2-5}
			    & \multicolumn{2}{c|}{fo-MAML $\alpha = 0.01$ (old)} & \multicolumn{2}{c|}{fo-MAML $\alpha \approx 0.1$} \\\hline\hline
	ILSVRC            & 36.09       & 1.01          & 45.51       & 1.11	\\\hline
	Omniglot          & 38.67       & 1.39          & 55.55       & 1.54	\\\hline
	Aircraft          & 34.50       & 0.90          & 56.24       & 1.11	\\\hline
 	Birds             & 49.10       & 1.18          & 63.61       & 1.06	\\\hline
	Textures          & 56.50       & 0.80          & 68.04       & 0.81	\\\hline
	Quick Draw        & 27.24       & 1.24          & 43.96       & 1.29	\\\hline
	Fungi             & 23.50       & 1.00          & 32.10       & 1.10	\\\hline
	VGG Flower        & 66.42       & 0.96          & 81.74       & 0.83	\\\hline
	Traffic Signs     & 33.23       & 1.34          & 50.93       & 1.51	\\\hline
	MSCOCO            & 27.52       & 1.11          & 35.30       & 1.23	\\\hline
\end{tabular}}

\subfloat[][Models trained on \textbf{all datasets}.]{
\begin{tabular}{|l *{2}{r@{$\pm$}l|}}
\hline
\multirow{2}{*}{Test Source}& \multicolumn{4}{c|}{Method: Accuracy (\%) $\pm$ confidence (\%)} \\ \cline{2-5}
			    & \multicolumn{2}{c|}{fo-MAML $\alpha = 0.01$ (old)} & \multicolumn{2}{c|}{fo-MAML $\alpha \approx 0.1$} \\\hline\hline
	ILSVRC            & 32.36       & 1.02          & 37.83       & 1.01	\\\hline
	Omniglot          & 71.91       & 1.20          & 83.92       & 0.95	\\\hline
	Aircraft          & 52.76       & 0.90          & 76.41       & 0.69	\\\hline
 	Birds             & 47.24       & 1.14          & 62.43       & 1.08	\\\hline
	Textures          & 56.66       & 0.74          & 64.16       & 0.83	\\\hline
	Quick Draw        & 50.50       & 1.19          & 59.73       & 1.10	\\\hline
	Fungi             & 21.02       & 0.99          & 33.54       & 1.11	\\\hline
	VGG Flower        & 70.93       & 0.99          & 79.94       & 0.84	\\\hline
	Traffic Signs     & 34.18       & 1.26          & 42.91       & 1.31	\\\hline
	MSCOCO            & 24.05       & 1.10          & 29.37       & 1.08	\\\hline
\end{tabular}}
\end{table*}

\subsection{The choice of a meta-validation procedure for \benchmark}

The design choice we made, as discussed in the main paper, is to use (the meta-validation set of) ImageNet only for model selection in all of our experiments. In the absence of previous results on the topic, this is a reasonable strategy since ImageNet has been known to consitute a useful proxy for performance on other datasets. However, it is likely that this is not the optimal choice: there might be certain hyperparameters for a given model that work best for held-out ImageNet episodes, but not for held-out episodes of other datasets. An alternative meta-validation scheme would be to use the average (across datasets) validation accuracy as an indicator for early stopping and model selection. We did not choose this method due to concerns about the reliability of this average performance. Notably, taking a simple average would over-emphasize larger datasets, or might over-emphasize datasets with natural images (as opposed to Omniglot and Quickdraw). Nevertheless, whether this strategy is beneficial is an interesting empirical question.

\subsection{Additional Per-dataset Analysis of Shots and Ways}
\begin{figure*}[htp]
\subfloat[ILSVRC]{\includegraphics[width=0.24\textwidth]{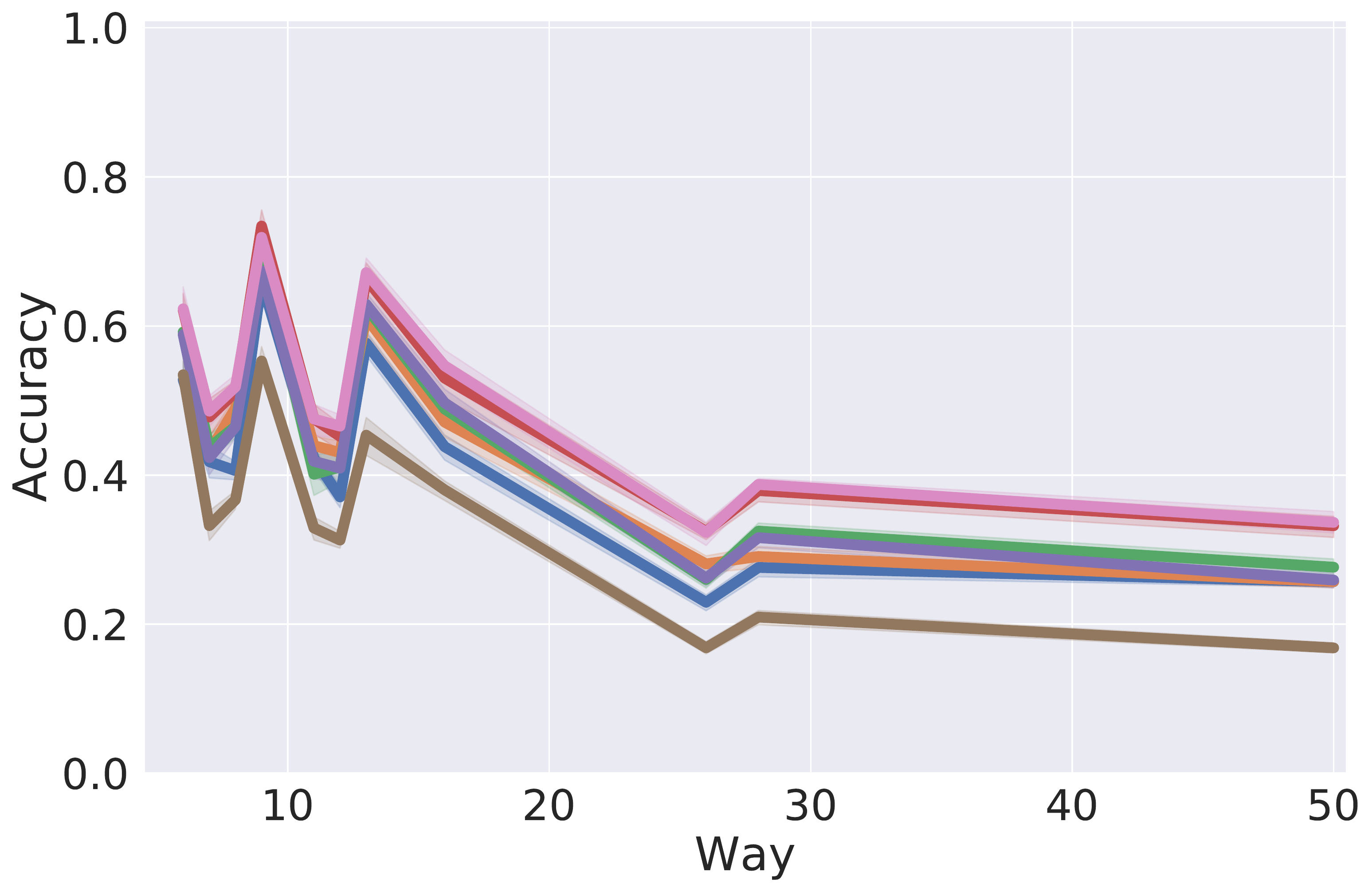}}\hspace{0.01cm}
\subfloat[Omniglot]{\includegraphics[width=0.24\textwidth]{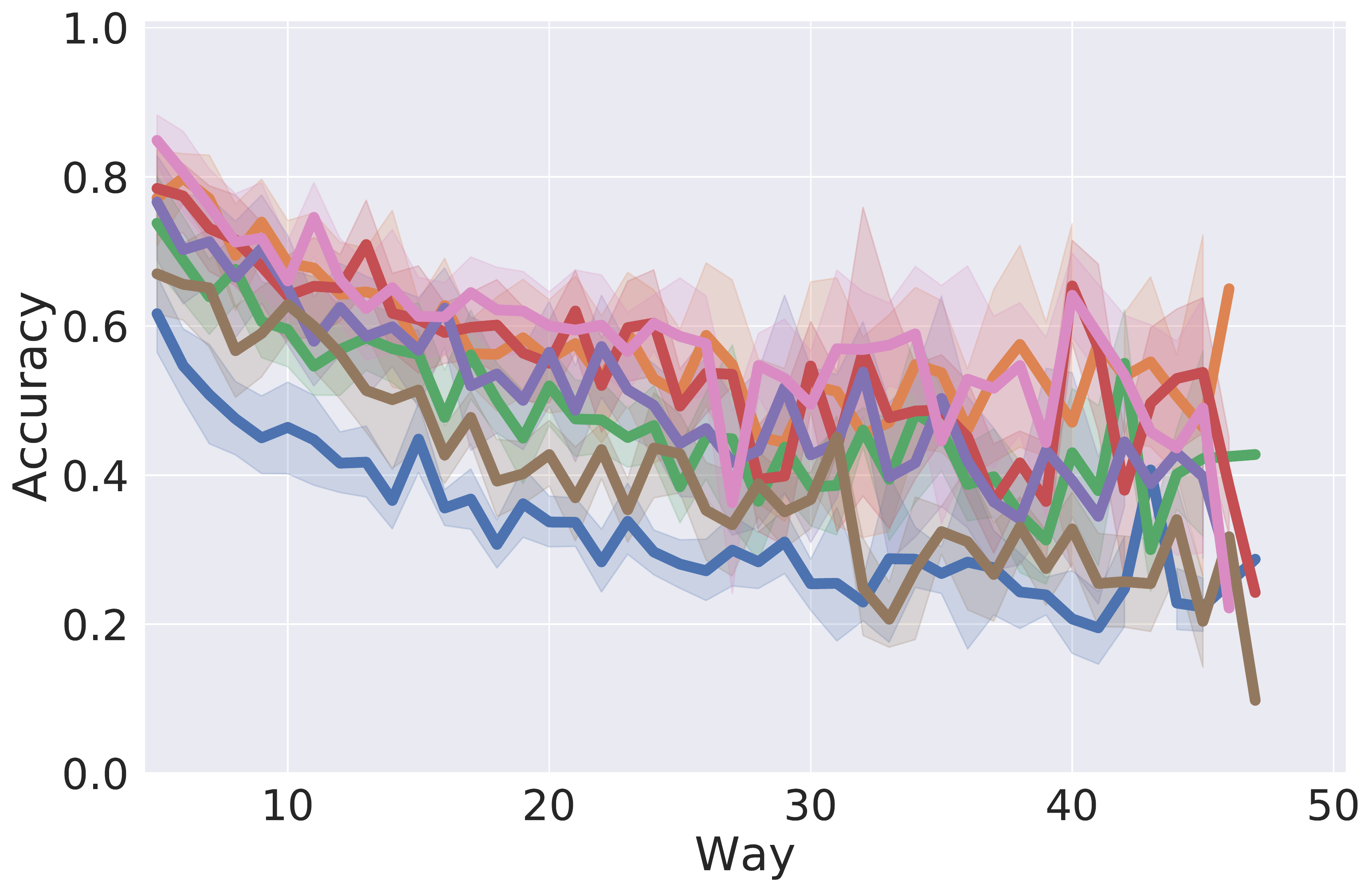}}\hspace{0.01cm}
\subfloat[Aircraft]{\includegraphics[width=0.24\textwidth]{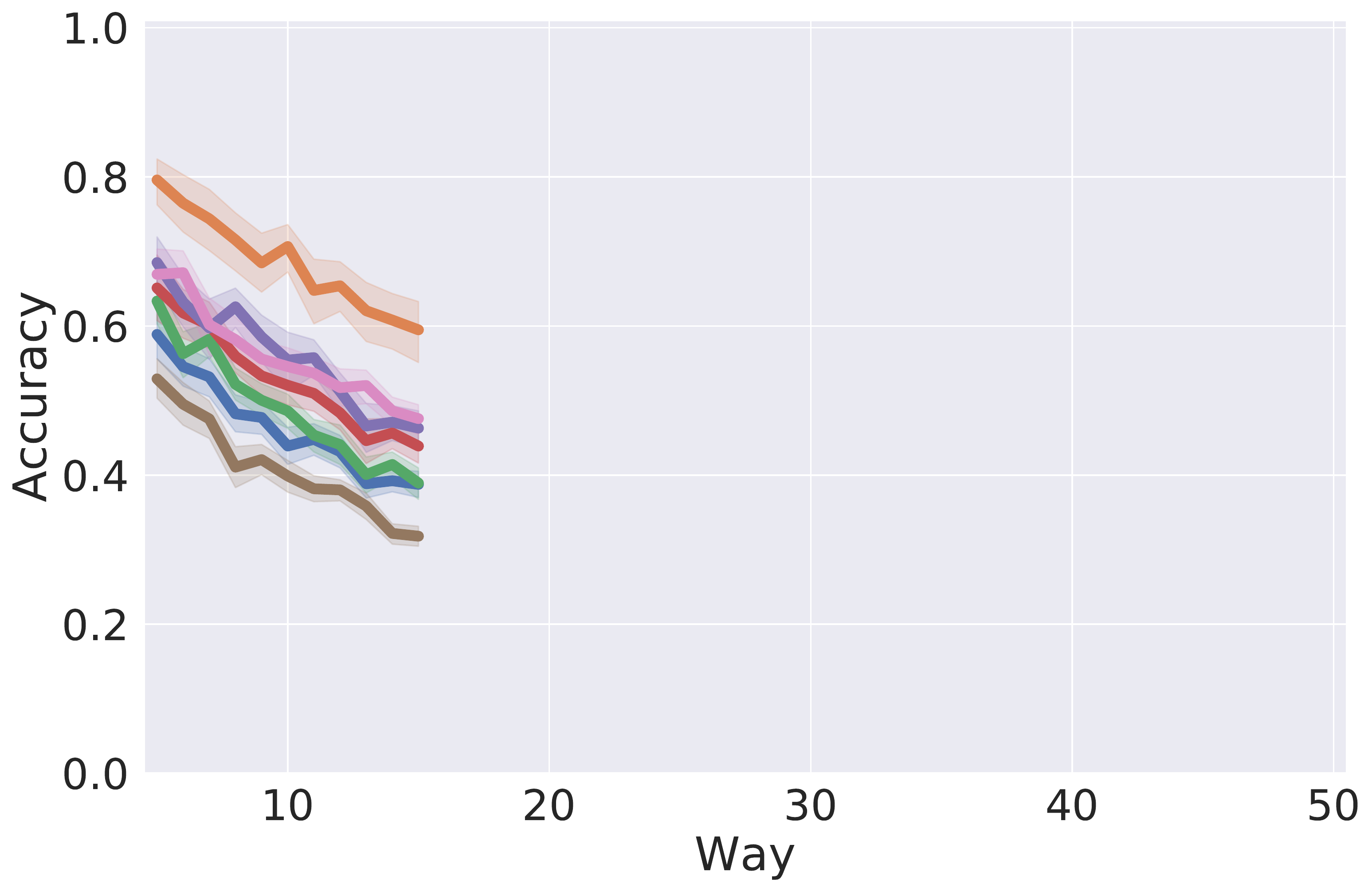}}\hspace{0.01cm}
\subfloat[Birds]{\includegraphics[width=0.24\textwidth]{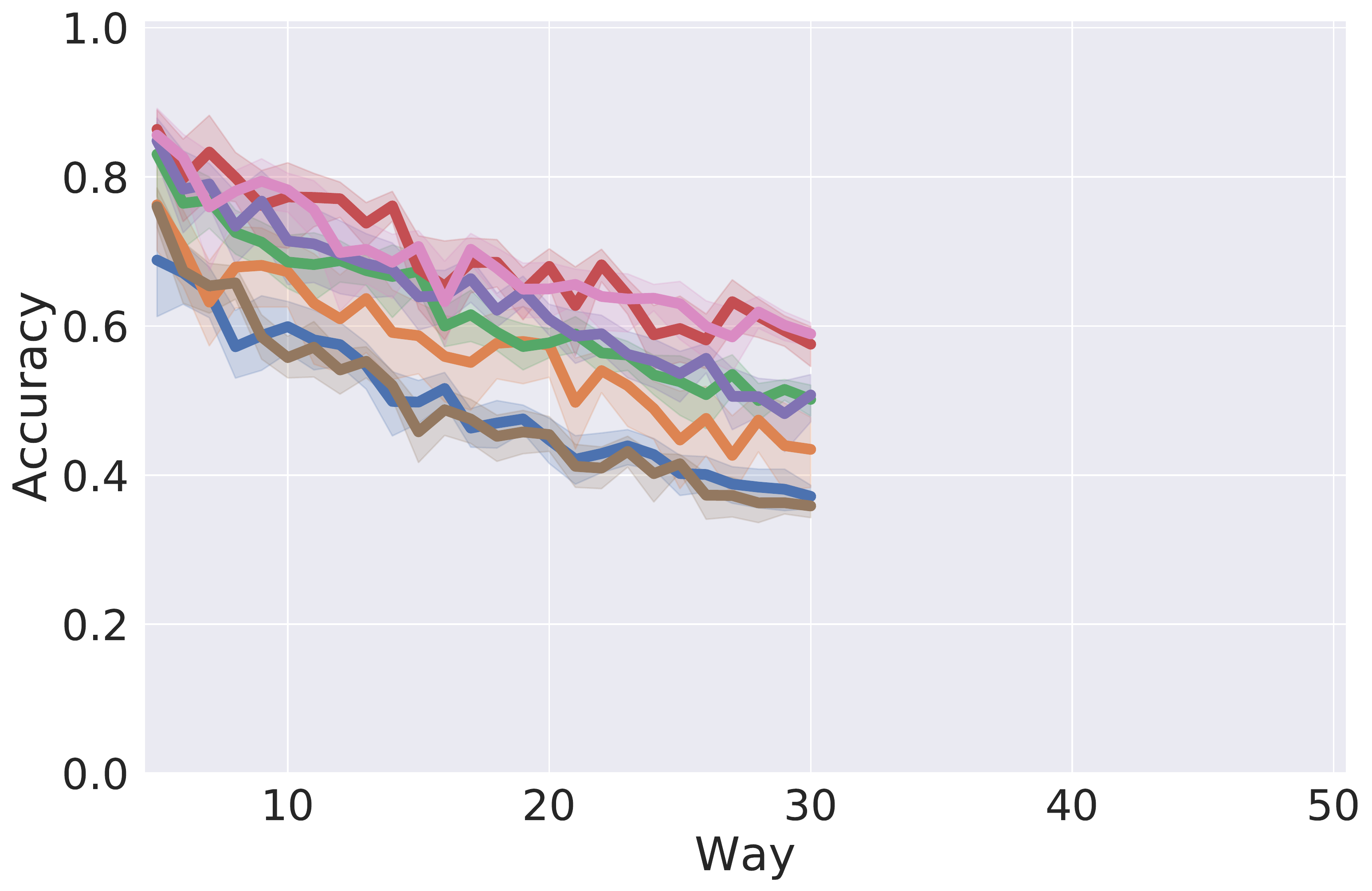}}\hspace{0.01cm}
\subfloat[Textures]{\includegraphics[width=0.24\textwidth]{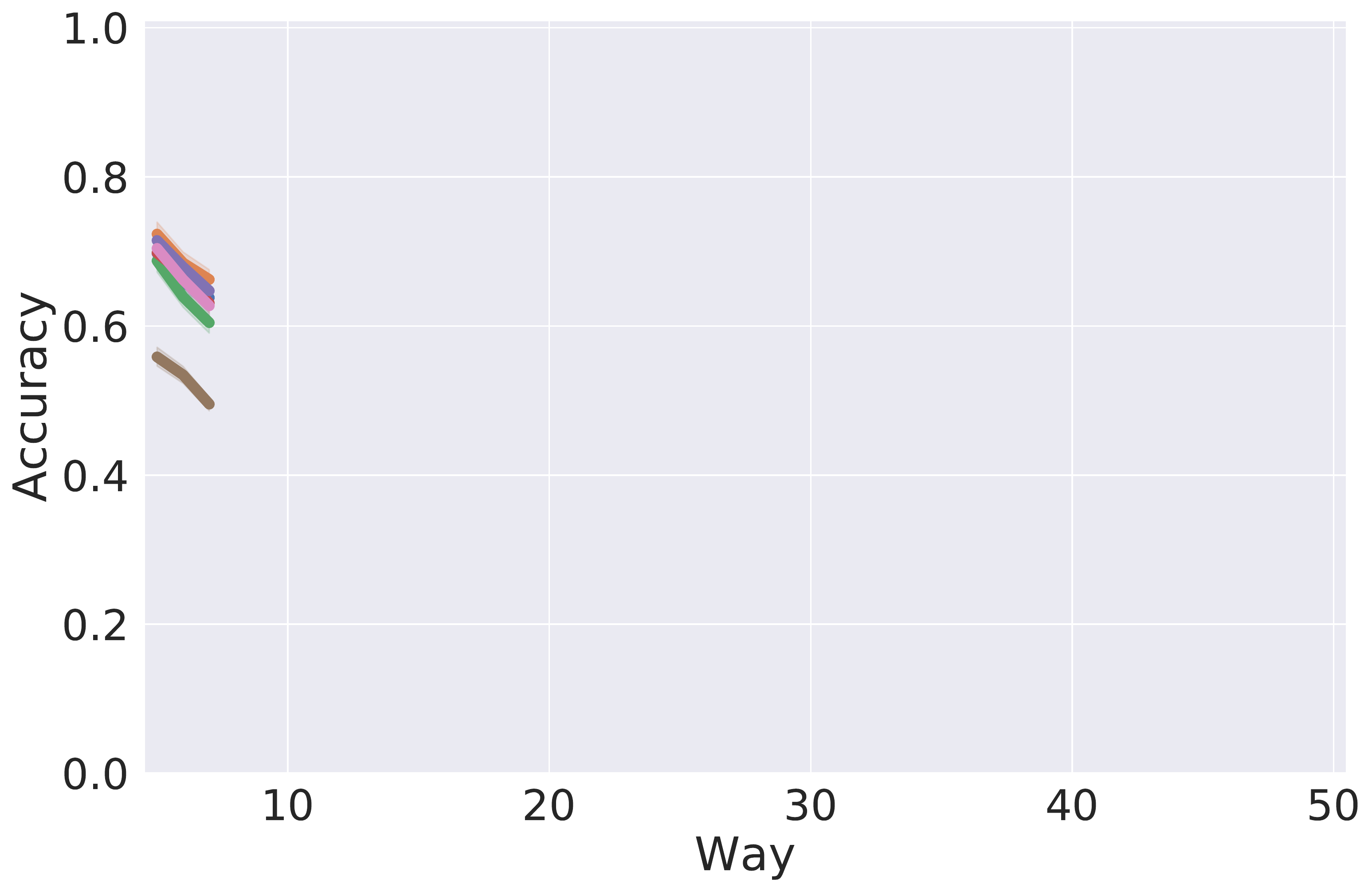}}\hspace{0.01cm}
\subfloat[Quickdraw]{\includegraphics[width=0.24\textwidth]{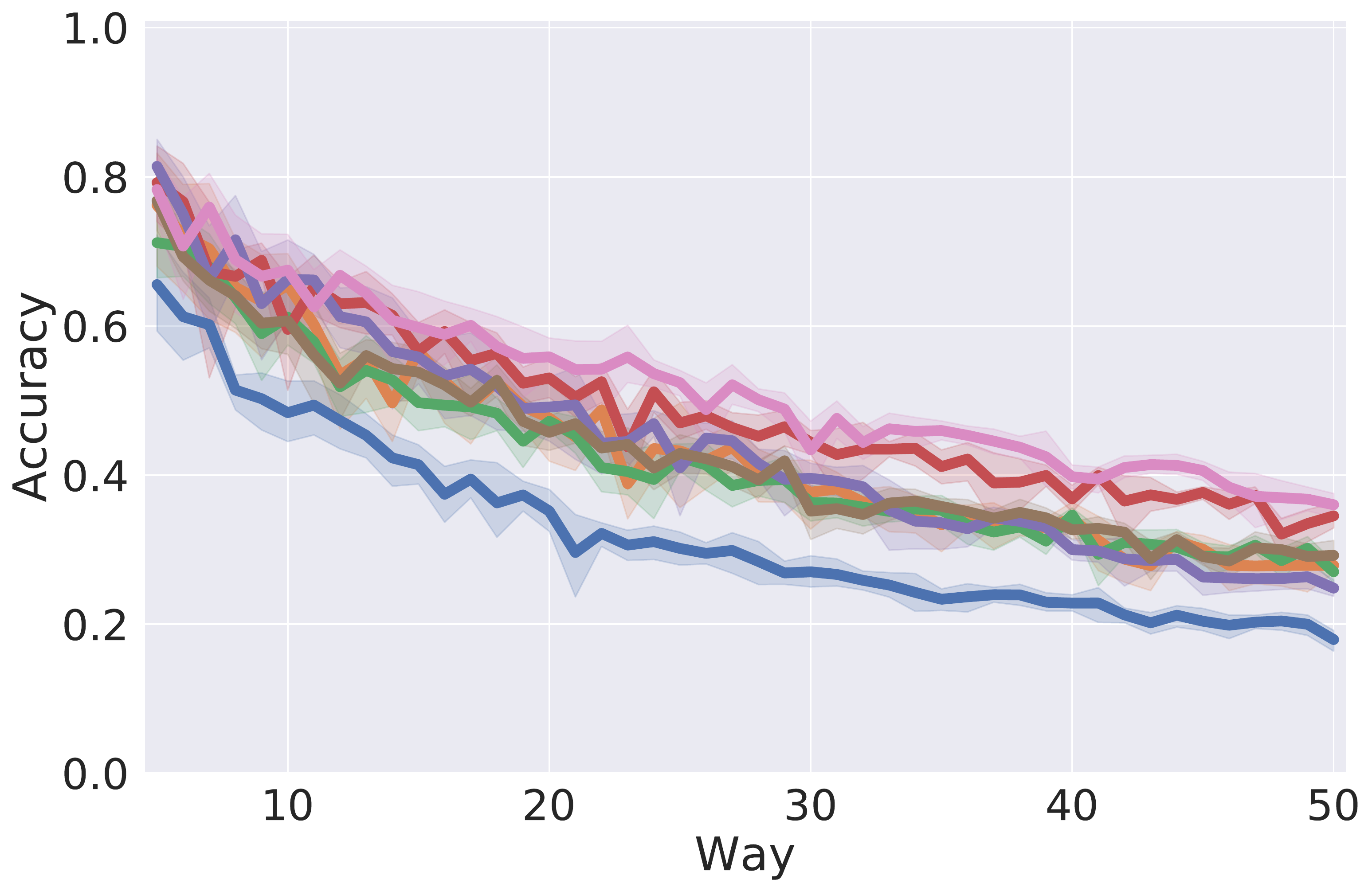}}\hspace{0.01cm}
\subfloat[Fungi]{\includegraphics[width=0.24\textwidth]{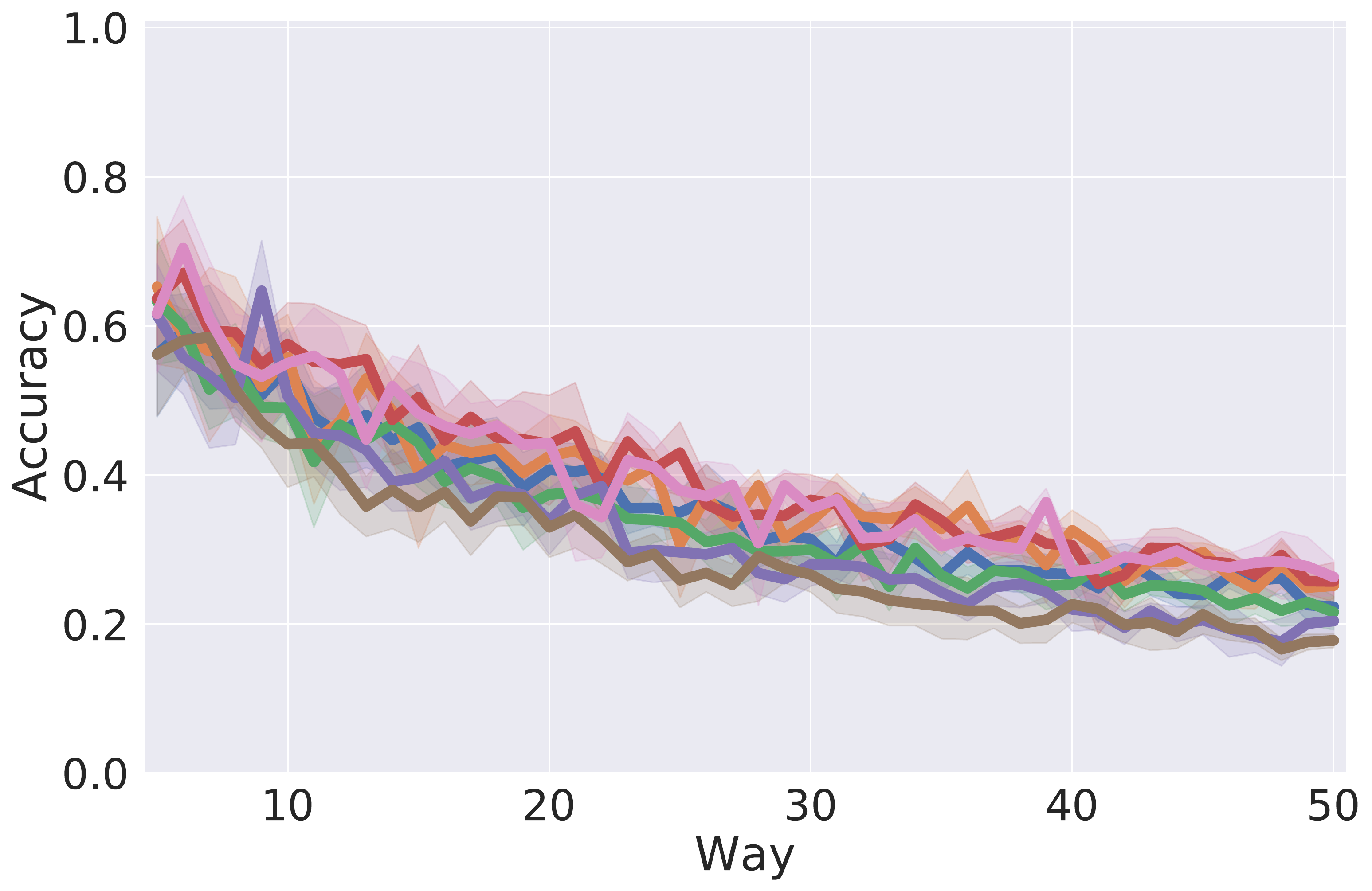}}\hspace{0.01cm}
\subfloat[VGG Flower]{\includegraphics[width=0.24\textwidth]{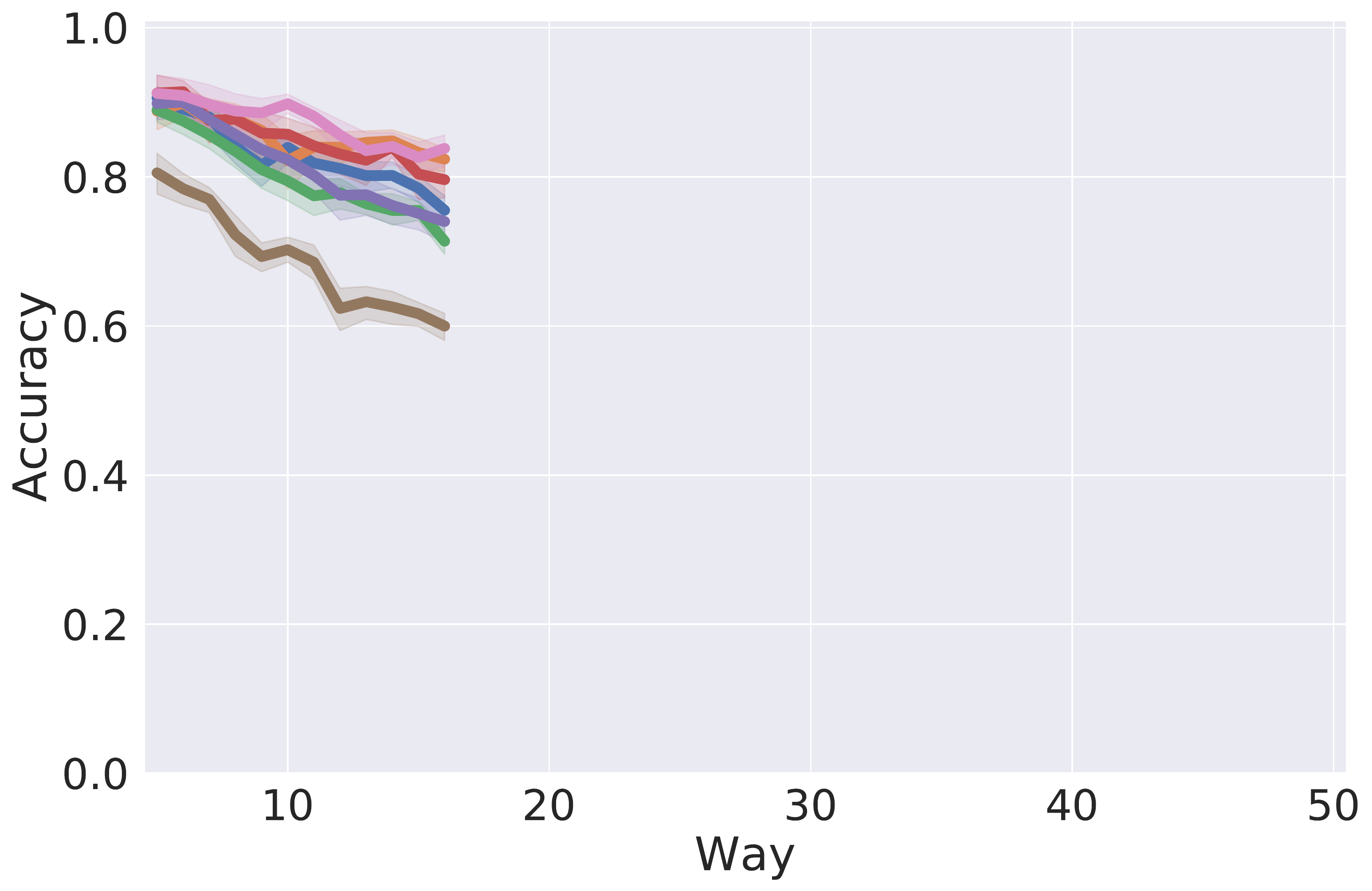}}\hspace{0.01cm}
\subfloat[Traffic Signs]{\includegraphics[width=0.24\textwidth]{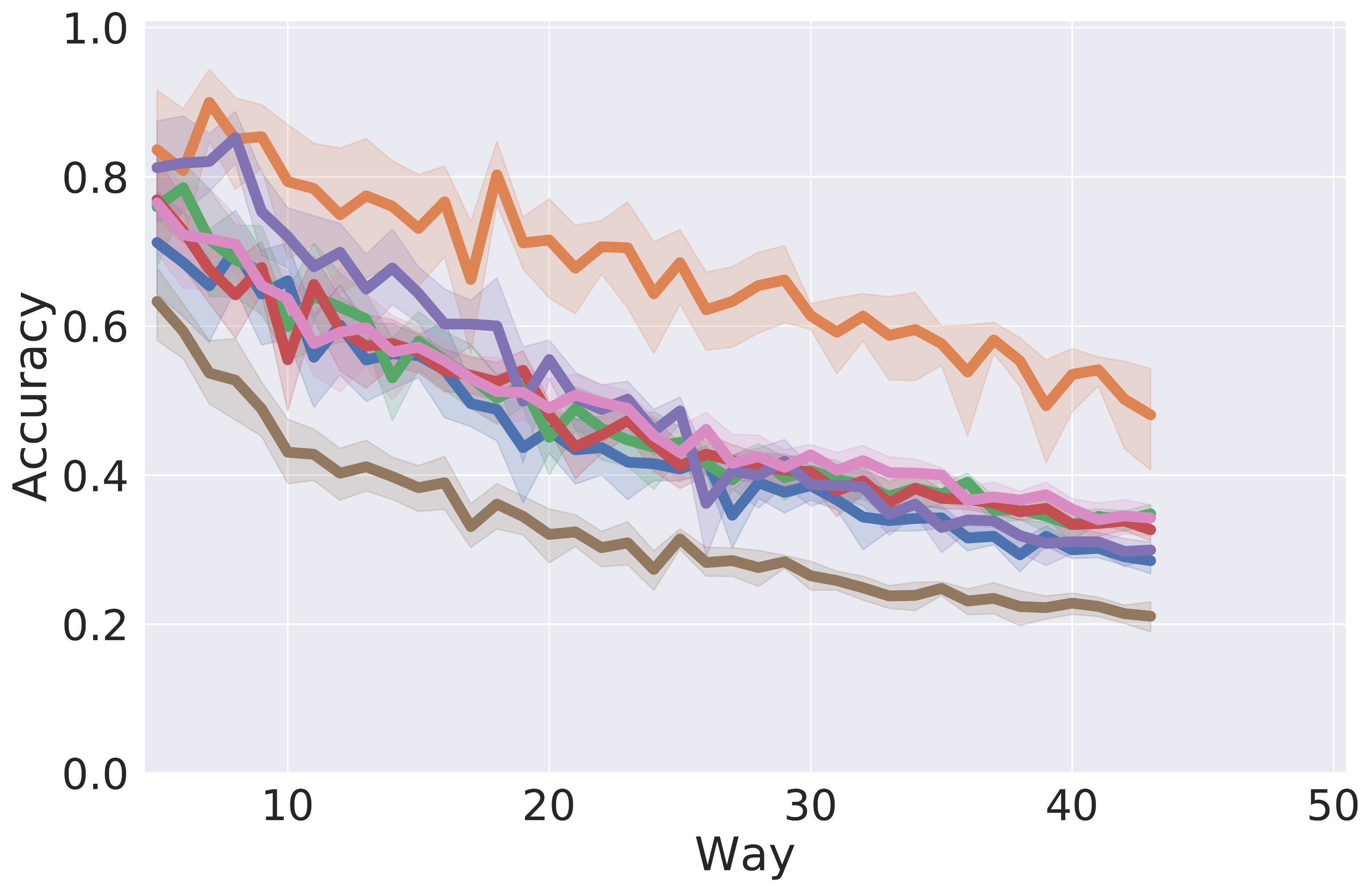}}\hspace{0.01cm}
\subfloat[MSCOCO]{\includegraphics[width=0.24\textwidth]{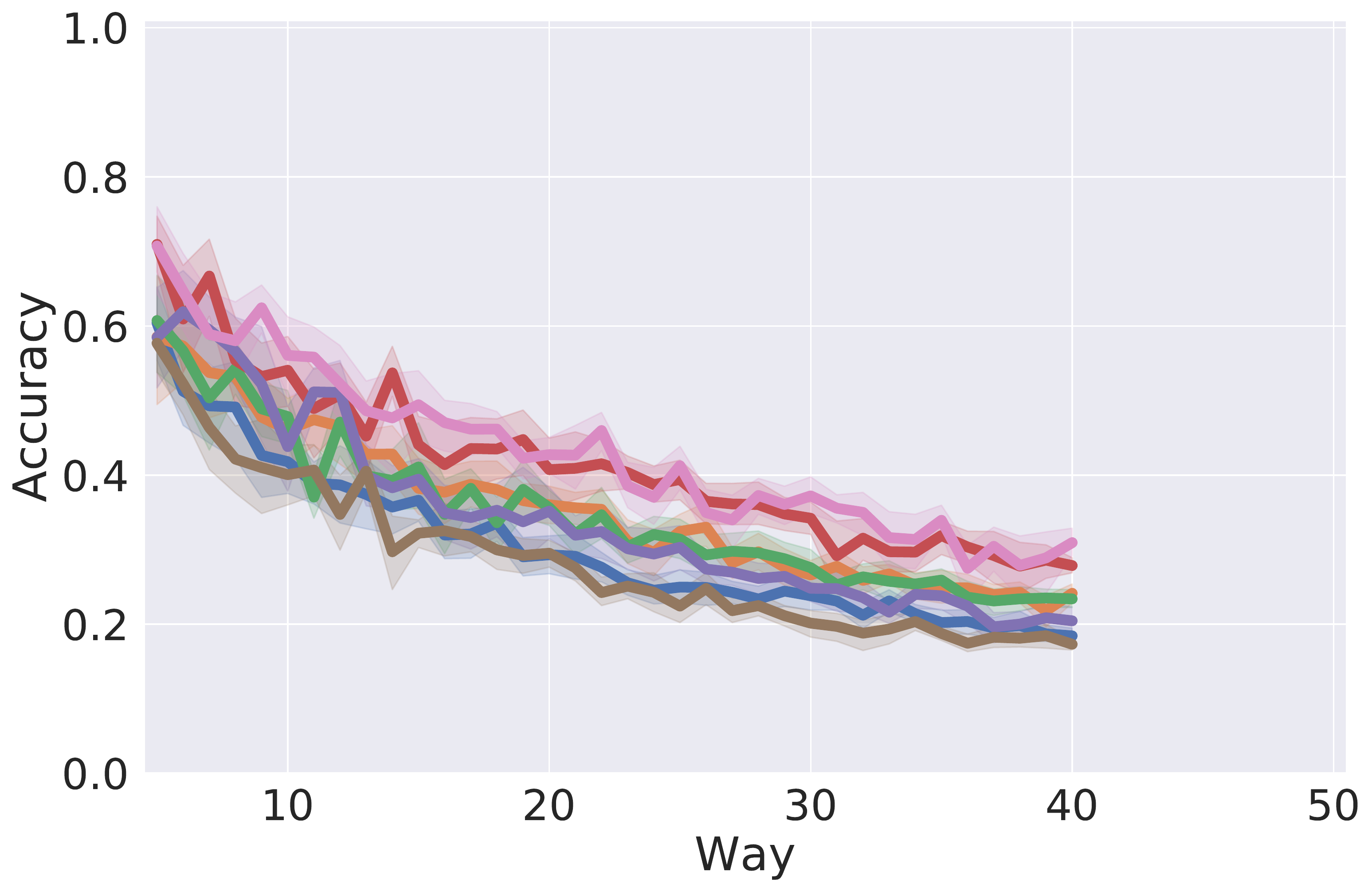}}\hspace{0.01cm}
\hspace{\fill}
\subfloat{\includegraphics[width=0.19\textwidth]{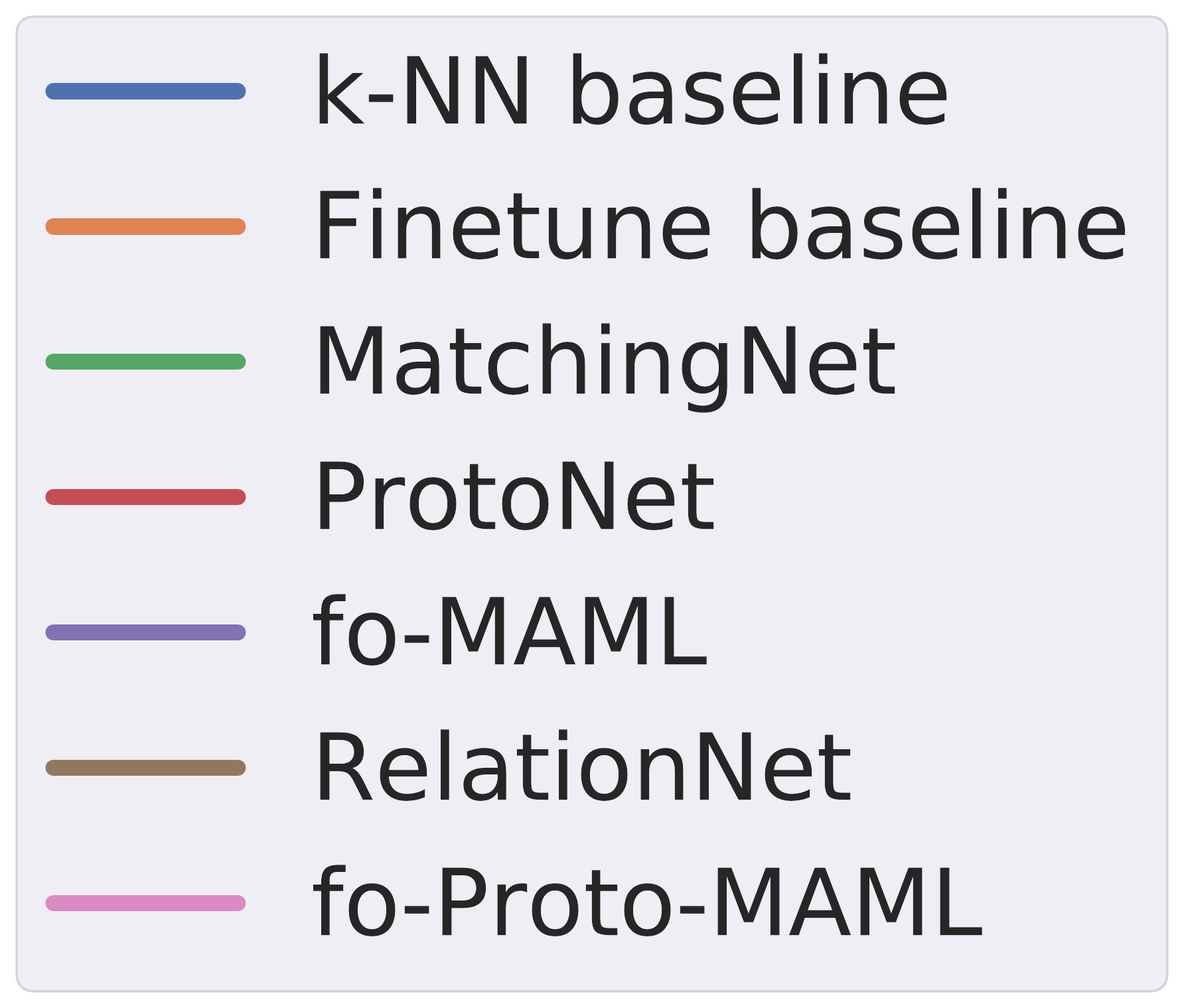}\hspace{0.01\textwidth}}\hspace{0.01cm}
   \caption{\label{fig:way_analysis_per_dataset} The performance across different ways, with 95\% confidence intervals, shown separately for each evaluation dataset. All models had been trained on ImageNet-only.}
 \end{figure*}

\begin{figure*}[htp]
\subfloat[ILSVRC]{\includegraphics[width=0.24\textwidth]{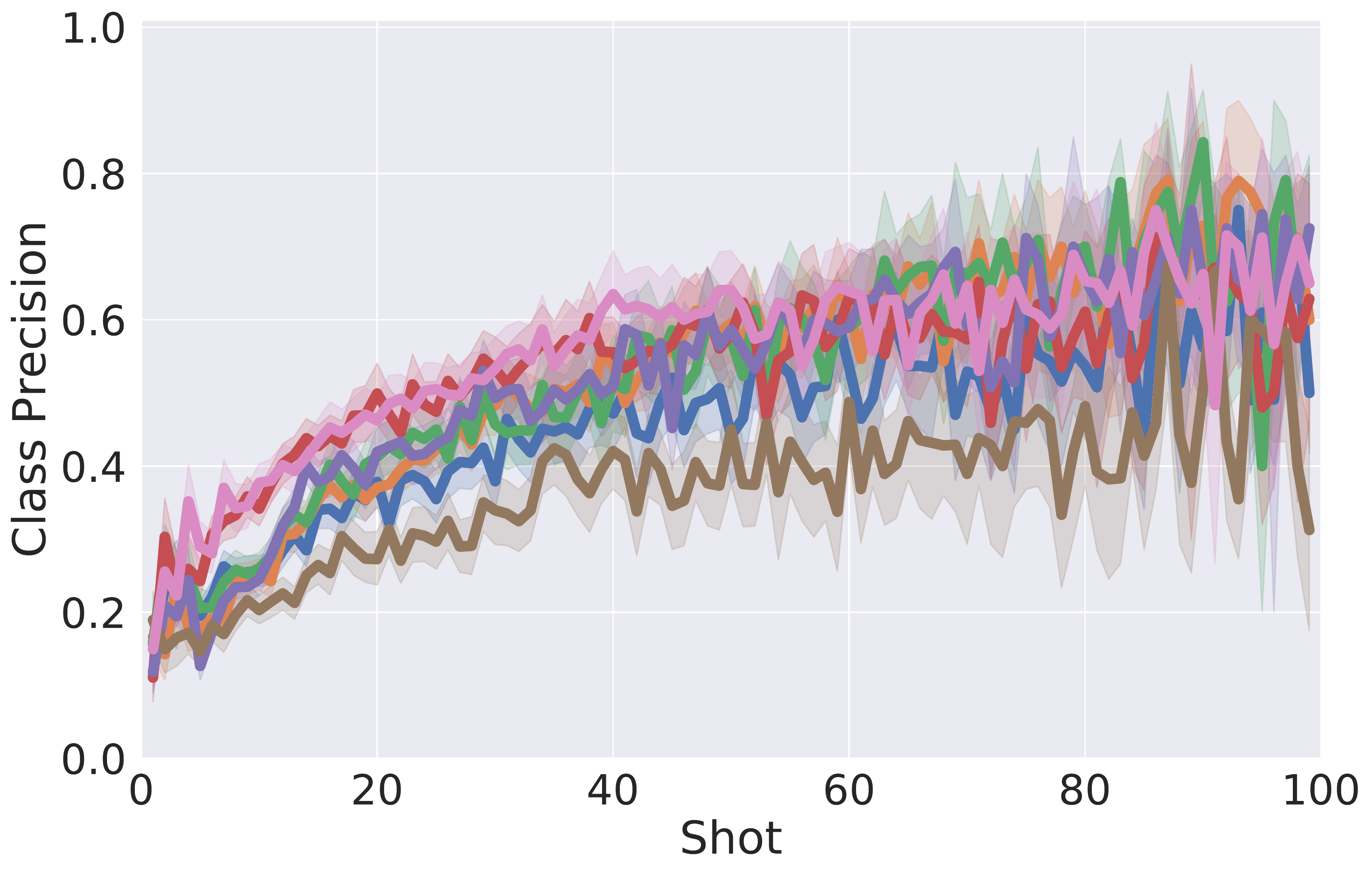}}\hspace{0.01cm}
\subfloat[Omniglot]{\includegraphics[width=0.24\textwidth]{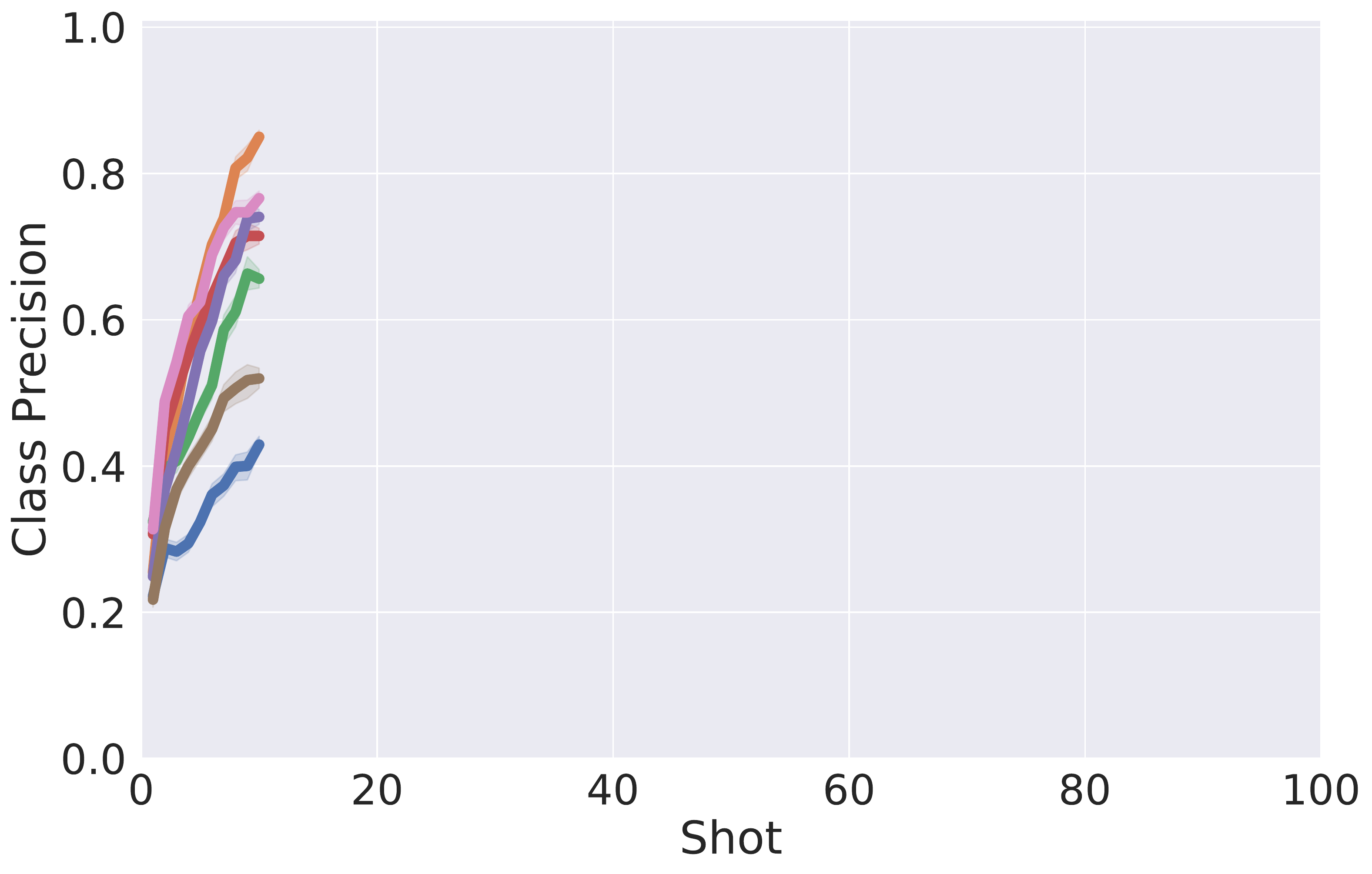}}\hspace{0.01cm}
\subfloat[Aircraft]{\includegraphics[width=0.24\textwidth]{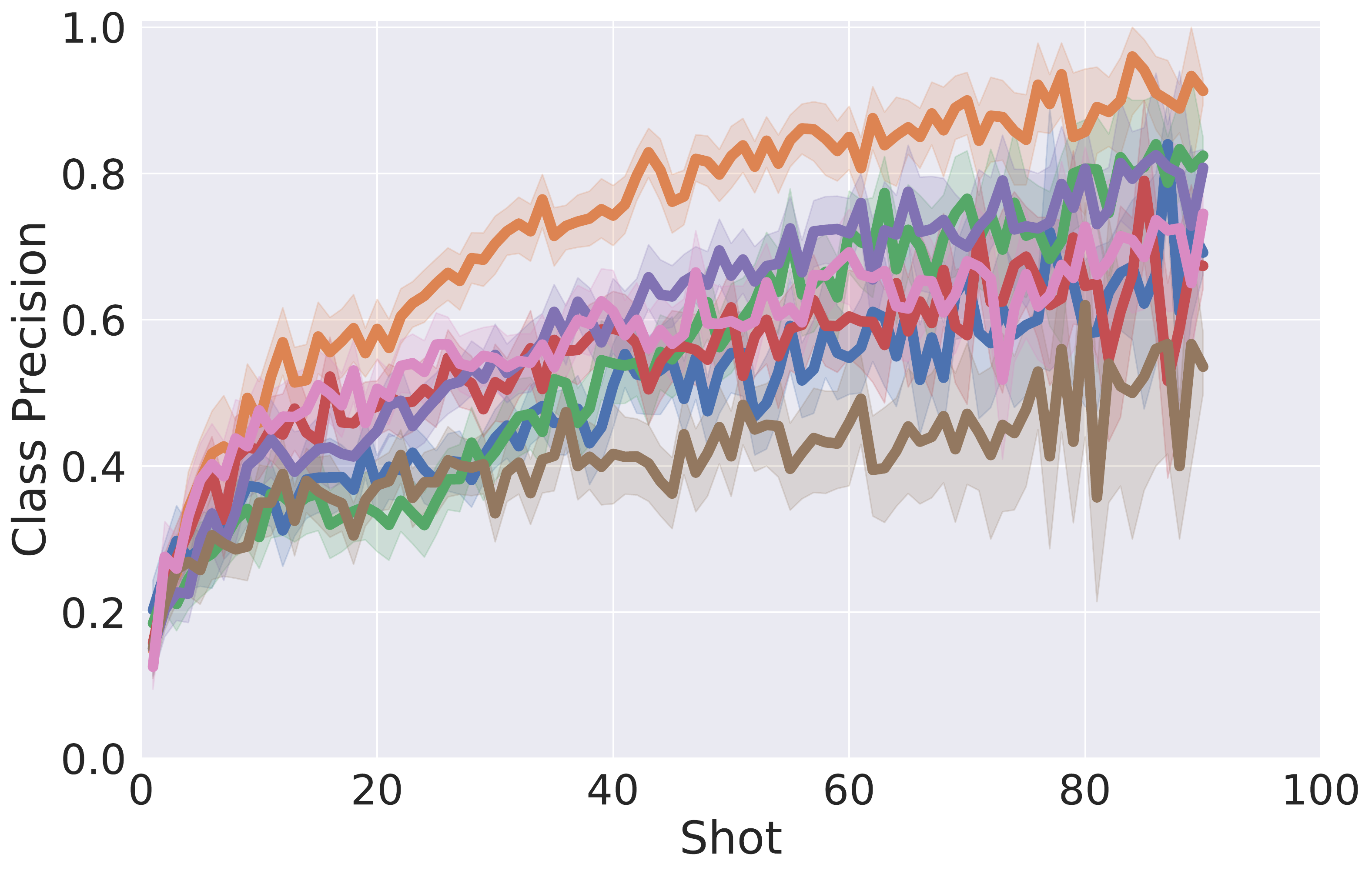}}\hspace{0.01cm}
\subfloat[Birds]{\includegraphics[width=0.24\textwidth]{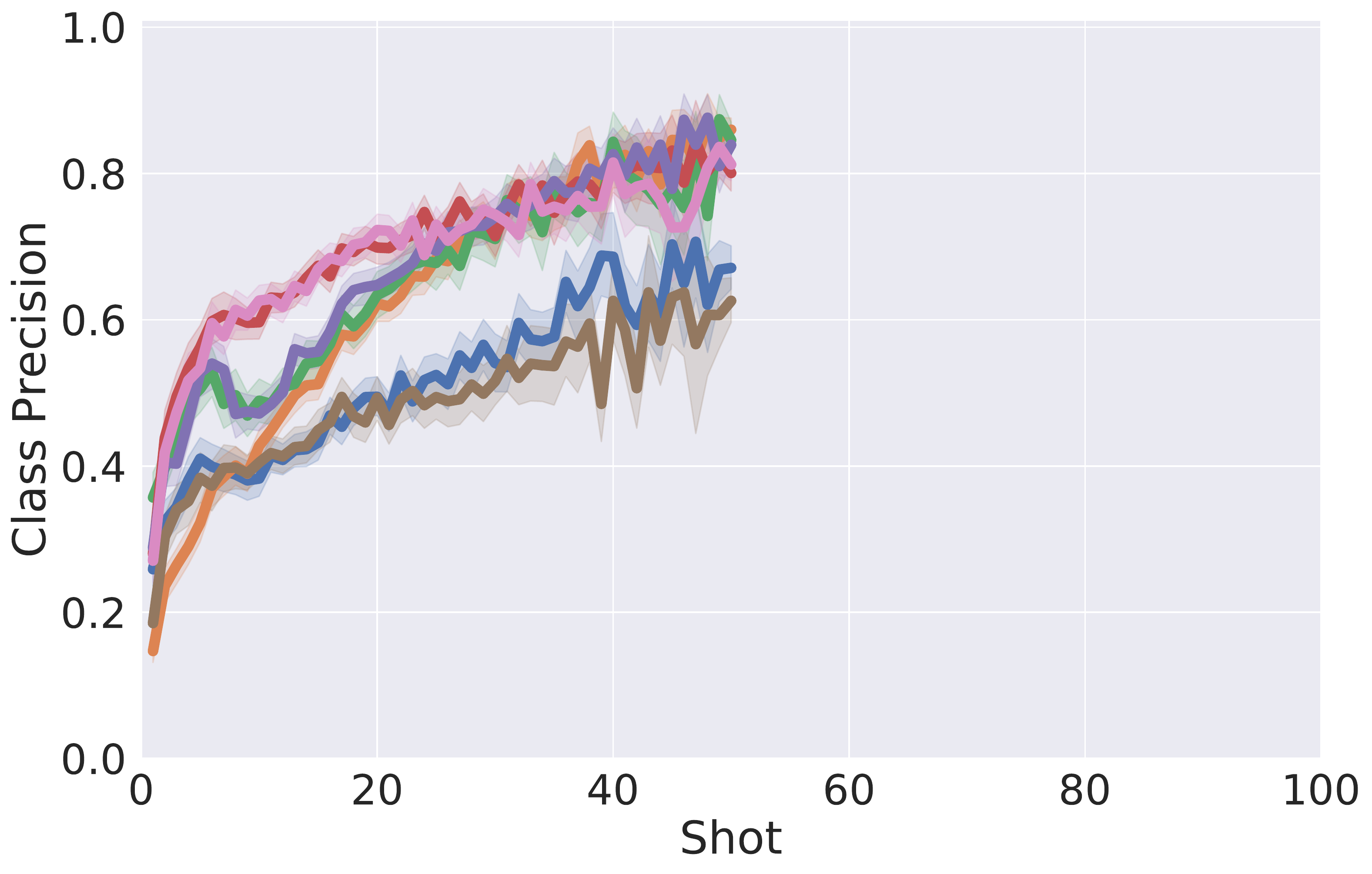}}\hspace{0.01cm}
\subfloat[Textures]{\includegraphics[width=0.24\textwidth]{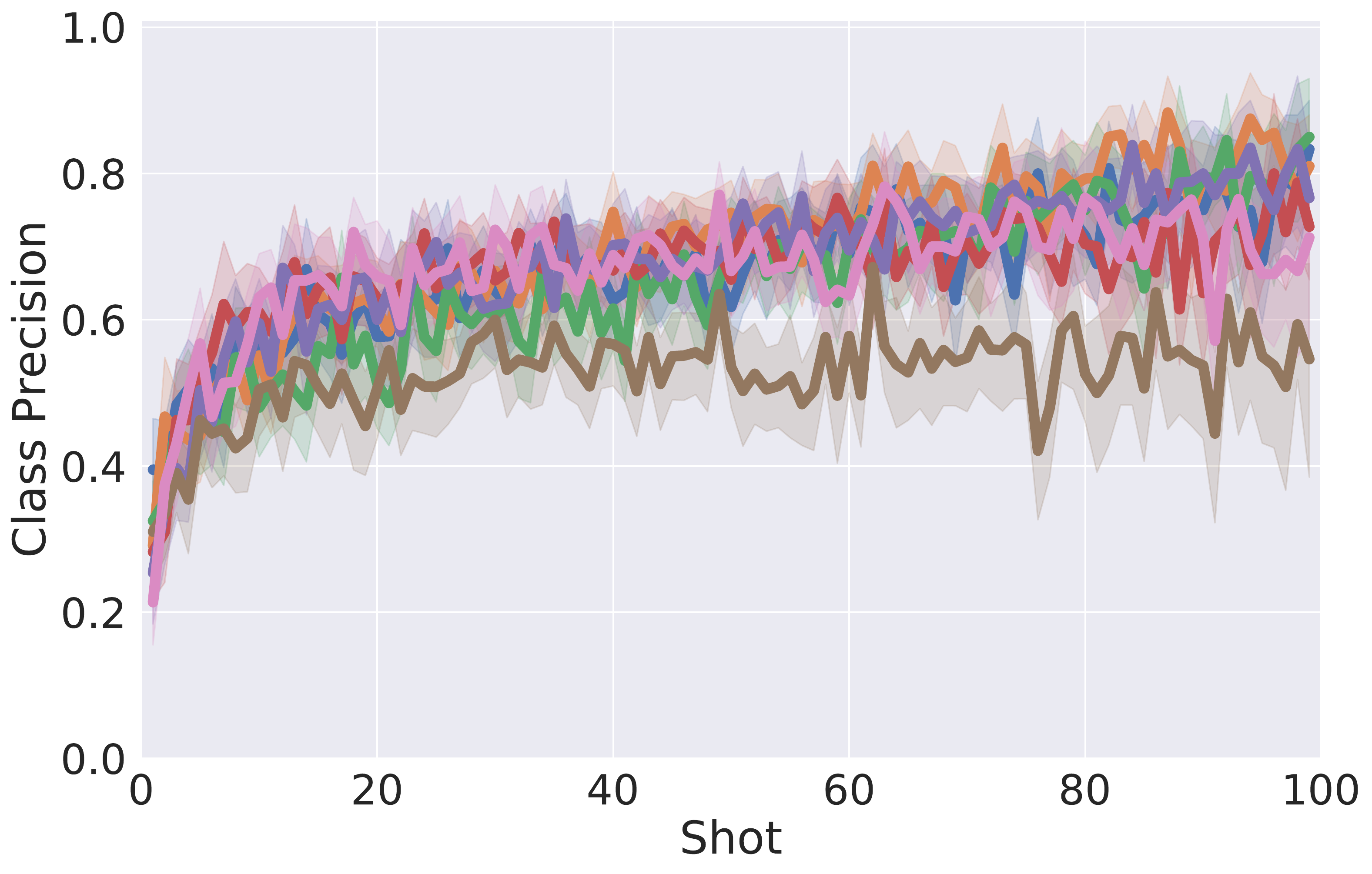}}\hspace{0.01cm}
\subfloat[Quickdraw]{\includegraphics[width=0.24\textwidth]{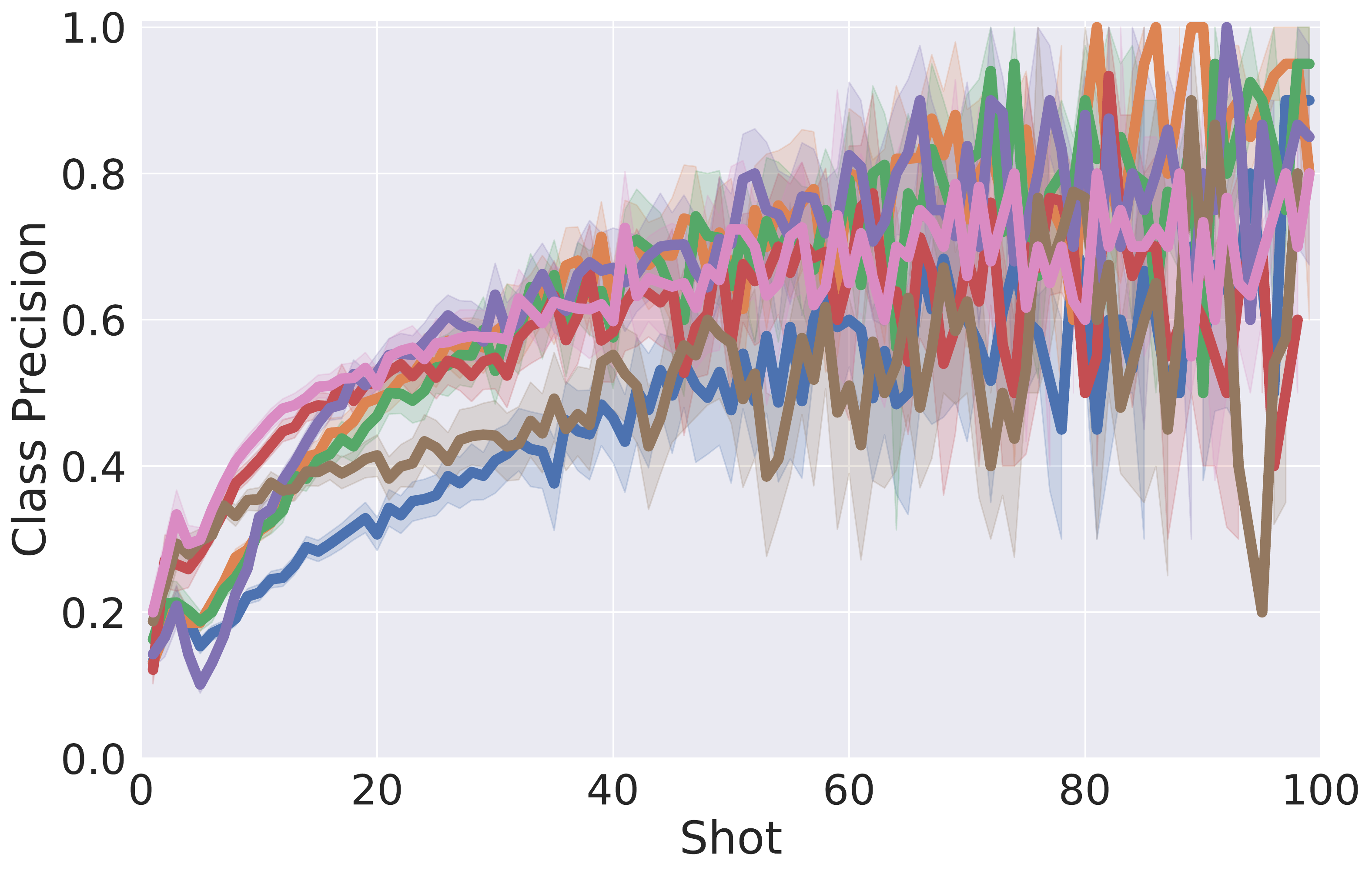}}\hspace{0.01cm}
\subfloat[Fungi]{\includegraphics[width=0.24\textwidth]{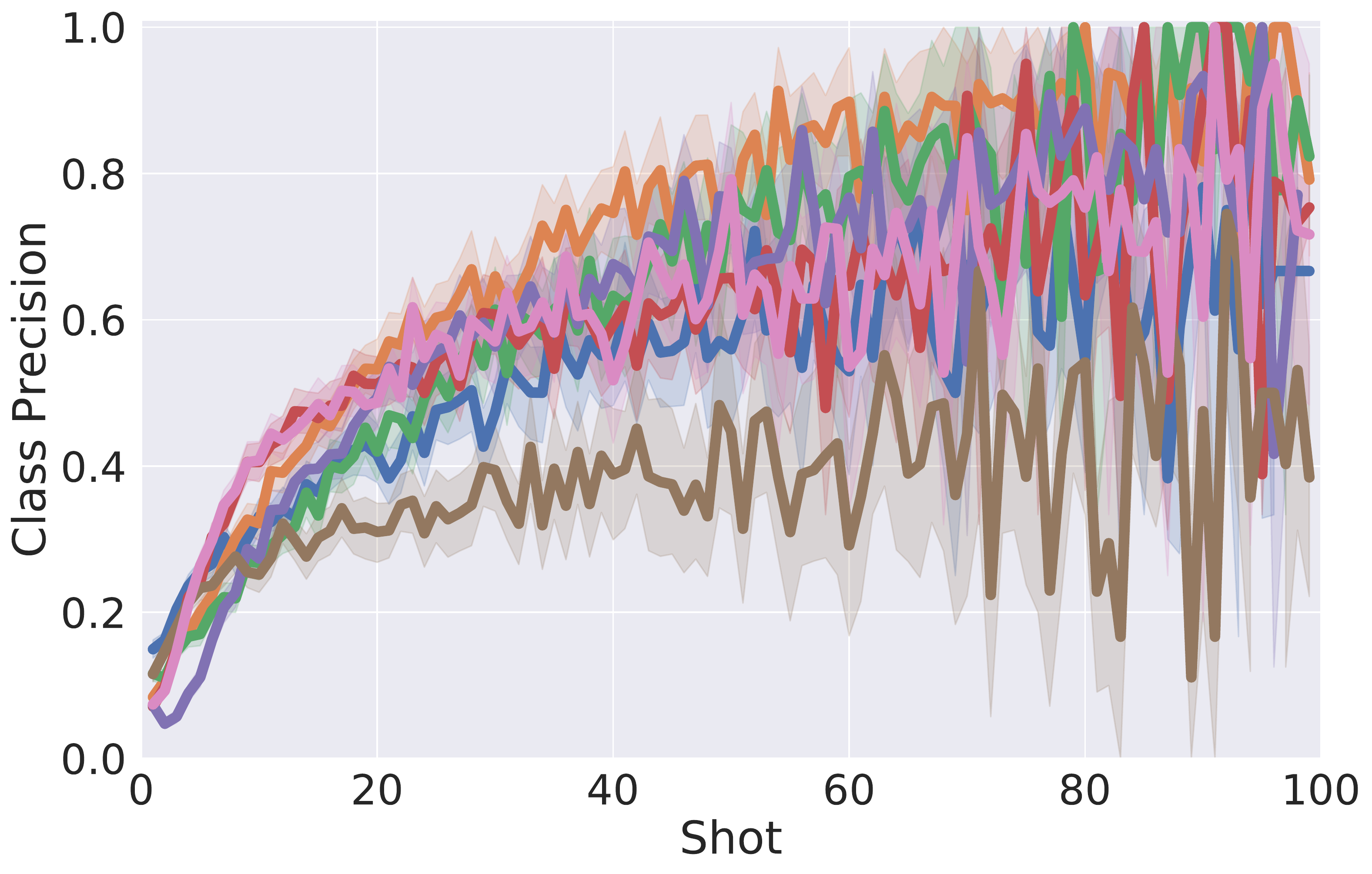}}\hspace{0.01cm}
\subfloat[VGG Flower]{\includegraphics[width=0.24\textwidth]{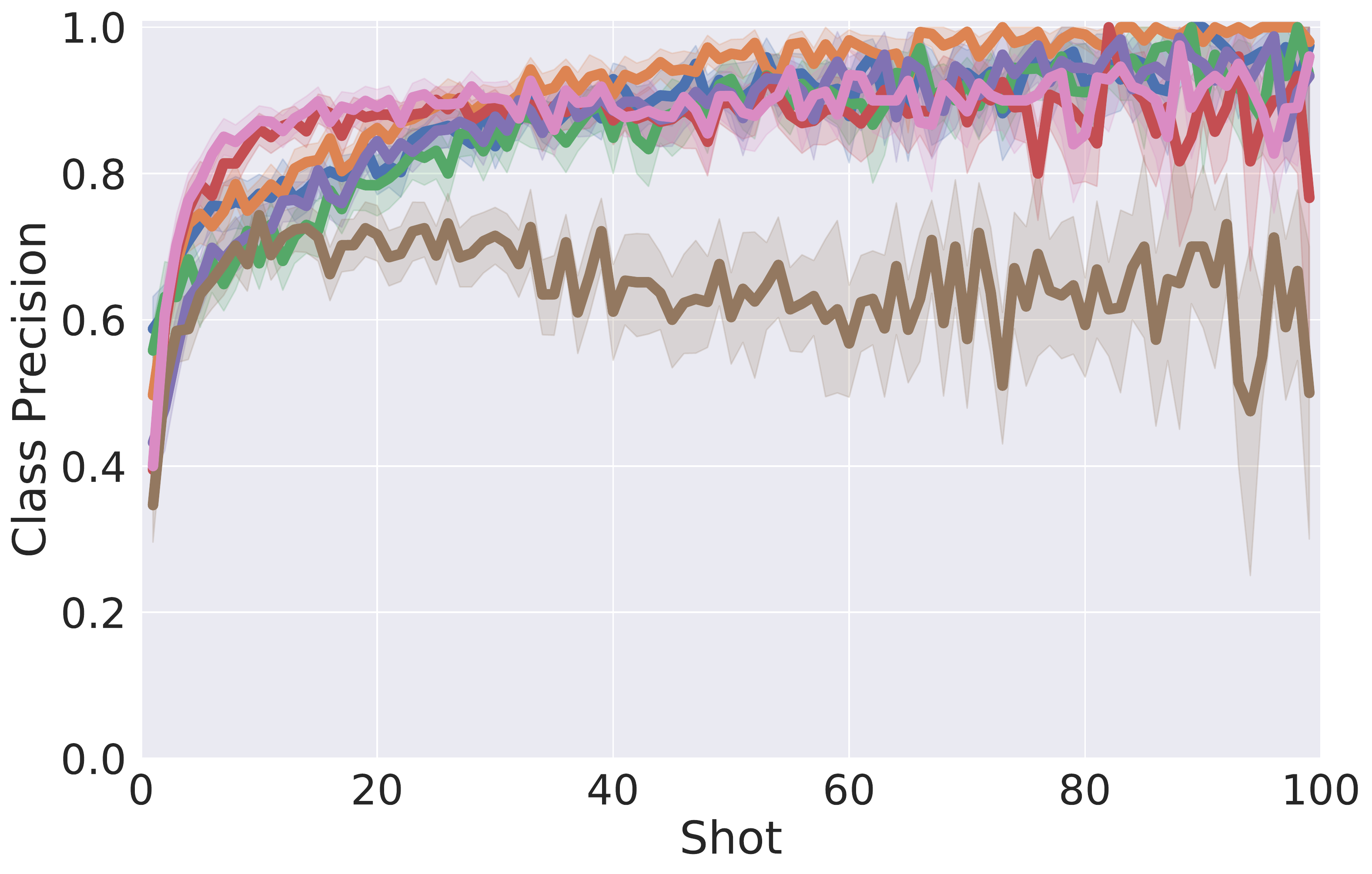}}\hspace{0.01cm}
\subfloat[Traffic Signs]{\includegraphics[width=0.24\textwidth]{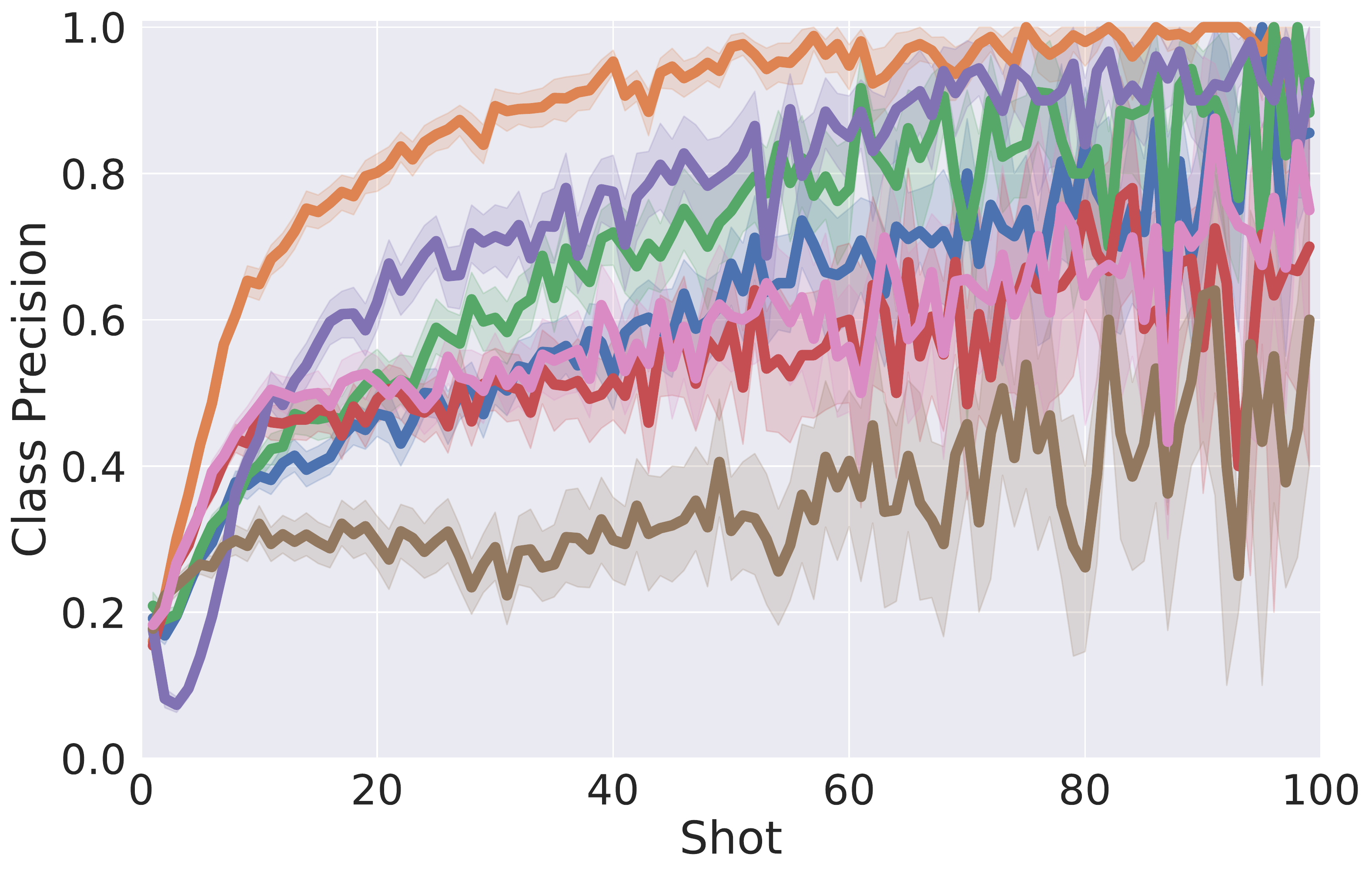}}\hspace{0.01cm}
\subfloat[MSCOCO]{\includegraphics[width=0.24\textwidth]{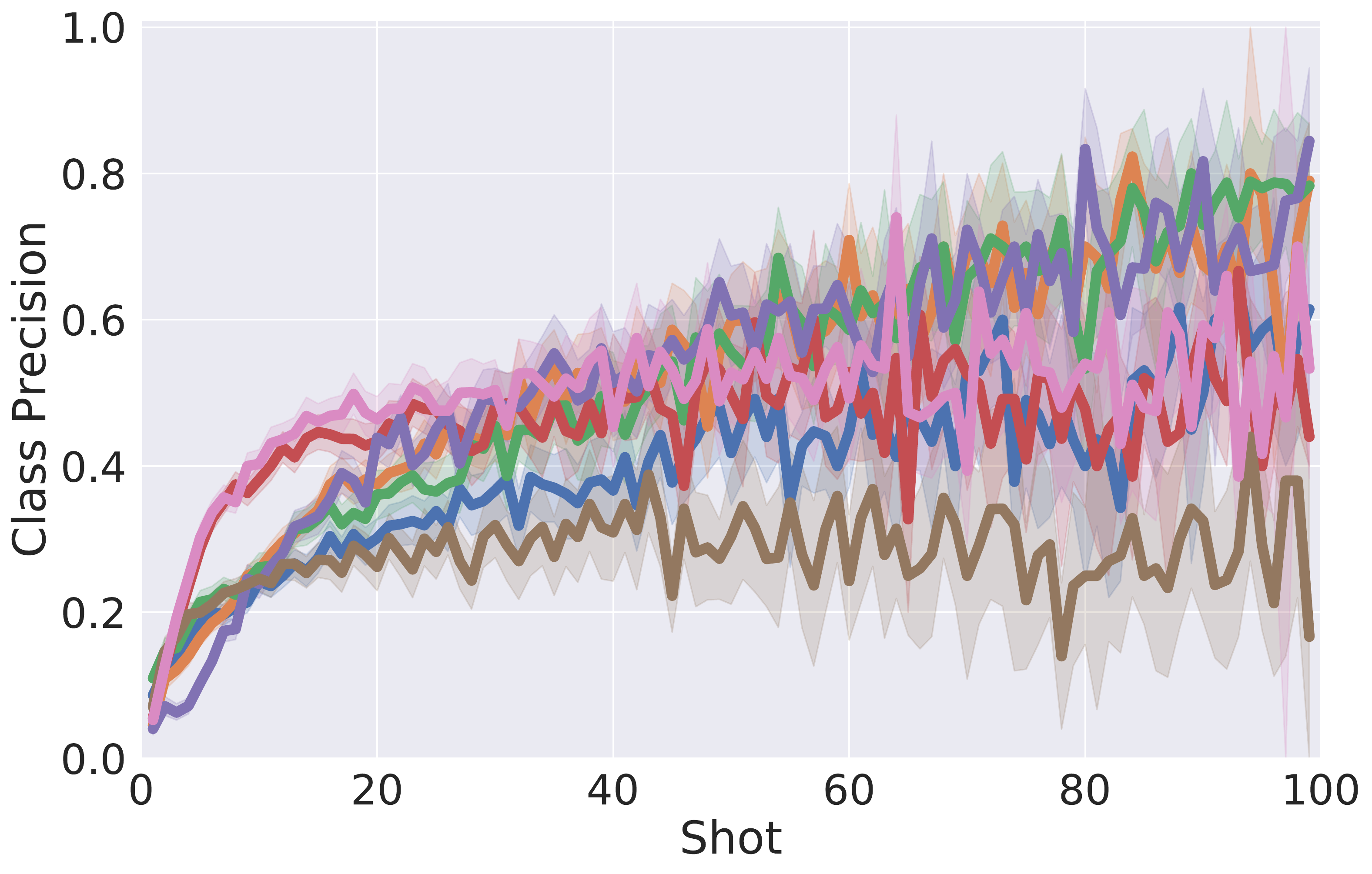}}\hspace{0.01cm}
\hspace{\fill}
\subfloat{\includegraphics[width=0.19\textwidth]{figures_iclr2020/ways_shots_legend.pdf}\hspace{0.01\textwidth}}\hspace{0.01cm}
   \caption{\label{fig:shots_analysis_per_dataset} The performance across different shots, with 95\% confidence intervals, shown separately for each evaluation dataset. All models had been trained on ImageNet-only.}
 \end{figure*}

\begin{figure*}[htp]
\subfloat[ILSVRC]{\includegraphics[width=0.24\textwidth]{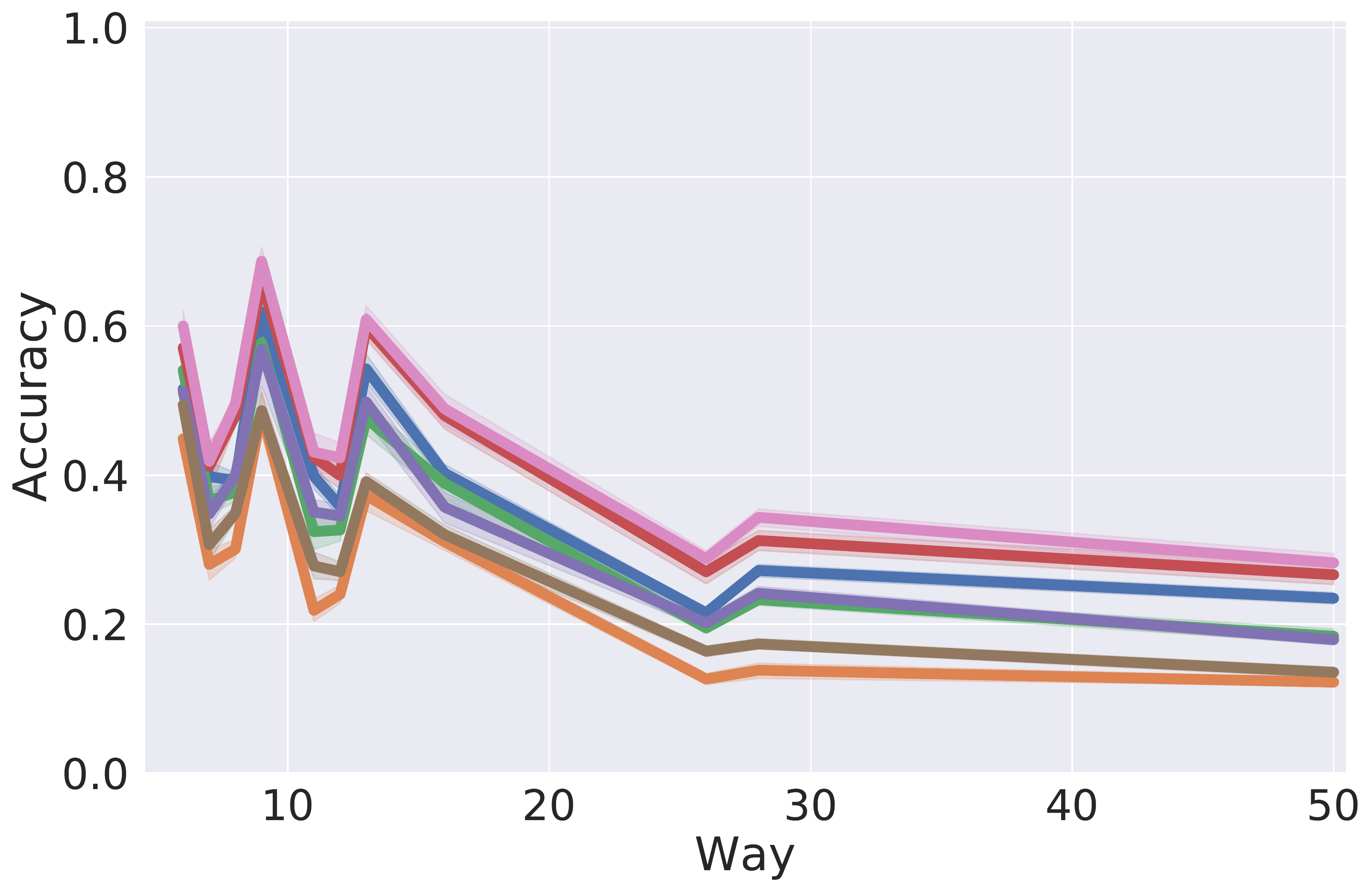}}\hspace{0.01cm}
\subfloat[Omniglot]{\includegraphics[width=0.24\textwidth]{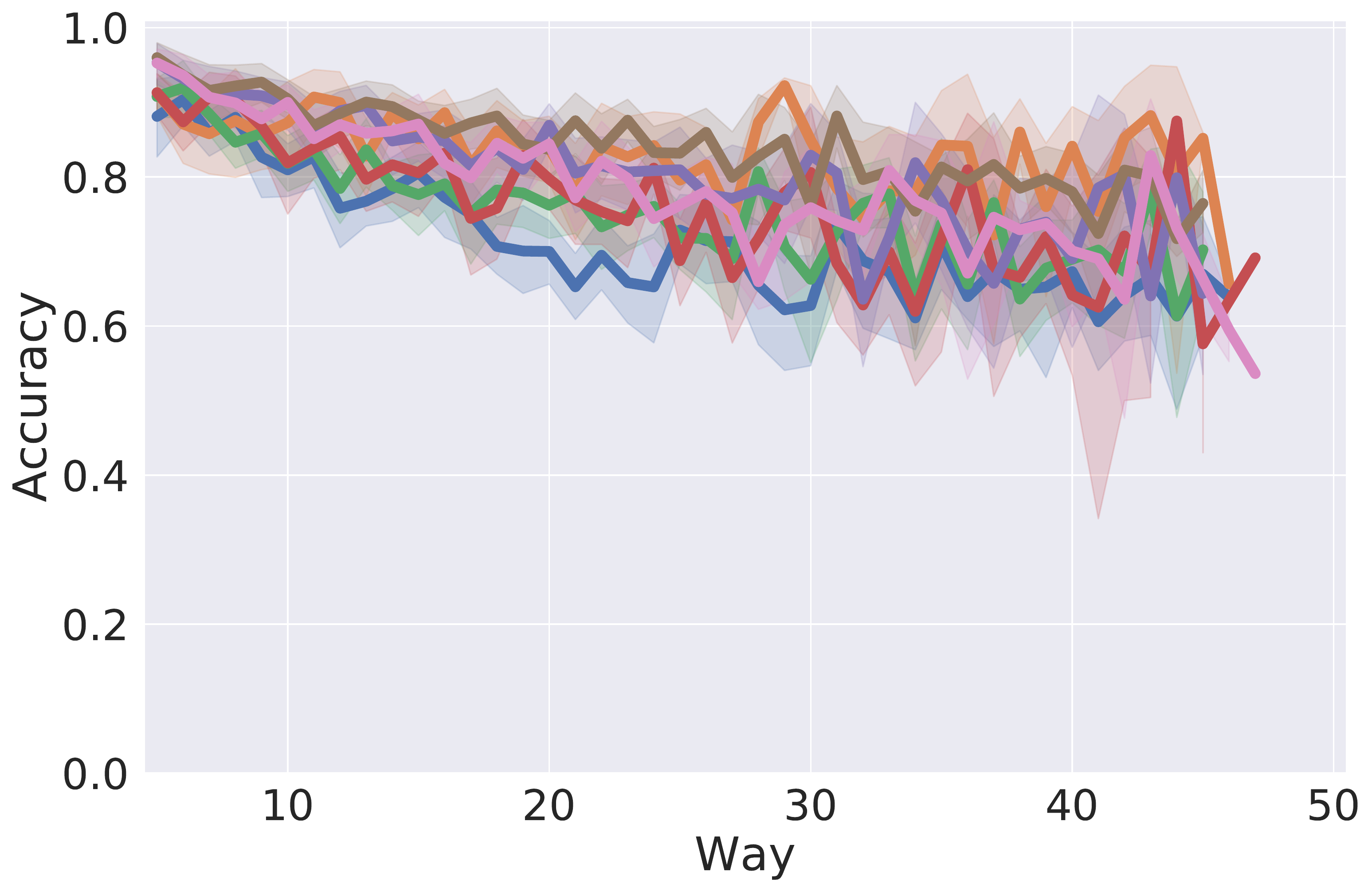}}\hspace{0.01cm}
\subfloat[Aircraft]{\includegraphics[width=0.24\textwidth]{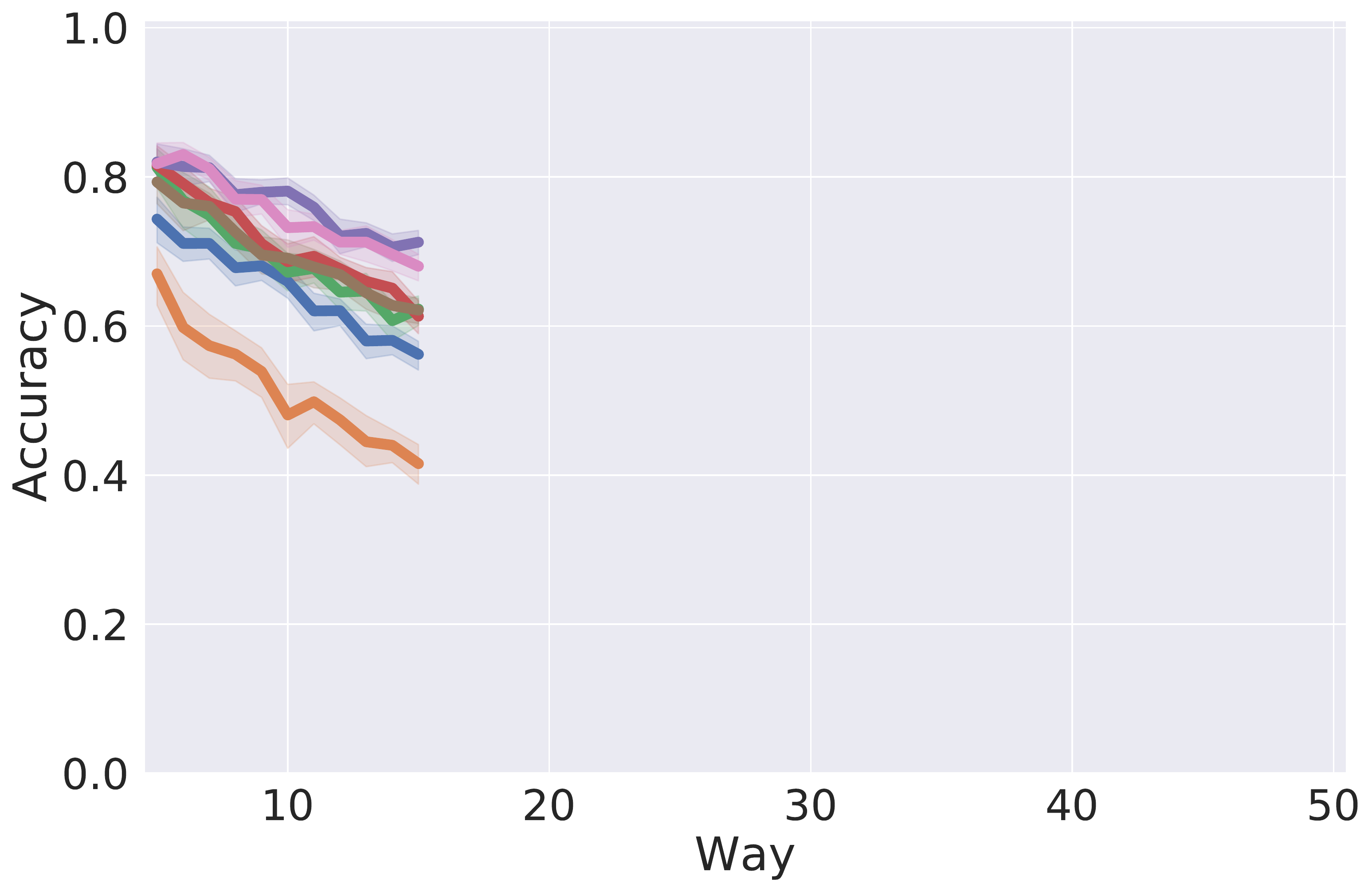}}\hspace{0.01cm}
\subfloat[Birds]{\includegraphics[width=0.24\textwidth]{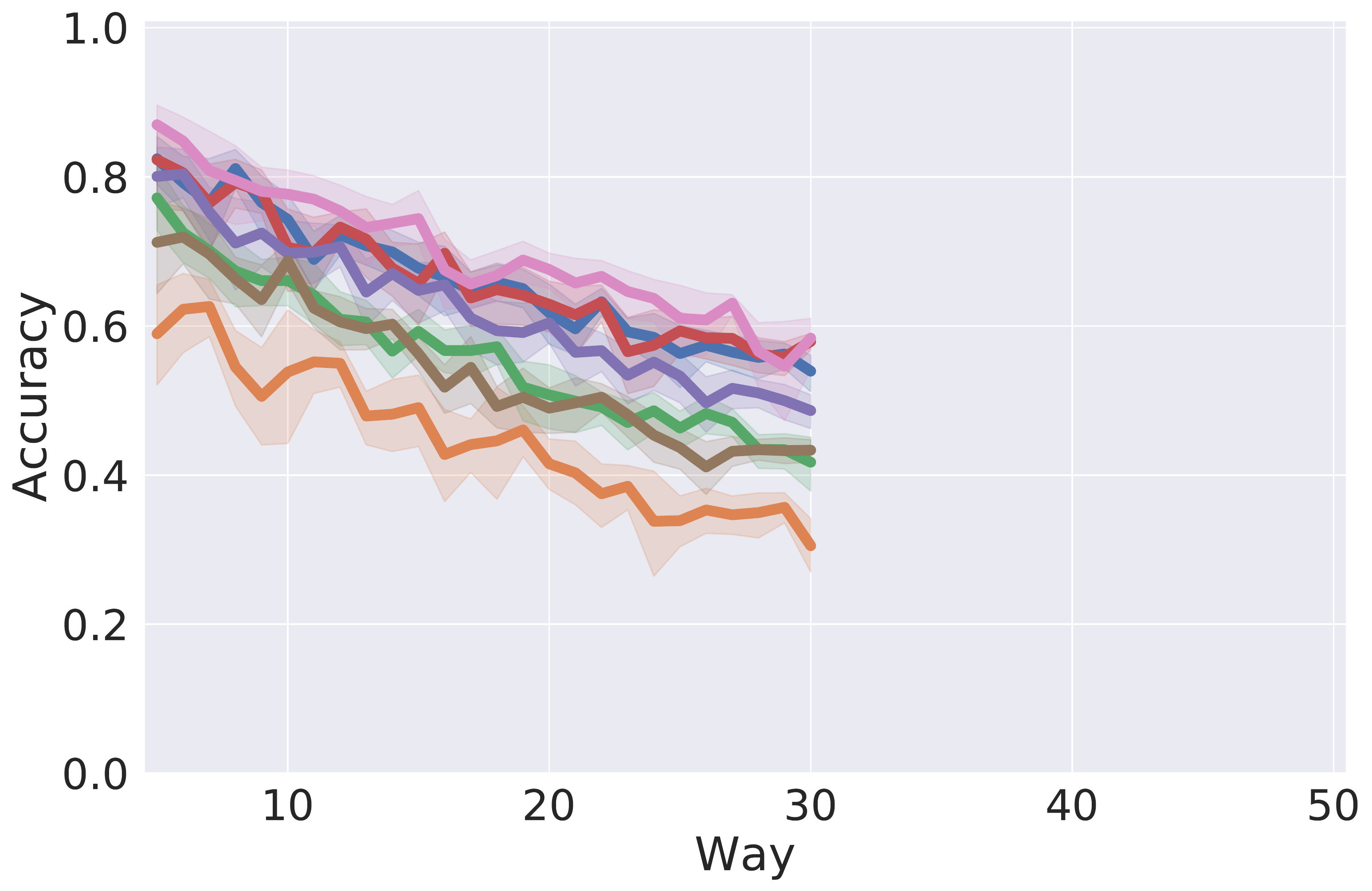}}\hspace{0.01cm}
\subfloat[Textures]{\includegraphics[width=0.24\textwidth]{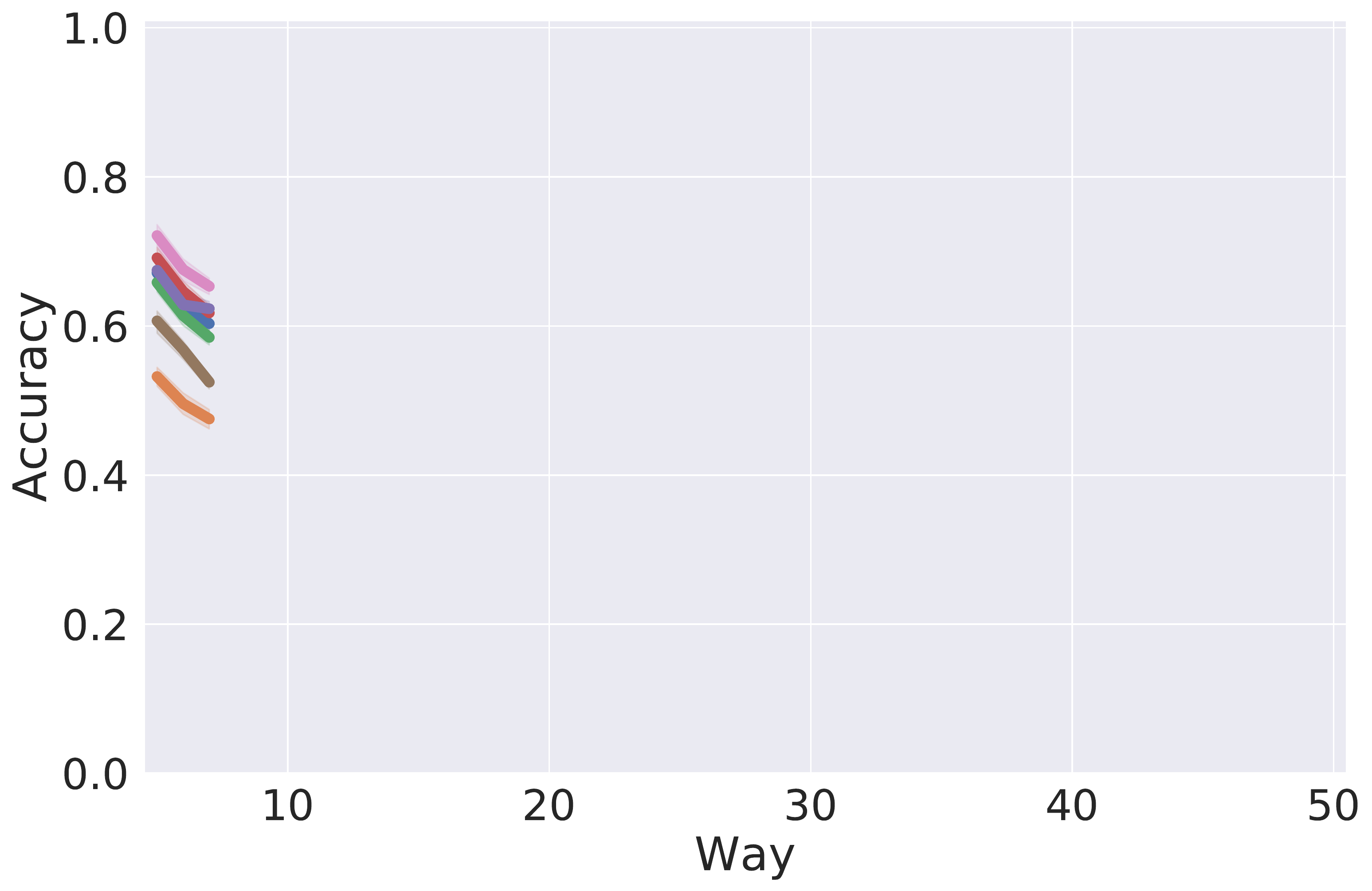}}\hspace{0.01cm}
\subfloat[Quickdraw]{\includegraphics[width=0.24\textwidth]{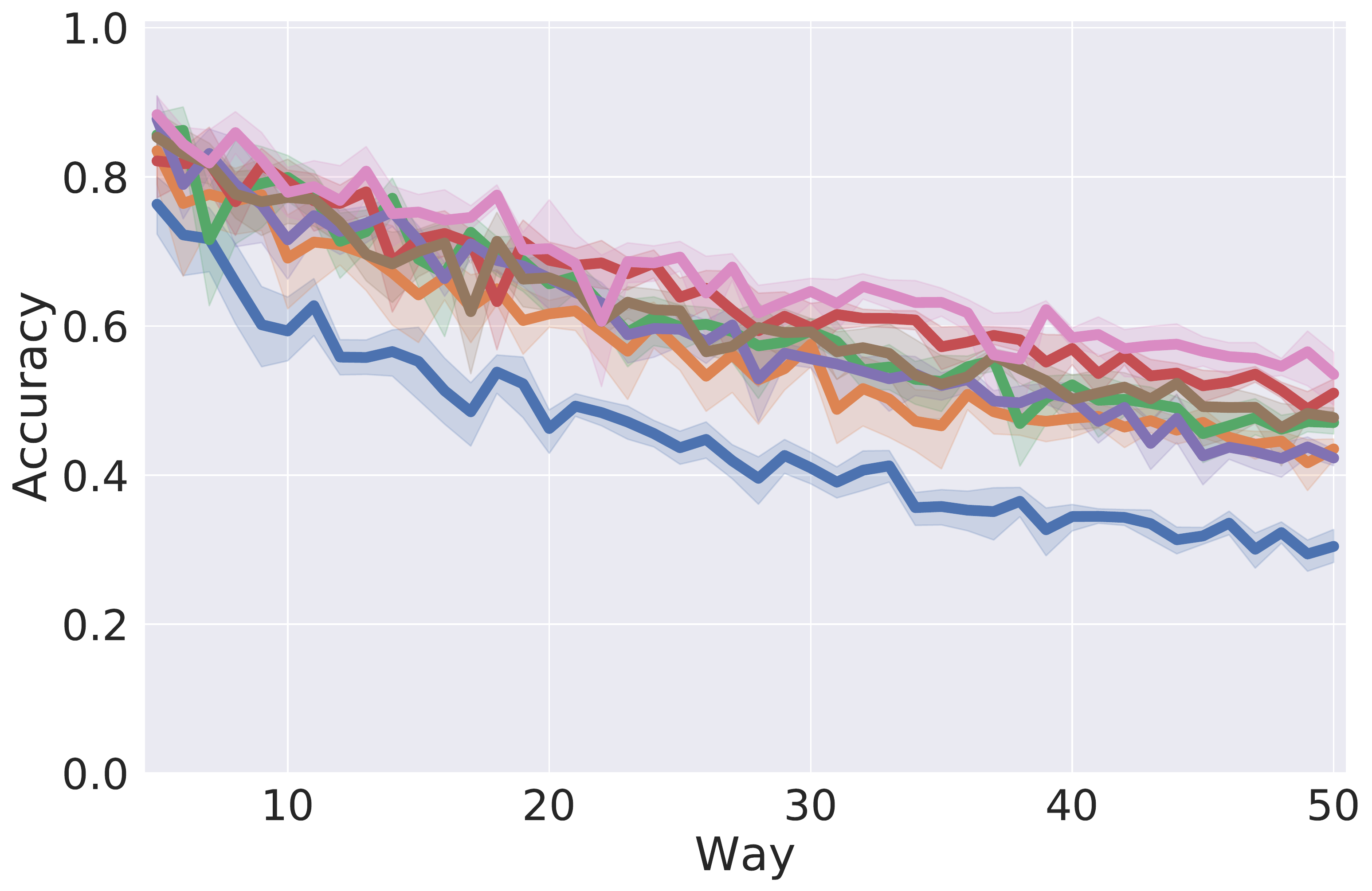}}\hspace{0.01cm}
\subfloat[Fungi]{\includegraphics[width=0.24\textwidth]{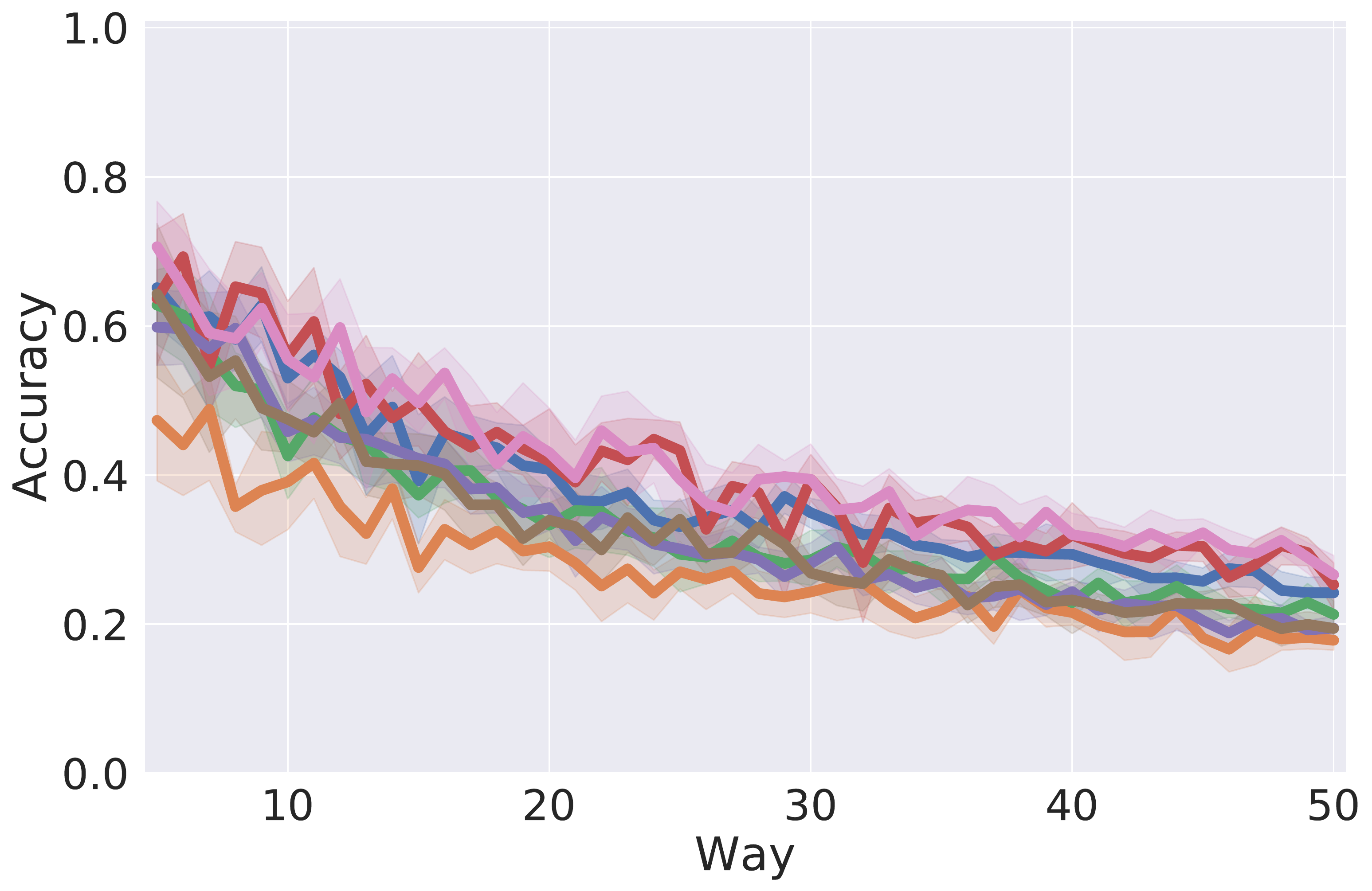}}\hspace{0.01cm}
\subfloat[VGG Flower]{\includegraphics[width=0.24\textwidth]{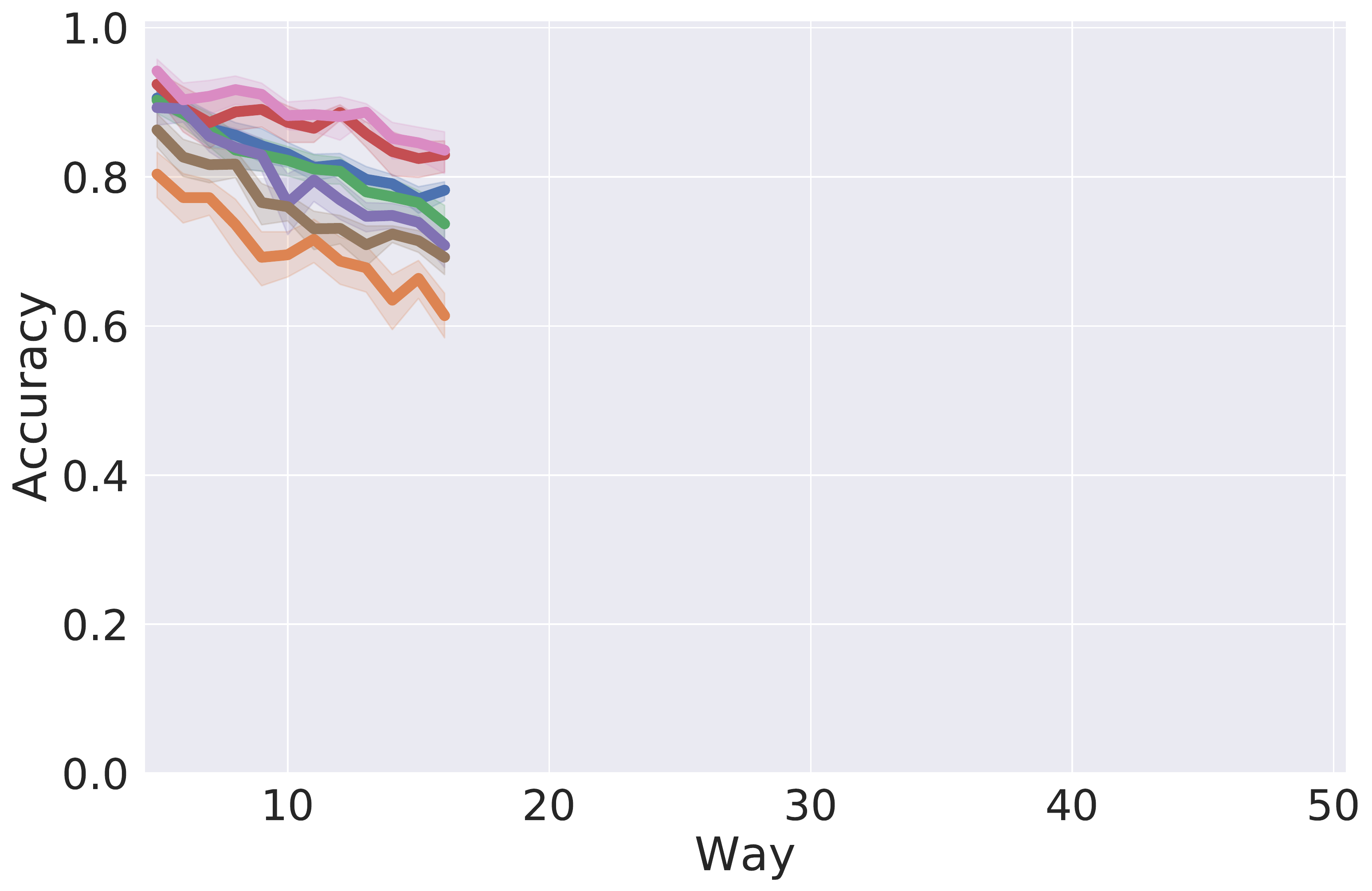}}\hspace{0.01cm}
\subfloat[Traffic Signs]{\includegraphics[width=0.24\textwidth]{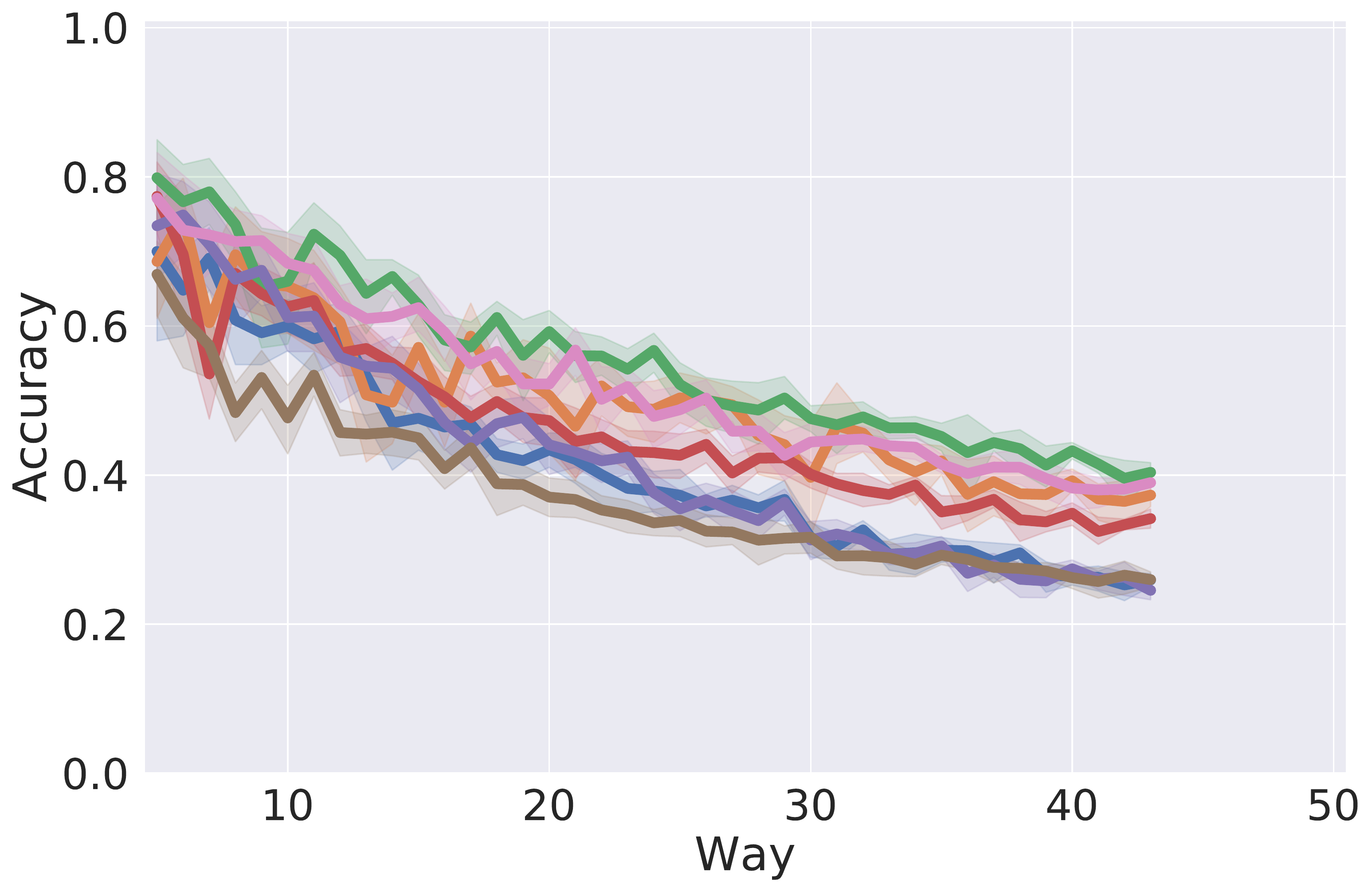}}\hspace{0.01cm}
\subfloat[MSCOCO]{\includegraphics[width=0.24\textwidth]{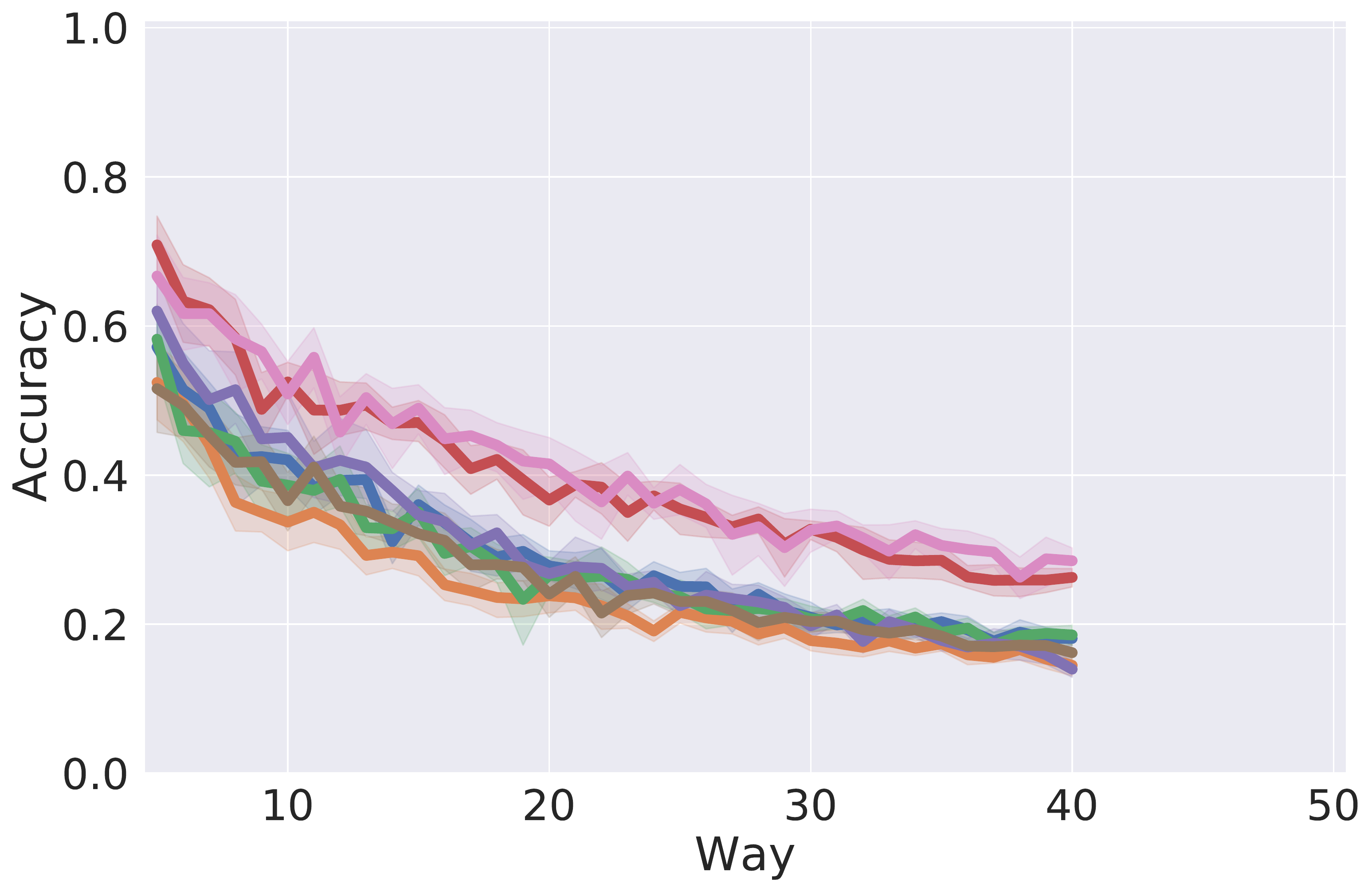}}\hspace{0.01cm}
\hspace{\fill}
\subfloat{\includegraphics[width=0.19\textwidth]{figures_iclr2020/ways_shots_legend.pdf}\hspace{0.01\textwidth}}\hspace{0.01cm}
	\caption{\label{fig:way_analysis_per_dataset_all_datasets} The performance across different ways, with 95\% confidence intervals, shown separately for each evaluation dataset. All models had been trained on (the training splits of) all datasets.}
 \end{figure*}

\begin{figure*}[htp]
\subfloat[ILSVRC]{\includegraphics[width=0.24\textwidth]{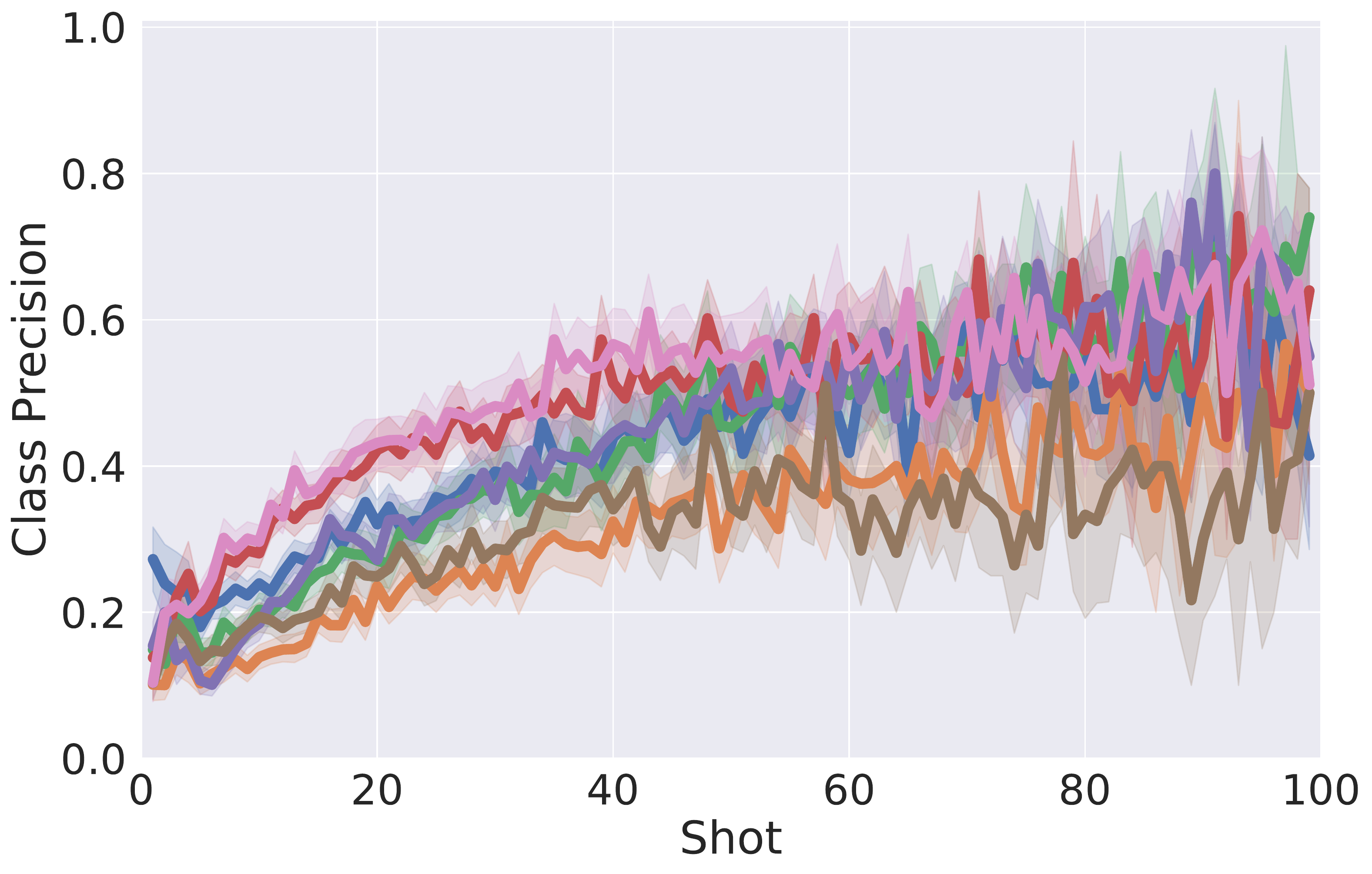}}\hspace{0.01cm}
\subfloat[Omniglot]{\includegraphics[width=0.24\textwidth]{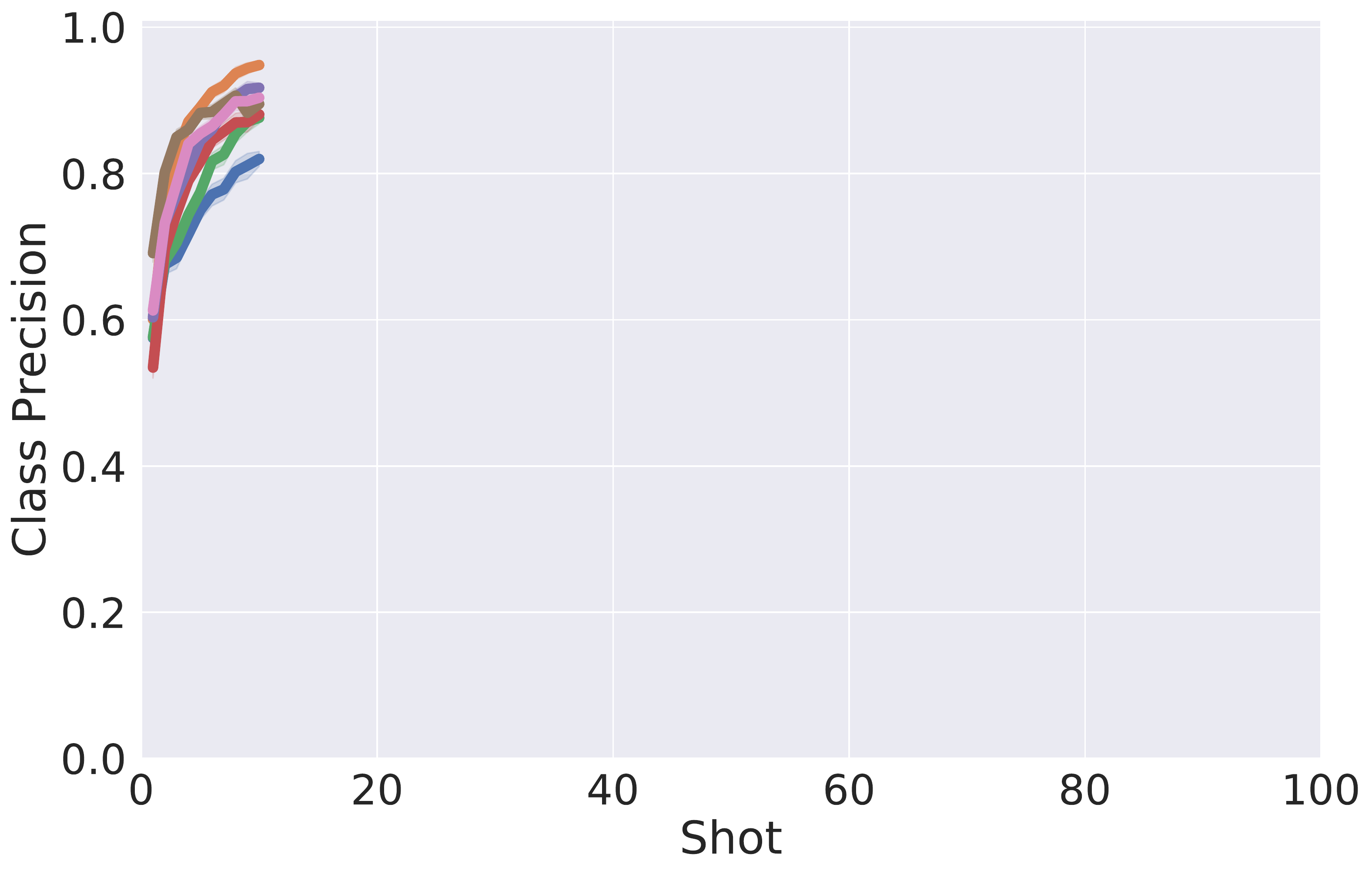}}\hspace{0.01cm}
\subfloat[Aircraft]{\includegraphics[width=0.24\textwidth]{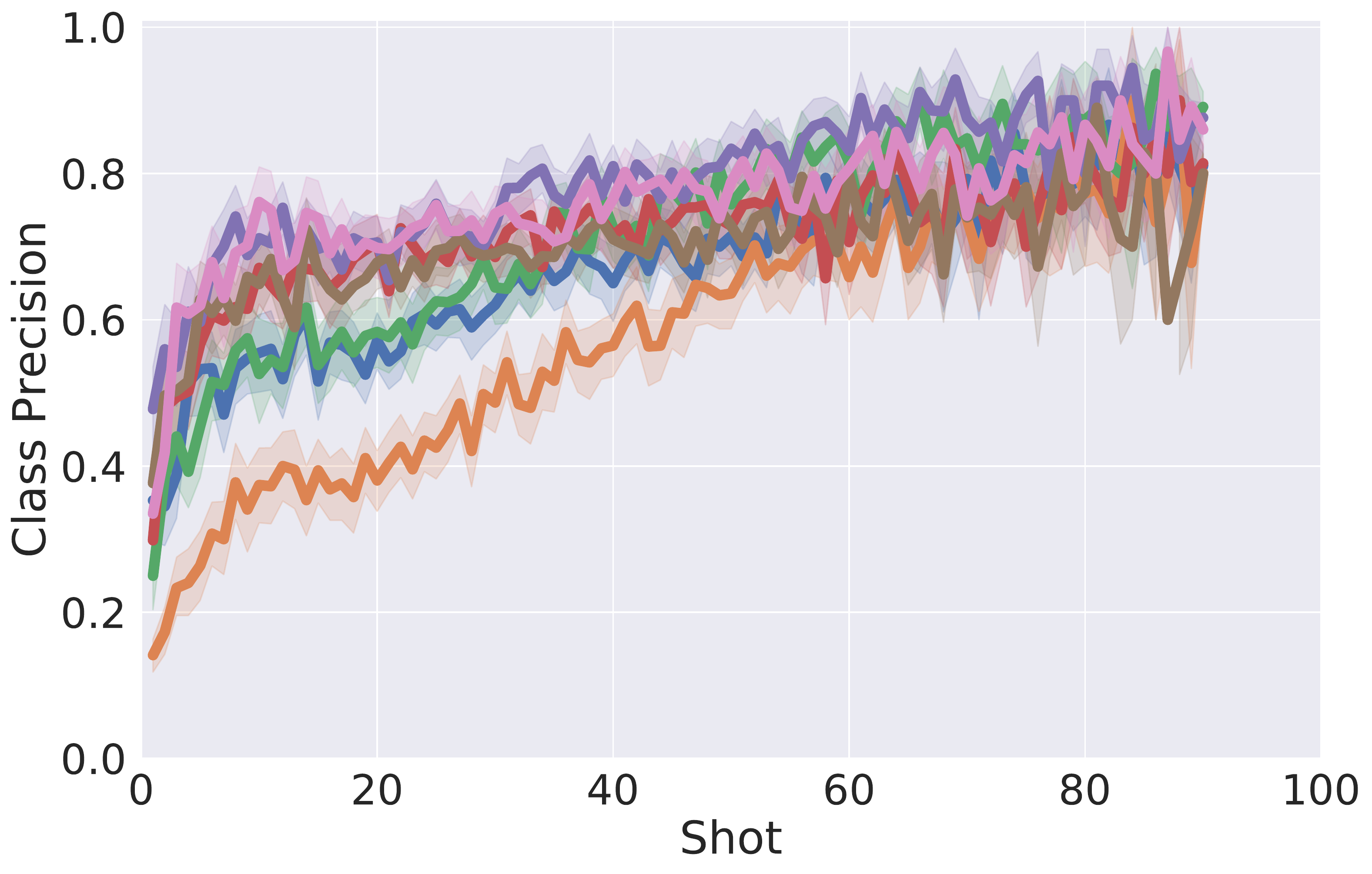}}\hspace{0.01cm}
\subfloat[Birds]{\includegraphics[width=0.24\textwidth]{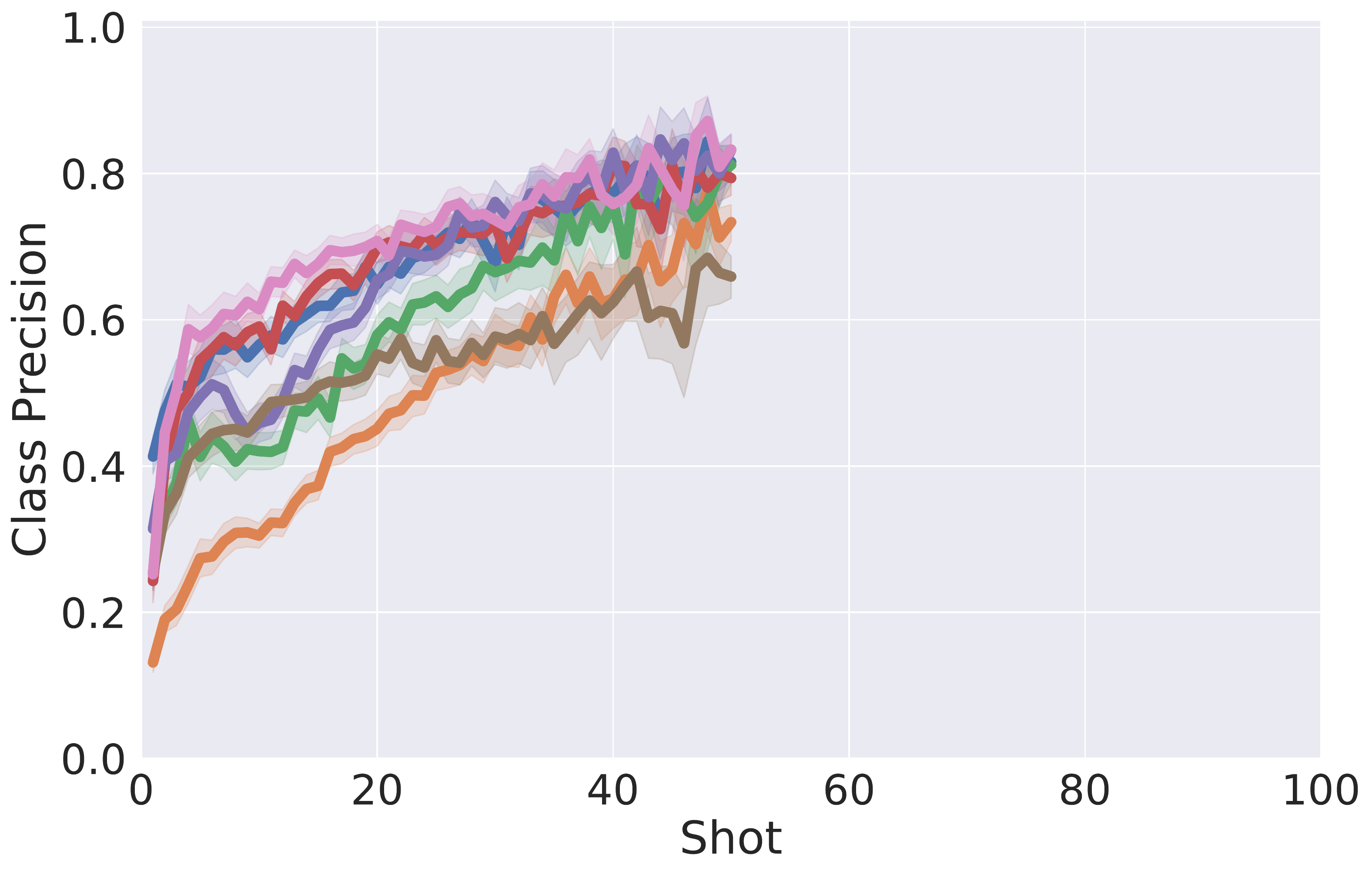}}\hspace{0.01cm}
\subfloat[Textures]{\includegraphics[width=0.24\textwidth]{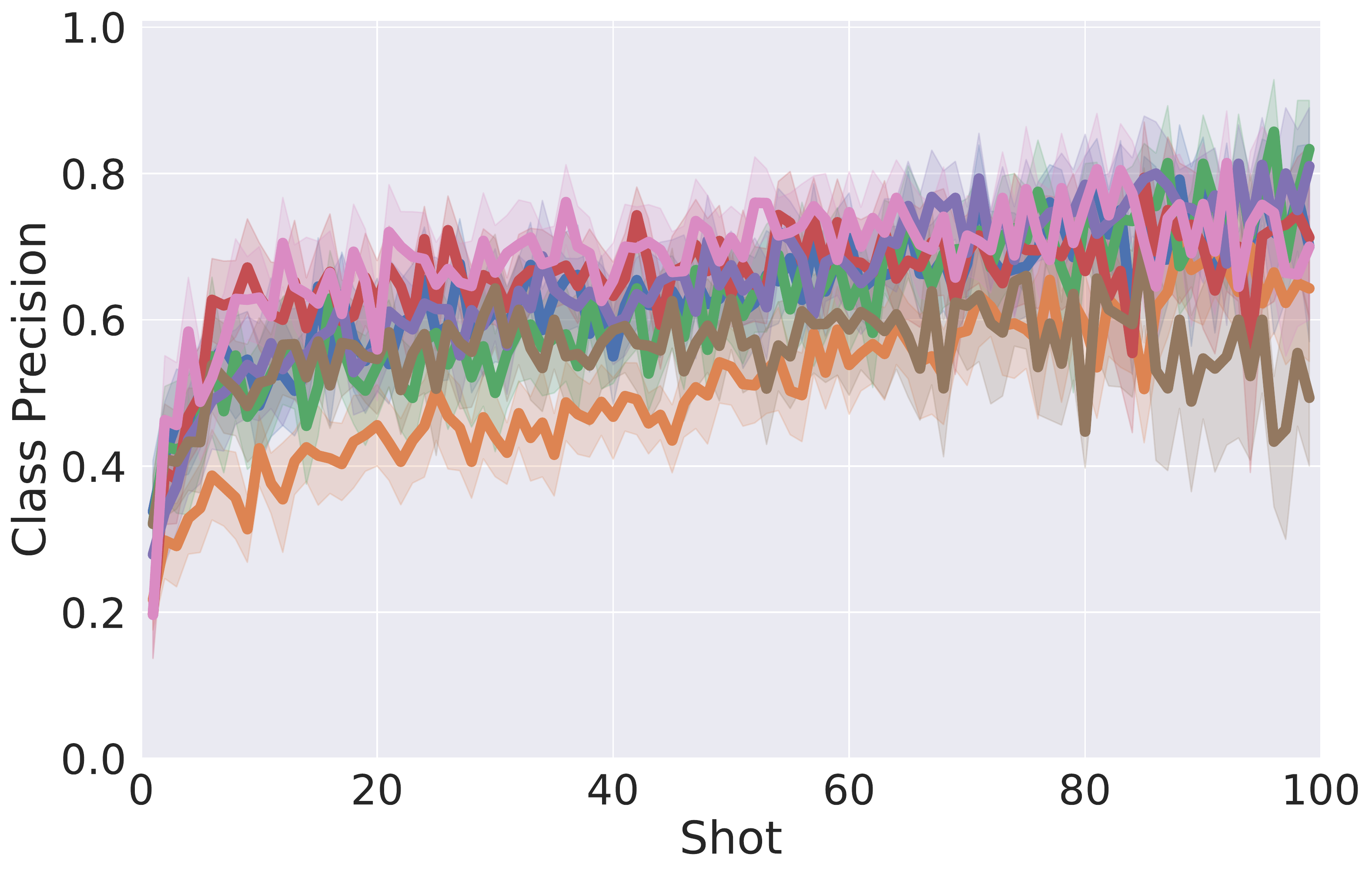}}\hspace{0.01cm}
\subfloat[Quickdraw]{\includegraphics[width=0.24\textwidth]{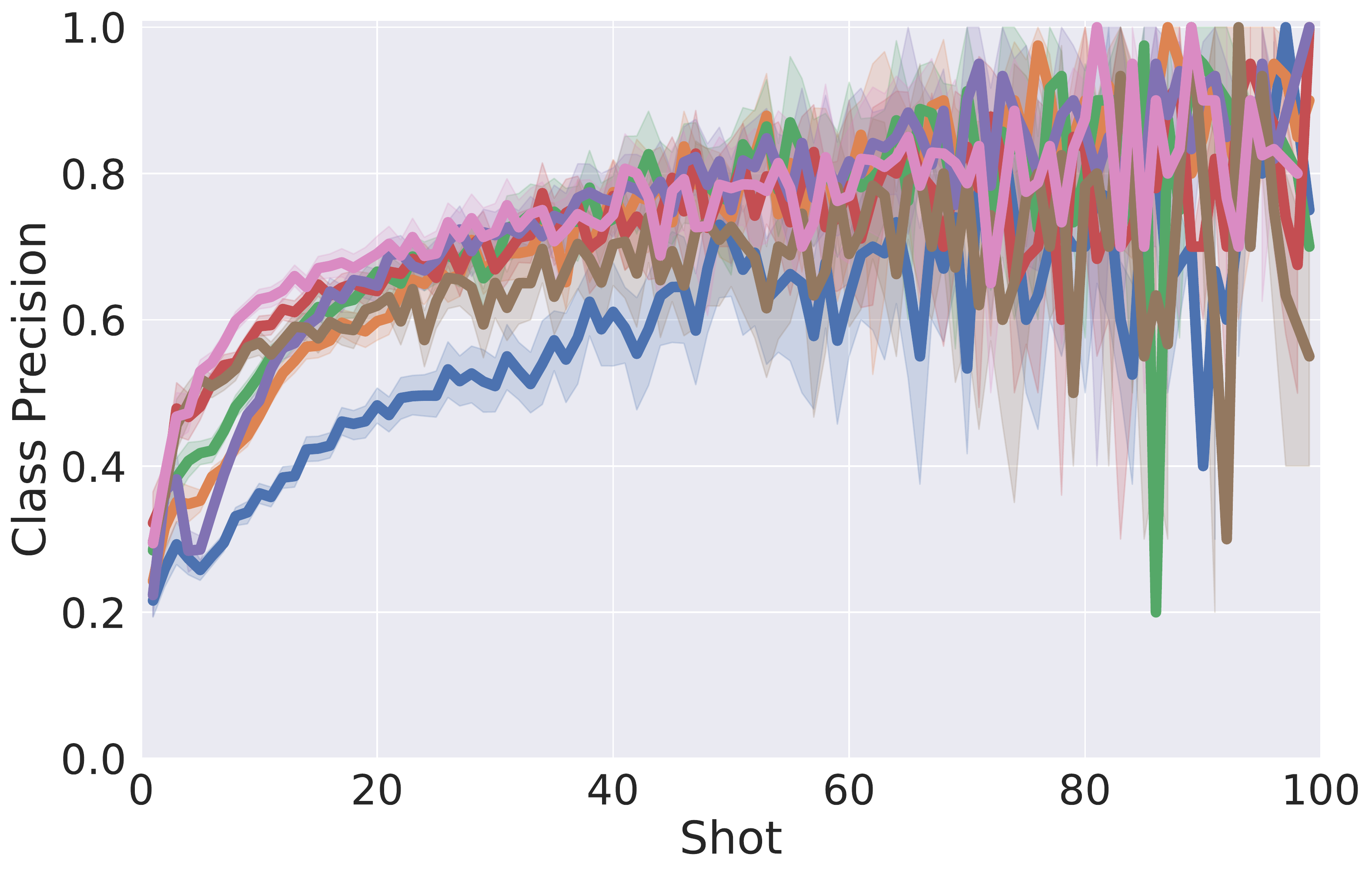}}\hspace{0.01cm}
\subfloat[Fungi]{\includegraphics[width=0.24\textwidth]{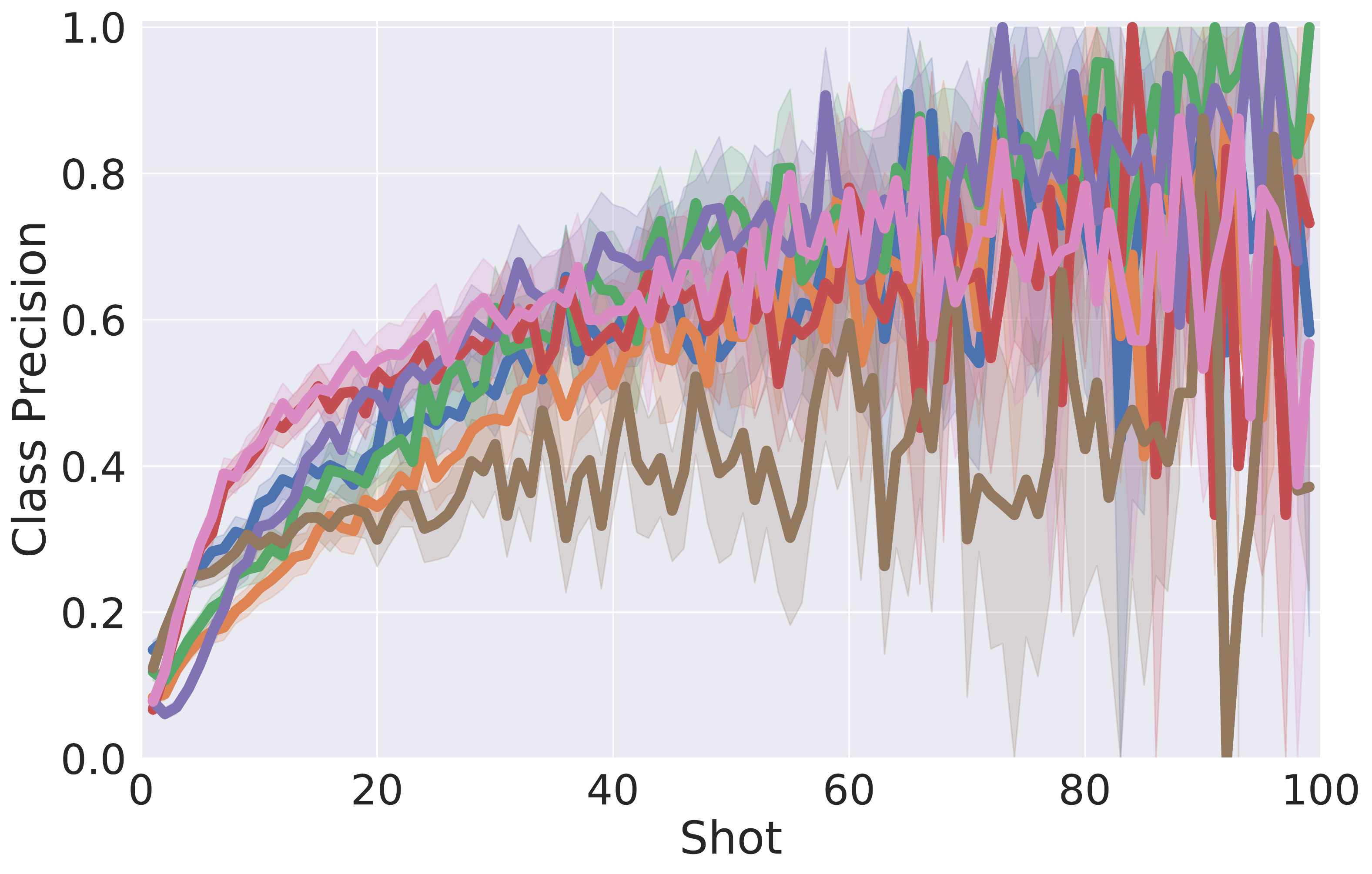}}\hspace{0.01cm}
\subfloat[VGG Flower]{\includegraphics[width=0.24\textwidth]{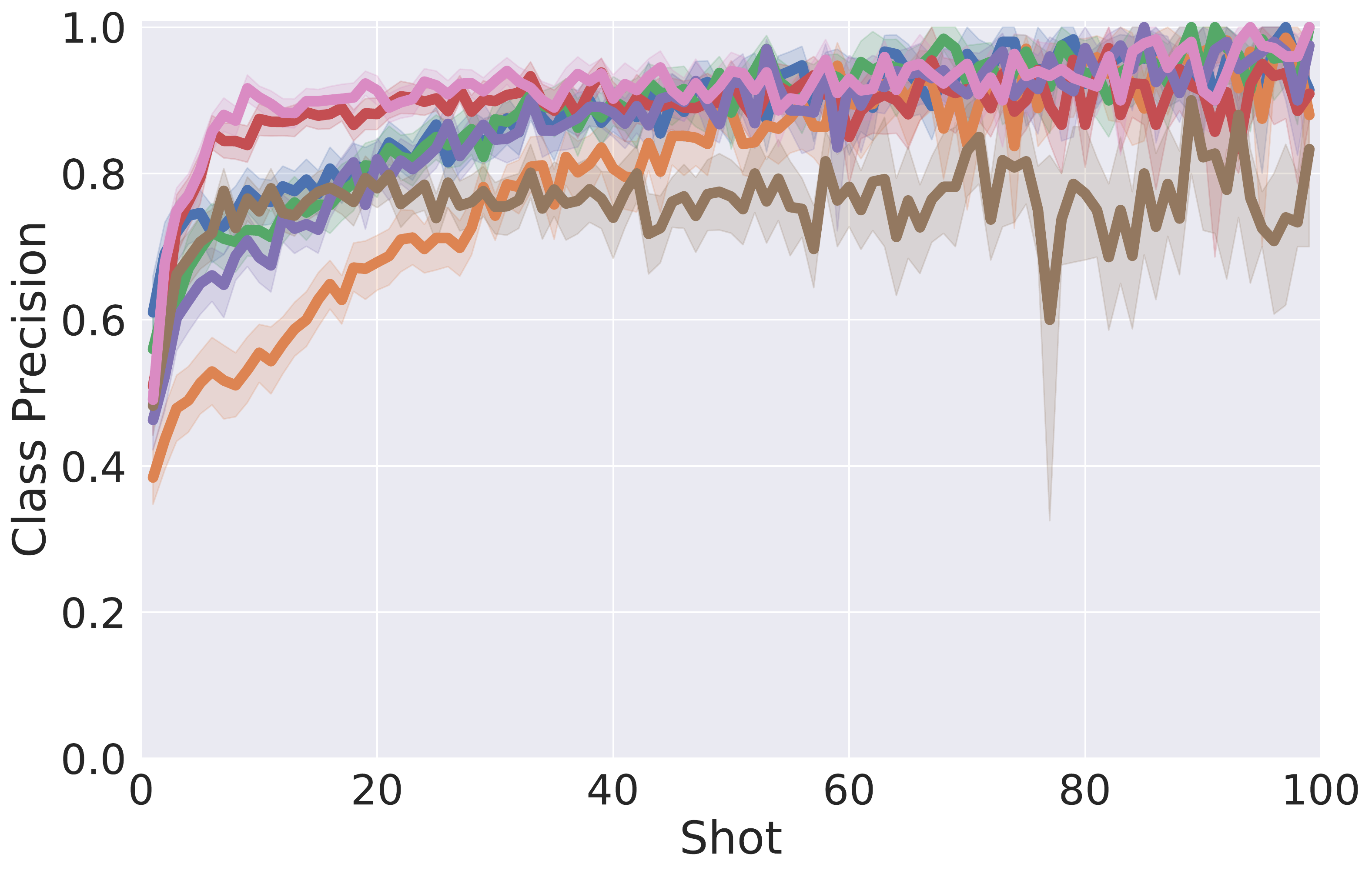}}\hspace{0.01cm}
\subfloat[Traffic Signs]{\includegraphics[width=0.24\textwidth]{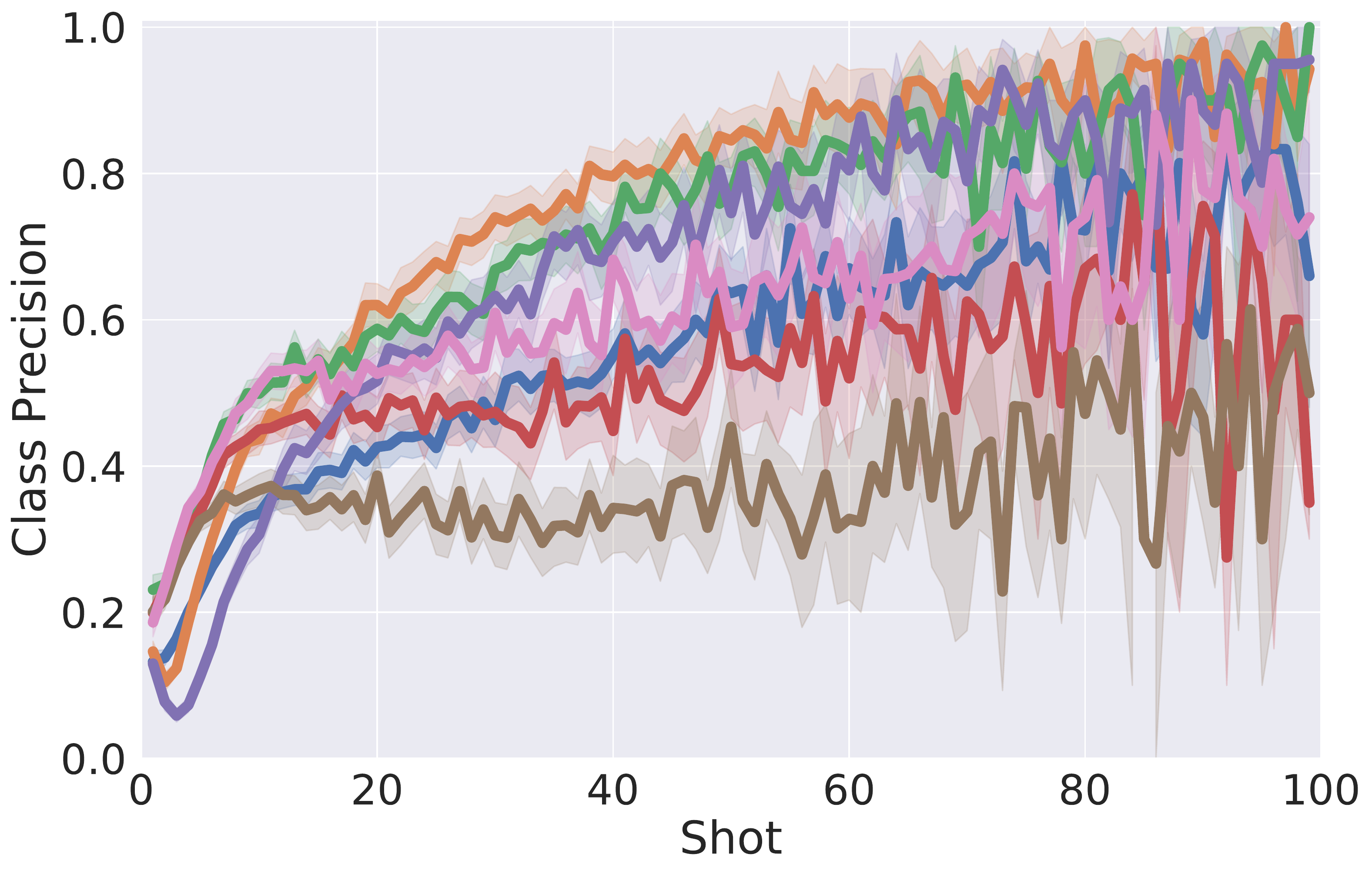}}\hspace{0.01cm}
\subfloat[MSCOCO]{\includegraphics[width=0.24\textwidth]{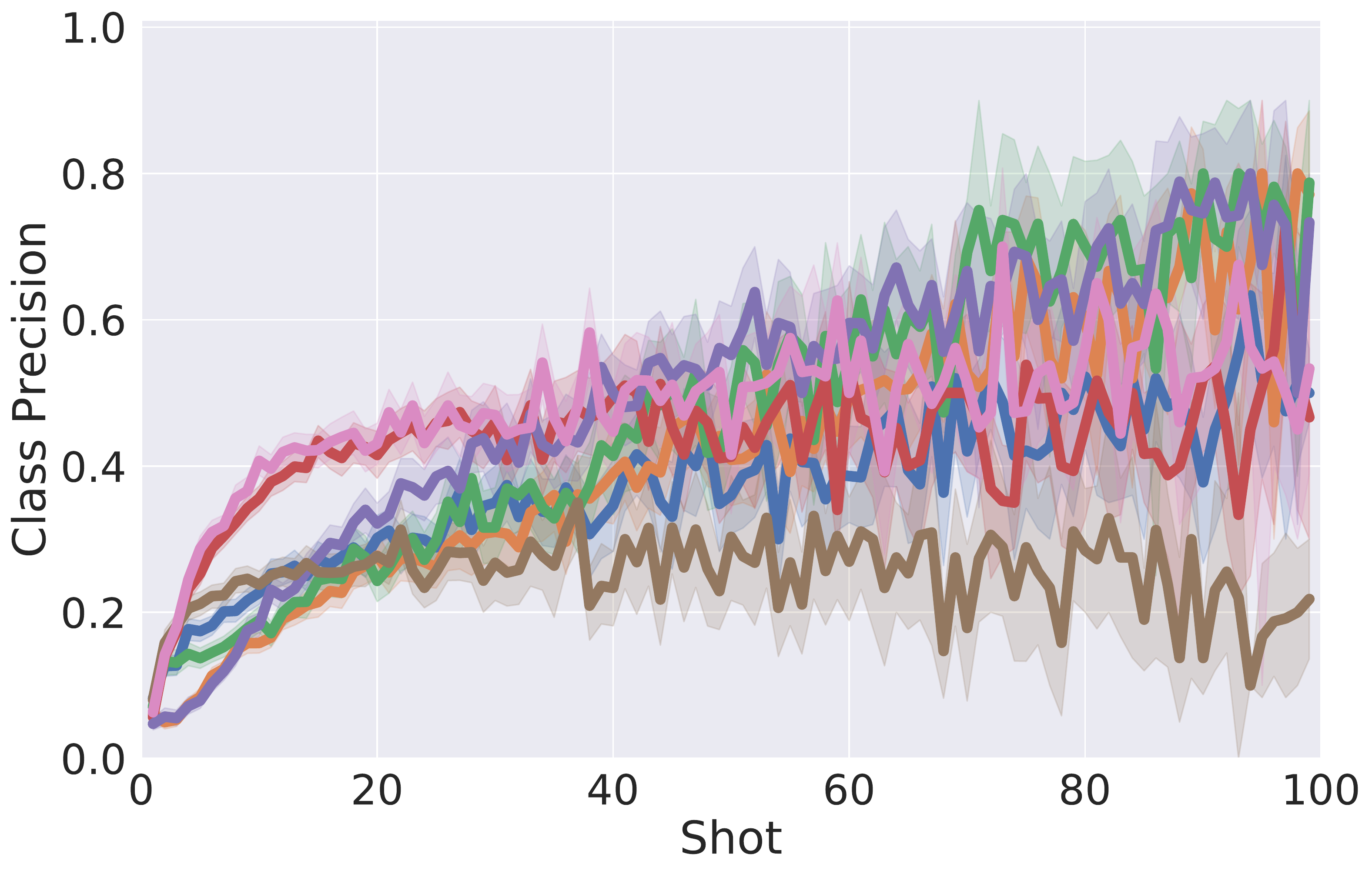}}\hspace{0.01cm}
\hspace{\fill}
\subfloat{\includegraphics[width=0.19\textwidth]{figures_iclr2020/ways_shots_legend.pdf}\hspace{0.01\textwidth}}\hspace{0.01cm}
	\caption{\label{fig:shots_analysis_per_dataset_all_datasets} The performance across different shots, with 95\% confidence intervals, shown separately for each evaluation dataset. All models had been trained on (the training splits of) all datasets.}
 \end{figure*}

In our previous analysis of performance across different shots and ways
(Figures~\ref{fig:imagenet_shots_analysis}, \ref{fig:imagenet_way_analysis},
and~\ref{fig:analyses_ways_and_shots_extra}), the performance is averaged over
all evaluation datasets. In this section we further break down those plots by
presenting the results separately for each dataset.
Figures~\ref{fig:way_analysis_per_dataset}
and~\ref{fig:shots_analysis_per_dataset} show the analysis of performance as a
function of ways and shots (respectively) for each evaluation dataset, for
models that were trained on ImageNet only. For completeness,
Figures~\ref{fig:way_analysis_per_dataset_all_datasets}
and~\ref{fig:shots_analysis_per_dataset_all_datasets} show the same for the
models trained on (the training splits of) all datasets.

\end{document}